\documentclass[10pt,journal,compsoc]{IEEEtran}

\usepackage{xcolor,soul,framed} %,caption

\colorlet{shadecolor}{yellow}
\usepackage[pdftex]{graphicx}
\graphicspath{{../pdf/}{../jpeg/}}
\DeclareGraphicsExtensions{.pdf,.jpeg,.png}

\usepackage[cmex10]{amsmath}
\usepackage{array}
\usepackage{mdwmath}
\usepackage{mdwtab}
\usepackage{eqparbox}
\usepackage{url}
\usepackage{graphicx}
\usepackage{amsmath}
\usepackage{amsfonts}
\usepackage{algorithm}
\usepackage{algpseudocode}
\usepackage{multirow}
\usepackage{footnote}
\usepackage{bm}
\usepackage{makecell}
\usepackage{pdflscape}
\usepackage{afterpage}
\usepackage{bbding}
\usepackage{lscape}
\usepackage{booktabs}
\usepackage{tikz}

\usepackage{pgfplots}
\pgfplotsset{compat=1.12}
\usepackage{subcaption}
    \pgfplotsset{
        layers/my layer set/.define layer set={
            background,
            main,
            foreground
        }{ },
        set layers=my layer set,
    }

\usepackage[square,sort,comma,numbers]{natbib} % multiple citation or 
\usepackage[top=0.4in,bottom=0.4in,left=0.4in,textwidth=7.7in]{geometry}

\usepackage[pagebackref=false,breaklinks=false,linkcolor=black,anchorcolor=black, citecolor=black,colorlinks,bookmarks=true]{hyperref}
\newcommand{\etal}{\textit{et al}. }
\newcommand{\etc}{\textit{etc}. }
\newcommand{\ie}{\textit{i}.\textit{e}., }
\newcommand{\eg}{\textit{e}.\textit{g}., }
\usepackage{amsfonts}
\newcommand{\tickYes}{\checkmark}
\usepackage{pifont}
\newcommand{\tickNo}{\hspace{1pt}\ding{55}}

\begin{document}

\title{Deep Learning for Instance Retrieval: A Survey}
 
\author{\IEEEauthorblockN{Wei Chen\thanks{W. Chen %(w.chen@liacs.leidenuniv.nl) 
is with Academy of Advanced Technology Research of Hunan, Changsha, China %and also with Leiden Institute of Advanced Computer Science, Leiden University, the Netherlands.
},
Yu Liu\thanks{Y. Liu %(liuyu8824@dlut.edu.cn) 
is with DUTRU International School
of Information Science and Engineering, Dalian University of Technology, China.},
Weiping Wang\thanks{W. Wang %(wpwang@nudt.edu.cn) 
is with Academy of Advanced Technology Research of Hunan, Changsha.},
Erwin M. Bakker\thanks{E. Bakker, T. Georgiou, and M. Lew %((E.M.Bakker, t.k.georgiou, m.s.lew)@liacs.leidenuniv.nl) 
are with Leiden Institute of Advanced Computer Science, Leiden University, the Netherlands.},
Theodoros Georgiou,\\
Paul Fieguth\thanks{P. Fieguth %(pfieguth@uwaterloo.ca) 
is with the Department of Systems Design Engineering, University of Waterloo, Canada.},
Li Liu\thanks{L. Liu is with Academy of Advanced Technology Research of Hunan, Changsha, China, and also with Center for Machine Vision and Signal Analysis, University of Oulu, Finland.},
and Michael S. Lew
\thanks{Corresponding author: Li Liu, li.liu@oulu.fi, dreamliu2010@gmail.com}
\thanks{This work was supported by China Scholarship Council (No. 201703170183), the Academy of Finland under grant 331883, Infotech Project FRAGES, and the National Natural Science Foundation of China under Grant 61872379, 62022091, 61825305, and 62102061.
}
}}

% \markboth{ In preparation for submitting to IEEE TPAMI}%
\markboth{IEEE Transactions on Pattern Analysis and Machine Intelligence}
{Chen \MakeLowercase{\textit{et al.}}: }

\IEEEtitleabstractindextext{%
\begin{abstract}
In recent years a vast amount of visual content has been generated and shared from many fields, such as social media platforms, medical imaging, and robotics. This abundance of content creation and sharing has introduced new challenges, particularly that of searching databases for similar content --- Content Based Image Retrieval (CBIR) --- a long-established research area in which improved efficiency and accuracy are needed for real-time retrieval. Artificial intelligence has made progress in CBIR and has significantly facilitated the process of instance search. In this survey we review recent instance retrieval works that are developed based on deep learning algorithms and techniques, with the survey organized by deep feature extraction, feature embedding and aggregation methods, and network fine-tuning strategies. Our survey considers a wide variety of recent methods, whereby we identify milestone work, reveal connections among various methods and present the commonly used benchmarks, evaluation results, common challenges, and propose promising future directions. 
\end{abstract}

\begin{IEEEkeywords}
Content based image retrieval, Instance retrieval, Deep learning, Convolutional neural networks, Literature survey
\end{IEEEkeywords}}

\maketitle
\IEEEdisplaynontitleabstractindextext
\IEEEpeerreviewmaketitle

\IEEEraisesectionheading{\section{Introduction}
\label{sec:intro}}

\IEEEPARstart{C}{ontent} Based Image Retrieval (CBIR) is the problem of searching for relevant images in an image gallery by analyzing the visual content (colors, textures, shapes, objects \emph{etc.}), given a query image \cite{smeulders2000content},\cite{lew2006content}. CBIR has been a longstanding research topic in the fields of computer vision and multimedia \cite{smeulders2000content},\cite{lew2006content}. With the exponential growth of image data, the development of appropriate information systems that efficiently manage such large image collections is of utmost importance, with image searching being one of the most indispensable techniques. Thus there is a nearly endless potential for applications of CBIR, such as person/vehicle reidentification \cite{zheng2015scalable},\cite{liu2019group}, landmark retrieval \cite{noh2017largescale}, %artwork search \cite{arandjelovic2011smooth}, 
remote sensing \cite{chaudhuri2019siamese}, online product searching \cite{liu2016deepfashion}.

Generally, CBIR methods can be grouped into two different tasks \cite{gordo2017beyond},\cite{barz2021content}: Category level Image Retrieval (CIR) and Instance level Image Retrieval (IIR). The goal of CIR is to find an arbitrary image representative of the same category as the query (\eg dogs, cars) \cite{wang2019multi},\cite{wang2014learning}.  By contrast, in the IIR task, a query image of a particular instance (\eg the Eiffel Tower, my neighbor's dog) is given and the goal is to find images containing the same instance that may be captured under different conditions like different imaging distances, viewing angles, backgrounds,  illuminations, and weather conditions (reidentifying exemplars of the same instance) \cite{babenko2015aggregating},\cite{zheng2018sift}. The focus of this survey is the IIR task\footnote{If not further specified, ``image retrieval'', ''IIR'', and ``instance retrieval'' are considered equivalent and will be used interchangeably.}.

In many real world applications, IIR is usually to find a desired image requiring a search among thousands, millions, or even billions of images. Hence, searching efficiently is as critical as searching accurately, to which continued efforts have been devoted  \cite{babenko2015aggregating},\cite{zhang2016hyperlink},\cite{kalantidis2016cross}. To enable accurate and efficient retrieval over a large-scale image collection, \emph{developing compact yet discriminative feature representations} is at the core of IIR.

\iffalse
%%% Fig 01
\begin {figure}[!t]
\centering
\begin{tabular}{c}
\includegraphics[width=0.9\columnwidth]{./Figures/TheProblem.pdf} \\
(\textit{a}) The CBIR problem \\
\includegraphics[width=0.9\columnwidth]{./Figures/InstancevsCategory.pdf} \\
(\textit{b}) instance-level retrieval versus category-level retrieval
 \end{tabular}
\caption{ Illustration of (a) the CBIR problem, and (b)
CBIR categorization. The images in green frame are retrieved correctly, while the ones in red frame are matched incorrectly.
}
\label{category_vs_instance}
\vspace{-1em}
\end {figure}
\fi

 %%% Tab 01
\begin{table*}[!ht]
\caption{A summary and comparison of the primary surveys in the field of image retrieval.}\label{SurveyCompare}
\vspace{-1em}
\centering
\setlength{\tabcolsep}{3.0mm}\footnotesize
\begin{tabular}{!{\vrule width0.3bp}p{5cm}|p{0.5cm}|p{1.6cm}|p{9.2cm}!{\vrule width0.3bp}}
\Xhline{1.0pt}
\multicolumn{1}{|c|}{ \multirow{2}*{ Title} } &  \multirow{2}*{ $\!\!\!\!$ Year}  & \multirow{2}*{ Published in}  &  \multicolumn{1}{c|}{ \multirow{2}*{ Main Content } }  \\ 
\hline

\multirowcell{2}{Content-Based Image Retrieval at the \\ End of the Early Years \cite{smeulders2000content} } & \multirow{2}*{2000}   &  \multirowcell{2}{ TPAMI } & \multirowcell{2}{This paper discusses the steps for image retrieval systems, including image \\ processing, feature extraction, user interaction, and similarity evaluation. }\\

\hline

\multirowcell{2}{Image Search from Thousands to Billions \\in 20 Years \cite{zhang2013image}}  & \multirow{2}*{2013}   &  \multirowcell{2}{ TOMM}  & \multirowcell{2}{This paper gives a good presentation of image search achievements from \\ 1970 to 2013, but the methods are not deep learning-based.} \\

\hline

\multirowcell{2}{Deep Learning for Content-Based Image \\ Retrieval: A Comprehensive Study \cite{wan2014deep}}  & \multirow{2}*{2014}   &  \multirowcell{2}{ ACM MM}  & \multirowcell{2}{This paper introduces supervised metric learning methods for fine-tuning \\ AlexNet. Details of instance-based image retrieval are limited.}\\
\hline

\multirowcell{2}{Semantic Content-based Image Retrieval:\\ A Comprehensive Study \cite{alzu2015semantic}}  & \multirow{2}*{2015}   &  \multirowcell{2}{ JVCI }  & \multirowcell{2}{This paper presents a comprehensive study about CBIR using traditional \\ methods; deep learning is introduced as a section with limited details.}\\
\hline

\multirowcell{3}{Socializing the Semantic Gap: A Compa-\\rative Survey on Image Tag Assignment, \\Refinement, and Retrieval \cite{li2016socializing}} & \multirowcell{1}{2016}  &  \multirowcell{3}{CSUR} &  \multirowcell{3}{A taxonomy is introduced to structure the growing literature of image \\ retrieval.  Deep learning methods for feature learning is introduced as \\future work.} \\
\hline

\multirowcell{2}{Recent Advance in Content based Image \\ Retrieval: A  Literature Survey \cite{zhou2017recent}}  & \multirow{2}*{2017}   &  \multirowcell{2}{ arXiv }  & \multirowcell{2}{This survey presents image retrieval from 2003 to 2016.  Neural networks \\ are introduced in a section and  mainly discussed as a future direction.}\\
\hline

\multirowcell{3}{Information Fusion in Content-based \\Image  Retrieval: A Comprehensive \\ Overview \cite{piras2017information}} & \multirowcell{1}{2017}  & \multirowcell{3}{Information\\Fusion} &  \multirowcell{3}{This paper presents information fusion strategies in CBIR. \\ Deep convolutional networks for feature learning are introduced \\ briefly but not covered thoroughly.} \\
\hline

\multirowcell{2} { $\!\!\!$ A Survey on Learning to Hash \cite{wang2018survey} }   & \multirow{2}*{2018}  & \multirowcell{2}{ TPAMI} &  \multirowcell{2}{This paper focuses on hash learning algorithms and introduces the \\ similarity-preserving methods and discusses their relationships. } \\
\hline

\multirowcell{2}{$\!\!\!$ SIFT Meets CNN: A Decade Survey of \\  Instance Retrieval \cite{zheng2018sift}}  & \multirow{2}*{2018}   &  \multirowcell{2}{ TPAMI}  & \multirowcell{2}{This paper presents a comprehensive review of instance retrieval based on \\SIFT and CNN methods.}\\
\hline

\multirowcell{3}{ \bf Deep Learning for Instance \\ \bf Retrieval: A Survey }  & \multirowcell{1}{\bf 2021}  & \multirowcell{3}{ \bf Ours} & \multirowcell{3}{\bf Our survey focuses on deep learning methods. We expand the review \\ \bf  with indepth details on IIR, including methods of feature extraction, \\ \bf feature embedding and aggregation, and network fine-tuning. }
 \\
\Xhline{1.0pt}
\end{tabular}
\end{table*}

During the past two decades, startling progress has been witnessed in image  representation which mainly consists of two important periods, \ie feature engineering and deep learning. In the feature engineering era, the field was dominated by various milestone handcrafted image representations like SIFT \cite{lowe2004distinctive} and Bag of visual Words (BoW) \cite{sivic2003video}. The deep learning era was reignited in 2012 when a Deep Convolutional Neural Network (DCNN) referred as ``AlexNet'' \cite{krizhevsky2012imagenet} won the first place in the ImageNet classification contest  with a breakthrough reduction in classification error rate. Since then, the dominating role of SIFT like local descriptors has been replaced by data driven Deep Neural Networks (DNNs) which can learn powerful feature representations with multiple levels of abstraction directly from data. During the past decade, DNNs have set the state of the art in various classical computer vision tasks, including image classification~\cite{krizhevsky2012imagenet},\cite{he2016deep}, object detection~\cite{ren2015faster}, semantic segmentation \cite{minaee2021image}, %image synthesis \cite{shamsolmoali2021image},
and image retrieval~\cite{wan2014deep}.

Given this period of rapid evolution, the goal of this paper is to provide a comprehensive survey of the recent achievements in IIR. In comparison with existing excellent surveys on traditional image retrieval \cite{zheng2018sift},\cite{alzu2015semantic},\cite{li2016socializing},\cite{zhou2017recent}, as contrasted in Table~\ref{SurveyCompare}, our focus in this paper is reviewing deep learning based methods for IIR, particularly on issues of retrieval accuracy and efficiency.
%, including feature extraction, feature fusion, feature embedding and aggregation, and network fine-tuning for IIR.

% \vspace{-0.5em}

\subsection{Summary of Progress since 2012}
\label{Summary_of_Progress}

After the highly successful image classification implementation of AlexNet~\cite{krizhevsky2012imagenet}, significant exploration of DCNNs for instance retrieval tasks was undertaken and representative methods are shown in Figure \ref{fig1}. Building on these methods, more recent progress for IIR can be achieved from the perspectives of off-the-shelf models and fine-tuned models, which form the basis for this survey.

Off-the-shelf models, based on DCNNs with fixed parameters \cite{sharif2014cnn},\cite{gong2014multi},\cite{yosinski2014transferable}, extract features at image scales or patch scales, which correspond to single-pass and multiple-pass schemes, respectively. These methods focus on effective feature use, for which researchers have proposed embedding and aggregation methods, such as R-MAC \cite{tolias2015particular}, CroW \cite{kalantidis2016cross}, and SPoC \cite{babenko2015aggregating} to promote the discriminativity of the extracted features. Fine-tuned models, based on DCNNs in which parameters are updated \cite{sharif2014cnn}, are more popular since deep networks themselves have been investigated extensively. To learn better retrieval features, researchers have proposed to improve the network architectures and/or update the pre-stored parameters \cite{yosinski2014transferable}. 

This survey will explore recent progress in detail in the context of the three following themes:

\begin{figure*}[!t]
\centering
\includegraphics[width=\textwidth]{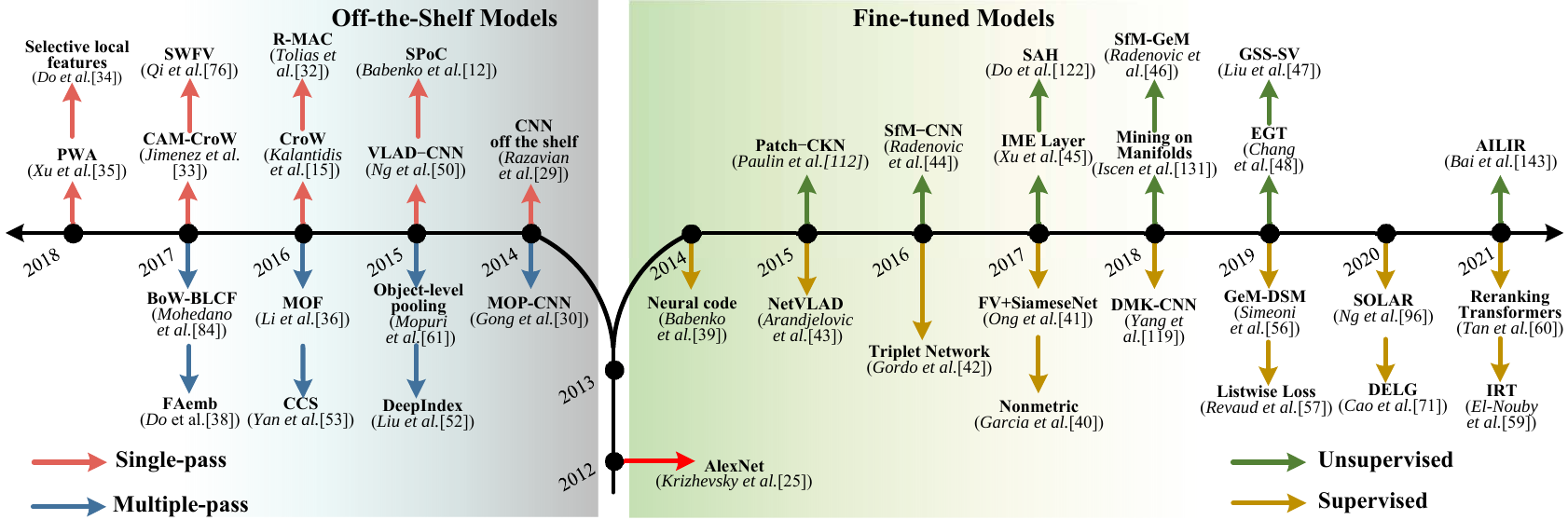} 
\vspace{-1.5em}
\caption{Representative methods in IIR. Off-the-shelf models have model parameters which are not further updated or tuned when extracting retrieval features. For single-pass schemes, the key step is the feature embedding and aggregation to promote the discriminativity of the extracted image-level activations \cite{kalantidis2016cross},\cite{jimenez2017class},\cite{do2018selective},\cite{xu2018unsupervised}, whereas for multiple-pass schemes the goal is to extract instance features at region scales and eliminate image clutter as much as possible \cite{gong2014multi},\cite{li2016exploiting},\cite{sharif2015baseline},\cite{do2017embedding}. In contrast, for fine-tuned models, the model parameters are tuned towards the retrieval task and address the issue of domain shifts. For supervised fine-tuning, the key step lies in the design of objective functions and sample sampling strategies \cite{babenko2014neural},\cite{garcia2017learning},\cite{ong2017siamese},\cite{gordo2016deep},\cite{arandjelovic2016netvlad}, while the success of unsupervised fine-tuning is to mine the relevance among training samples \cite{radenovic2016cnn},\cite{xu2017iterative},\cite{radenovic2018fine},\cite{liu2019guided},\cite{chang2019explore}. See Sections \ref{Retrieval_with_Off_the_Shelf_DCNN_Models} and \ref{Retrieval_via_Learning_DCNN_Representations} for details.}

\label{fig1}
\end{figure*}

\medskip

%\noindent (1) \emph{Improvements in Network Architectures} \hfill %(Section \ref{Popular_Backbone_Architectures})%
%\smallskip
%
%
%\noindent
%What network architectures and variations have been proposed?  DCNNs with more stacked layers (\ie with ever more linear convolution filters and non-linear activation functions) offer a stronger ability to extract high-level abstract and semantic-aware features~\cite{he2016deep,simonyan2014very}. It is also possible to concatenate multi-scale features in parallel \cite{szegedy2015going}. Recently,  convolution-free models that only rely on transformer layers have shown competitive performance and been used as a powerful alternative to DCNNs, such as IRT \cite{el2021training}, reranking Transformers \cite{tan2021instance}.
%\medskip

\noindent (1) \emph{Deep Feature Extraction} \hfill (Section \ref{Deep_Feature_Extraction})%
\smallskip

\noindent 
One key step in IIR is to make the descriptors as semantic-aware \cite{he2016deep},\cite{simonyan2014very} as possible. For this, some recent work focus on the input data of DCNNs, thereby instance features can be extracted from the whole image, \eg CroW \cite{kalantidis2016cross}, VLAD-CNN \cite{yue2015exploiting} or from image patches, \eg MOP-CNN \cite{gong2014multi}, FAemb \cite{do2017embedding}. For instance, evaluated on the Holidays dataset \cite{jegou2008hamming}, patch-level input scheme can improve mAP by 8.29\% compared to the results (70.53\%) achieved using image-level input \cite{gong2014multi}. Others focus on exploring different feature extractors, \eg one layer of a given DCNN, to get the output activations. Initially, fully-connected layers are usually chosen to extract features \cite{liu2015deepindex},\cite{yan2016cnn}, and then convolutional layers are popularly used \cite{babenko2015aggregating},\cite{tolias2015particular}. Afterwards, some work leverage the complementarity of different extractors to explore layer-level fusion, such as  MoF \cite{li2016exploiting}, and model-level fusion, such as DeepIndex  \cite{liu2015deepindex} to promote the retrieval performance.

\medskip

\noindent 
(2) \emph{Feature Embedding and Aggregation} \hfill (Section \ref{Deep_Feature_Enhancement})%
\smallskip

\noindent 
%How do we promote the discriminativity of the image-level or region-level activations extracted from a certain DCNN layer?
Recent work revisit the classical embedding and aggregation methods and apply on deep features. Most work have a preference towards mapping individual vector from convolutional layer \cite{sivic2003video},\cite{jegou2010aggregating},\cite{perronnin2007fisher} and then aggregating into a global feature. The mapping process can be realized by using a pre-trained codebook (\ie built separately), such as VLAD-CNN \cite{yue2015exploiting}, DeepIndex \cite{liu2015deepindex} or learned as parameters during training (built simultaneously), such as NetVLAD \cite{arandjelovic2016netvlad}, GeM-DSM \cite{simeoni2019local}. Some work aggregate local features into a global one by direct pooling \cite{piras2017information} or sophisticated pooling-based methods \cite{babenko2015aggregating} without the aggregation operation, such as R-MAC \cite{tolias2015particular}.

% How do we promote the discriminativity of the image-level or region-level activations extracted from a certain DCNN layer? Individual feature vectors can be mapped into a high-dimensional space \cite{gong2014multi,sivic2003video,jegou2010aggregating,perronnin2007fisher} and then aggregated into a global feature. This mapping process can be realized by using a pre-trained codebook (\ie built separately) \cite{liu2015deepindex,yue2015exploiting} or learned as parameters during training (built simultaneously) \cite{arandjelovic2016netvlad,simeoni2019local}. Features may be aggregated into a global feature by direct pooling \cite{piras2017information} or sophisticated pooling-based methods \cite{babenko2015aggregating,tolias2015particular} without feature embedding. 
%Further, hashing methods \cite{wang2018survey} encode the real-valued features into binary codes to significantly improve retrieval efficiency.%  The feature embedding strategy can significantly influence the efficiency of image retrieval.
\medskip

\noindent 
(3) \emph{Network Fine-tuning for Learning Representations} \hfill (Section \ref{Retrieval_via_Learning_DCNN_Representations})%
\smallskip

\noindent 
DCNNs pretrained on source datasets for image classification are influenced by domain shifts when performing retrieval tasks on new datasets. Thus, it is necessary to fine-tune deep networks to the specific domain \cite{babenko2014neural}, by using supervised or unsupervised fine-tuning methods. As depicted in Figure \ref{fig1}, recent supervised fine-tuning methods focus on designing objective functions (\eg Listwise loss \cite{revaud2019learning}) and sample sampling strategies, such as NetVLAD \cite{arandjelovic2016netvlad}, Triplet Network \cite{gordo2016deep}. Unsupervised methods focus on mining the relevance among training samples by using clustering, such as SfM-GeM \cite{radenovic2018fine} or manifold learning, such as AILIR \cite{bai2018regularized}.  Recently,  convolution-free models that only rely on transformer layers have shown competitive performance and been used as a powerful alternative to DCNNs, such as IRT \cite{el2021training}, reranking Transformers \cite{tan2021instance}.

%The key ingredient of the success of supervised fine-tuning is the design of objective functions and sample  sampling  strategies \cite{gordo2016deep}, \cite{arandjelovic2016netvlad} while the success of unsupervised approaches is to mine the relevance among training samples \cite{xu2017iterative}, \cite{radenovic2018fine}.

\subsection{Key Challenges}
\label{Keychallenges}
The goal of each of the preceding three themes is to address the competing objectives of \emph{accuracy} and \emph{efficiency}, with both objectives continuing to present challenges:

% \textbf{1) Accuracy related challenges.} For IIR, the differences between query image and database images are mainly caused by imaging environment changes like variations in scale, rotation \emph{etc} \cite{liu2019guided}. These lead to the significant variations in the object's appearance, which is called the sensory gap \cite{smeulders2000content}. In addition, in CBIR applications, it has been generally agreed that the user searches for semantic similarity, but computers can only compute visual similarity of data driven features, which is called the semantic gap \cite{smeulders2000content,barz2021content,li2016socializing}. Usually, there are different object instances to search among, and some negative instances may share very high similarity with the query example. Moreover, there are many objects may be even challenging to search, such as textureless objects \cite{cao2017local}, objects with reflective surface \cite{mishchuk2017working} and cluttered backgrounds \cite{zhang2016hyperlink}, % transparent objects \cite{???}, nonrigid objects \cite{???}, 
% \etc Therefore, there are two pivotal subproblems to achieve accurate instance retrieval. First, how to extract discriminative yet invariant features to represent the unique characteristic of instances (maybe from a global or local perspective)? Second, how to evaluate the similarity between a pair of images based on their representations?

\textbf{A) Accuracy related challenges} depend on the input data, the feature extractor, and the way in which the extracted features are processed:
\begin{itemize}

\item Invariance challenge: The instance in an image may be rotated, translated, or scaled differently, so the final features are affected by these transformations and retrieval accuracy may be degraded \cite{gong2014multi}. It is necessary to incorporate invariance into the IIR pipeline \cite{reddy2015object},\cite{morere2017nested}.

\item Distraction challenge: IIR systems may need to focus on only a certain object, or even only a small portion. DCNNs may be affected by image clutter or background, so multiple-pass schemes have been examined where region proposals are studied before feature extraction.

\item Discriminativity challenge: Deep features for IIR are needed to be as discriminative as possible to distinguish instances with small differences, leading to many explorations in feature processing.  These include feature embedding and aggregation methods, to promote feature discriminativity; and attention mechanisms, to highlight the most relevant regions within the extracted features or to enable deep networks to focus on regions of interest. 
\item Fine-tune challenge: DCNNs can be fine-tuned as powerful extractors to capture fine semantic distinctions among instances. These explorations have offered improved accuracy, however the area remains a major challenge. 
\end{itemize}

%%% 
\begin{figure*}[!t]
\centering
\includegraphics[width=0.95\textwidth]{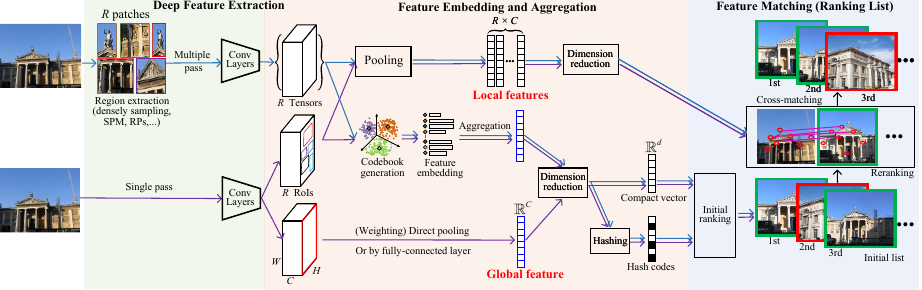}
 \vspace{-0.5em}
\caption{General framework of IIR, which includes feature extraction on image or image patches, followed by feature embedding and aggregation methods to improve feature discriminativity. Feature matching can be performed by using global features (initial filtering) or use local features to rerank the top-ranked images matched by global features. } 
\label{PipelineofImageRetrieval}
\end{figure*}

\textbf{B) Efficiency related challenges} are important, especially for large-scale datasets \cite{weyand2020google}. Retrieval systems should respond quickly when given a query image. Deep features are high-dimensional and contain semantic-aware information to support higher accuracy, yet is often at the expense of efficiency. 

On the one hand, the efficiency is related to the 
format of features, \ie real valued or binary. Hash codes have advantages in storage and searching  \cite{wang2018survey},\cite{babenko2014neural}, however for hashing methods one needs to carefully consider the loss function design \cite{song2017deep},\cite{lin2018unsupervised}, to obtain optimal codes for high retrieval accuracy. 

On the other hand, efficiency is also related to the mechanism of feature matching. For example, instead of time-consuming cross-matching between local features, one can choose to use global features to perform an initial ranking and then a post-step re-ranking via the features of top-ranked images.

\vspace{-1em}
\section{General Framework of IIR} 
\label{General_Framework_of_IIR}

Figure \ref{PipelineofImageRetrieval} offers an overview of the general framework for deep-learning-based IIR, involving three main stages. %, the first two nicely aligned with the themes of Section~\ref{Summary_of_Progress}:
\medskip
 
\noindent 
\textbf{1) Deep feature extraction}: \hfill (Section \ref{Deep_Feature_Extraction})
\smallskip

\noindent
Feature extraction is the first step of IIR and can be realized in a single-pass or multiple-pass way. Single-pass methods take as input the whole image, whereas multiple-pass methods depend on region extraction, as depicted in Figure \ref{MultiplePass}.

The activations from fully-connected layers of a given DCNN can be used as retrieval features whether based on a whole image or on patches. The tensors from convolutional layers can be used when further processed by sophisticated pooling, as shown in Figure \ref{PipelineofImageRetrieval}. Different layers of the same deep network can be combined as a more powerful extractor % \ie layer-level fusion
\cite{li2016exploiting},\cite{zhou2017collaborative}. Furthermore, it is possible to fuse the activations from layers of different models %\ie model-level fusions
\cite{liu2016deepvehicles},\cite{ozaki2019large}. Feature extraction is the step to produce vanilla network activations (\ie 3D tensors or a single vector), these activations, in most cases, are needed to be further processed. 
\medskip
 
\noindent 
\textbf{2) Embedding and aggregation}: \hfill (Section \ref{Deep_Feature_Enhancement})
\smallskip

\noindent
%Based on the extracted image-level or region-level descriptors, 
Feature embedding and aggregation are two essential steps to produce global or local features. Feature embedding maps individual local features into higher-dimensional space whereas feature aggregation summarizes the multiple mapped vectors or all individual features into a global vector. Global features may come from pooling convolutional feature maps directly \cite{razavian2016visual},\cite{pang2018unifying} or using some sophisticated weighting methods \cite{babenko2015aggregating},\cite{kalantidis2016cross} (\ie both without feature embedding). Feature embedding method using a pre-generated codebook can be performed to encode individual convolutional vectors and then aggregated \cite{sivic2003video},\cite{jegou2010aggregating},\cite{perronnin2007fisher}. For local features, the well-embedded representations for all regions of interest are stored individually and used for cross-matching in the reranking stage without aggregation.
\medskip

\noindent 
\textbf{3) Feature matching}:
\smallskip

\noindent
Feature matching is a process to measure the feature similarity between images and then return a ranked list.  Global matches can be computed efficiently via such as Euclidean distance. For local features \cite{noh2017largescale},\cite{cao2020unifying}, the image similarity is usually evaluated by summarizing the similarities across local features, using classical RANSAC \cite{fischler1981random} or more recent variations \cite{tolias2016image},\cite{teichmann2019detect}. Storing local features separately and then estimating their similarity individually lead to additional memory and search costs \cite{cao2020unifying},\cite{teichmann2019detect}, therefore in most cases local features are used to re-rank the initial ranking image matched by global features \cite{tolias2015particular},\cite{song2017deep},\cite{cao2020unifying},\cite{pang2018improving}.

The three preceding stages for IIR rely on DCNNs as backbone architectures. In almost all cases, pre-stored parameters in these backbones can be fine-tuned (Section \ref{Retrieval_via_Learning_DCNN_Representations}) to be better suited for instance retrieval and to contribute to better performance.  

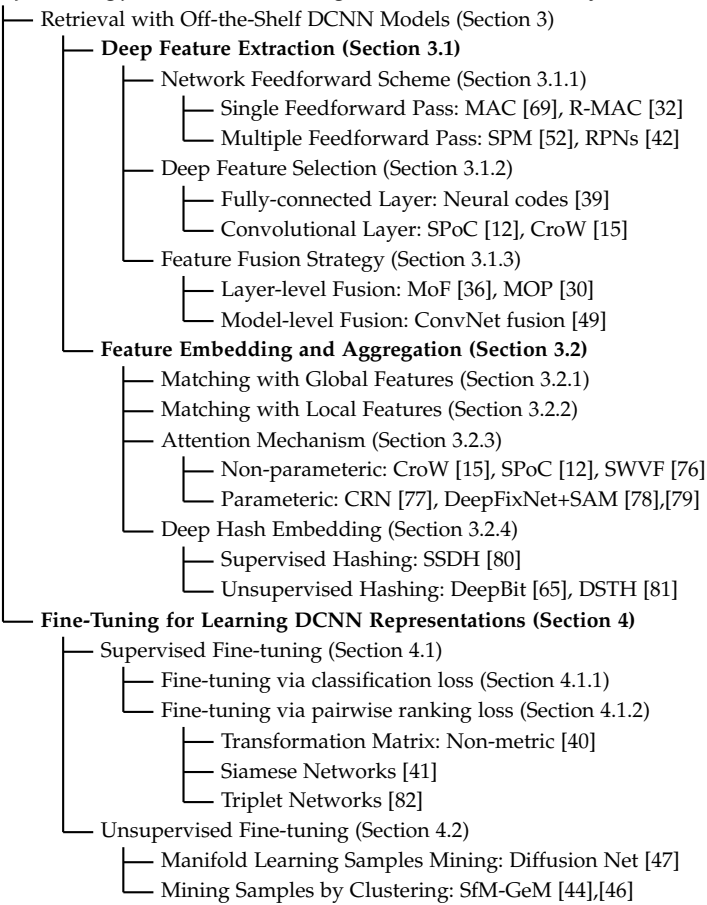
\begin{figure}[!htbp]
\centering
\footnotesize
\begin{tikzpicture}[xscale=0.8, yscale=0.4]

\draw [thick, -] (0, 16.5) -- (0, -4); \node [right] at (-0.5, 17) {\bf \em Deep Learning for Instance-level Image Retrieval (overall survey)};
% \draw [thick, -] (0, 16) -- (0.5, 16);\node [right] at (0.5, 16) {\bf Improvement in Deep Network Architectures (Section \ref{Popular_Backbone_Architectures})};
% \draw [thick, -] (1, 15.5) -- (1, 14);
% \draw [thick, -] (1, 15) -- (1.5, 15);\node [right] at (1.5, 15) {Deepen Networks: AlexNet \cite{krizhevsky2012imagenet}, VGG \cite{simonyan2014very}, ResNet \cite{he2016deep}, \etc};
% \draw [thick, -] (1, 14) -- (1.5, 14);\node [right] at (1.5, 14) {Widen Networks: GoogLeNet \cite{szegedy2015going}, DenseNet \cite{huang2017densely}, \etc };

% %%% category 2

% \draw [thick, -] (0, 13) -- (0.5, 13);\node [right] at (0.5, 13) { General Framework of Instance-level Image Retrieval (Section \ref{General_Framework_of_IIR})};

%%% category 3
\draw [thick, -] (0, 16) -- (0.5, 16);\node [right] at (0.5, 16) {Retrieval with Off-the-Shelf DCNN Models (Section \ref{Retrieval_with_Off_the_Shelf_DCNN_Models})};
\draw [thick, -] (1, 15.5) -- (1, 5);

\draw [thick, -] (1, 15) -- (1.5, 15);\node [right] at (1.5, 15) {\bf Deep Feature Extraction (Section \ref{Deep_Feature_Extraction})};
\draw [thick, -] (2, 14.5) -- (2, 8);

\draw [thick, -] (2, 14) -- (2.5, 14);\node [right] at (2.5, 14) {Network Feedforward Scheme (Section \ref{Network_Feedforward_Scheme})};
\draw [thick, -] (3, 13.5) -- (3, 12); 

\draw [thick, -] (3, 13) -- (3.5, 13);\node [right] at (3.5, 13) {Single Feedforward Pass: MAC \cite{razavian2016visual}, R-MAC \cite{tolias2015particular} };

\draw [thick, -] (3, 12) -- (3.5, 12);\node [right] at (3.5, 12) {Multiple Feedforward Pass: SPM \cite{liu2015deepindex}, RPNs \cite{gordo2016deep} };

%%%
\draw [thick, -] (2, 11) -- (2.5, 11);\node [right] at (2.5, 11) {Deep Feature Selection (Section \ref{Deep_Feature_Selection}) };
\draw [thick, -] (3, 10.5) -- (3, 9); 

\draw [thick, -] (3.0, 10) -- (3.5, 10);\node [right] at (3.5, 10) { Fully-connected Layer: Neural codes \cite{babenko2014neural} };

\draw [thick, -] (3.0, 9) -- (3.5, 9);\node [right] at (3.5, 9) { Convolutional Layer: SPoC \cite{babenko2015aggregating}, CroW \cite{kalantidis2016cross} };

\draw [thick, -] (2.0, 8) -- (2.5, 8);\node [right] at (2.5, 8) {Feature Fusion Strategy (Section \ref{Feature_Fusion_Strategy}) };
\draw [thick, -] (3.0, 7.5) -- (3.0, 6);

\draw [thick, -] (3.0, 7) -- (3.5, 7);\node [right] at (3.5, 7) { Layer-level Fusion: MoF \cite{li2016exploiting}, MOP \cite{gong2014multi} };

\draw [thick, -] (3.0, 6) -- (3.5, 6);\node [right] at (3.5, 6) { Model-level Fusion: ConvNet fusion \cite{simonyan2014very} };

\draw [thick, -] (1, 5) -- (1.5, 5);\node [right] at (1.5, 5) {\bf Feature Embedding and Aggregation (Section \ref{Deep_Feature_Enhancement}) };
\draw [thick, -] (2, 4.5) -- (2, -1); 

\draw [thick, -] (2.0, 4) -- (2.5, 4);\node [right] at (2.5, 4) { Matching with Global Features (Section \ref{Matching_with_Global_Feature})};

\draw [thick, -] (2.0, 3) -- (2.5, 3);\node [right] at (2.5, 3) {Matching with Local Features (Section \ref{Matching_with_Local_Feature})};

\draw [thick, -] (2.0, 2) -- (2.5, 2);\node [right] at (2.5, 2) { Attention Mechanism (Section \ref{Attention_Mechanism}) };
\draw [thick, -] (3.0, 1.5) -- (3.0, 0);

\draw [thick, -] (3.0, 1) -- (3.5, 1);\node [right] at (3.5, 1) { Non-parameteric: CroW \cite{kalantidis2016cross}, SPoC \cite{babenko2015aggregating}, SWVF \cite{qi2017spatial} };

\draw [thick, -] (3.0, 0) -- (3.5, 0);\node [right] at (3.5, 0) { Parameteric: CRN \cite{kim2017learned}, DeepFixNet+SAM \cite{mohedano2017saliency},\cite{yang2017two} };

\draw [thick, -] (2.0, -1) -- (2.5, -1);\node [right] at (2.5, -1) {Deep Hash Embedding (Section \ref{Deep_Hash_Embedding}) };
\draw [thick, -] (3.0, -1.5) -- (3.0, -3);

\draw [thick, -] (3.0, -2) -- (3.5, -2);\node [right] at (3.5, -2) { Supervised Hashing: SSDH \cite{yang2018supervised}  };

\draw [thick, -] (3.0, -3) -- (3.5, -3);\node [right] at (3.5, -3) { Unsupervised Hashing: DeepBit \cite{lin2018unsupervised}, DSTH  \cite{liu2018deep}};

%%% category 4
\draw [thick, -] (0, -4) -- (0.5, -4);\node [right] at (0.5, -4) {\bf Fine-Tuning for Learning DCNN Representations (Section \ref{Retrieval_via_Learning_DCNN_Representations})};
\draw [thick, -] (1, -4.5) -- (1, -11);

\draw [thick, -] (1, -5) -- (1.5, -5);\node [right] at (1.5, -5) {Supervised Fine-tuning (Section \ref{Supervised_Fine-tuning})  };
\draw [thick, -] (2, -5.5) -- (2, -7); 

\draw [thick, -] (2.0, -6) -- (2.5, -6);\node [right] at (2.5, -6) {Fine-tuning via classification loss (Section \ref{Classification-based_Fine-tuning}) }; 

\draw [thick, -] (2.0, -7) -- (2.5, -7);\node [right] at (2.5, -7) {Fine-tuning via pairwise ranking loss (Section \ref{Verification_based_Learning}) };
\draw [thick, -] (3.0, -7.5) -- (3.0, -10);

\draw [thick, -] (3.0, -8) -- (3.5, -8);\node [right] at (3.5, -8) { Transformation Matrix: Non-metric \cite{garcia2017learning} };

\draw [thick, -] (3.0, -9) -- (3.5, -9);\node [right] at (3.5, -9) { Siamese Networks \cite{ong2017siamese}};

\draw [thick, -] (3.0, -10) -- (3.5, -10);\node [right] at (3.5, -10) { Triplet Networks \cite{gordo2017end} };

\draw [thick, -] (1.0, -11) -- (1.5, -11);\node [right] at (1.5, -11) { Unsupervised Fine-tuning (Section \ref{Unsupervised_Fine-tuning})};
\draw [thick, -] (2, -11.5) -- (2, -13); 

\draw [thick, -] (2.0, -12) -- (2.5, -12);\node [right] at (2.5, -12) { Manifold Learning Samples Mining: Diffusion Net \cite{liu2019guided} };

\draw [thick, -] (2.0, -13) -- (2.5, -13);\node [right] at (2.5, -13) {Mining Samples by Clustering: SfM-GeM \cite{radenovic2016cnn},\cite{radenovic2018fine} };

\end{tikzpicture}
\vspace{-2em}
\caption{This survey is organized around three key themes in instance-level image retrieval, shown in bold.}
\label{FourAspectsofprogress}
\end{figure}

The detailed categorization of the material of the following sections is shown in Figure~\ref{FourAspectsofprogress}.

\section{Retrieval with Off-the-Shelf DCNN Models}
\label{Retrieval_with_Off_the_Shelf_DCNN_Models}

\textcolor{black}{Because of their size, DCNNs need to be trained, initially for classification tasks, on exceptionally large-scale datasets and be able to recognize images from different classes.} One possible scheme, then, is that DCNNs effectively trained for classification directly serve as off-the-shelf feature detectors for the image retrieval task, the topic of this survey.  That is, one can propose to undertake image retrieval on the basis of DCNNs, trained for classification and with their pre-trained parameters frozen. 

%%% Four aspects of recent progress
% \medskip

There are limitations to this approach, most fundamentally that there is a model-transfer or domain-shift challenge between tasks~\cite{zheng2018sift},\cite{yosinski2014transferable},\cite{azizpour2016factors}, meaning that models trained for classification do not necessarily extract features well suited to image retrieval. In particular, a classification decision can be successful as long as features remain within classification boundaries, however features from such models may show insufficient capacity for retrieval where feature matching itself is more important than classification. This section will survey the strategies which have been developed to improve the quality of feature representations, particularly based on feature extraction / fusion (Section \ref{Deep_Feature_Extraction}) and feature embedding / aggregation (Section \ref{Deep_Feature_Enhancement}).

\subsection{ Deep Feature Extraction}
\label{Deep_Feature_Extraction}

Feature extraction is about the mechanism by which retrieval features can be extracted from off-the-shelf DCNNs. %, which primarily involves three aspects: network feedforward scheme, feature selection, and feature fusion. 
For an input image $x$ and a network $f(\cdot; \bm{\theta})$, we denote its features from a convolutional layer as $ \boldsymbol{A} := f_{conv}(x) \in \mathbb{R}^{H \times W \times C}$ with height $H$, width $W$, and channels $C$ while that from a fully-connected layer as $ \boldsymbol{B} := f_{fc}(x) \in \mathbb{R}^{D \times 1}$ with the dimensional $D$.

\subsubsection{ Network Feedforward Scheme}
\label{Network_Feedforward_Scheme}

Network feedforward schemes focus on how images are fed into a DCNN, which includes single-pass and multiple-pass.

\begin{flushleft}
\emph{a. Single Feedforward Pass Methods}.   
\end{flushleft}

Single feedforward pass methods take the whole image and feed it into an off-the-shelf model to extract features. The approach is relatively efficient since the input image is fed only once. For these methods, both the fully-connected layer and last convolutional layer can be used as feature extractors \cite{mohedano2016bags}.

Early network-based IIR work focused on leveraging DCNNs as a fixed extractor to obtain global features, especially based on the fully-connected layers \cite{sharif2014cnn},\cite{babenko2014neural}, requiring close to zero engineering effort. However, extracting features in this way affects retrieval accuracy since the extracted features may include background information or activations for irrelevant objects.

The key to single-pass schemes is to embed and aggregate features to improve their discriminativity, such that features of two related images (\ie including the same object) are more similar than these of two unrelated images \cite{babenko2015aggregating}. For this purpose, it is possible to first map the features $ \boldsymbol{B} $  into a high-dimensional space and then to aggregate them into a final global feature \cite{gong2014multi}. Another direction is to treat regions in convolutional features $ \boldsymbol{A} $ as different sub-vectors, such that a combination of sub-vectors of all feature maps are used to represent the input image \cite{kalantidis2016cross},\cite{tolias2015particular}.

\begin{flushleft}
\emph{b. Multiple Feedforward Pass Methods}.   
\end{flushleft}

Compared to single-pass schemes, multiple-pass methods are more time-consuming \cite{zheng2018sift} because several patches are generated and then fed into the network, % before being aggregated as a final global feature. However, 
\textcolor{black}{
multiple-pass schemes are more helpful for addressing the ``\textit{invariance challenges} and ``\textit{distraction challenges}'' in Section \ref{Keychallenges}. Local patches at multiple scales become more robust for image translation, scaling and rotation \cite{gong2014multi},\cite{ reddy2015object}. Also, these patches are helpful to filter several irrelevant background information. 
}

\begin {figure}[!t]
\centering
 {    
   \includegraphics[width=\columnwidth]{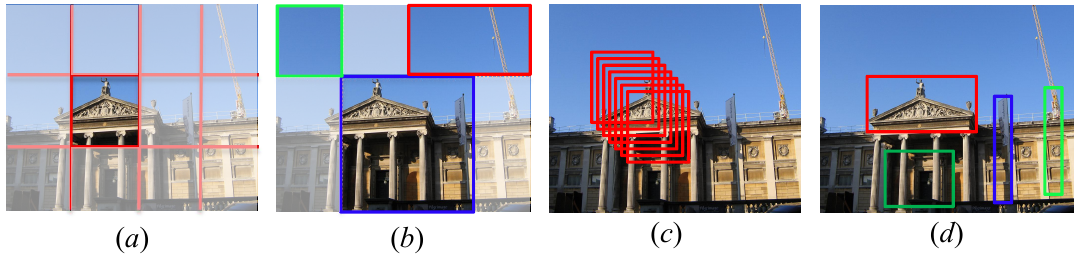} 
 }
 \vspace{-2em}
\caption{Image patch generation schemes: (a) Sliding windows \cite{sharif2015baseline},\cite{do2017embedding}; (b) Spatial pyramid modeling \cite{zhao2017spatial}; (c) Dense sampling \cite{gong2014multi},\cite{li2016exploiting}; (d) Region proposals from region proposal networks \cite{ren2015faster},\cite{gordo2016deep}.}
\vspace{-1em}
\label{MultiplePass}
\end {figure}

The representations are usually produced from two stages: patch detection and patch description. Multi-scale image patches are obtained using sliding windows \cite{sharif2015baseline},\cite{do2017embedding} or spatial pyramid model (SPM) \cite{liu2015deepindex},\cite{zhao2017spatial},\cite{zheng2016accurate}, as shown in Figure \ref{MultiplePass}. For example, Zheng \etal \cite{zheng2016accurate} partition an image by using SPM and extract features at increasing scales, thus enabling the integration of global, regional, local contextual information.

Patch detection methods lack retrieval efficiency since irrelevant patches are also detected \cite{tolias2015particular}. For example, Cao \etal \cite{cao2016focus} propose to merge image patches into larger regions with different hyper-parameters, where the hyper-parameter selection is viewed as an optimization problem to maximize the similarity between query and candidate features.

Instead of generating multi-scale image patches randomly or densely, region proposal methods introduce a degree of purpose. Region proposals can be generated using object detectors, such as selective search \cite{reddy2015object}, edge boxes \cite{zitnick2014edge},\cite{yu2017efficient}, and BING \cite{sun2015scalable}. For example, Yu \etal \cite{yu2017efficient} propose fuzzy object matching (FOM) for instance search in which the fuzzy objects are generated from 300 object proposals and then clustered to filter out overlapping proposals. Region proposals can also be learned using such as region proposal networks (RPNs) \cite{ren2015faster},\cite{gordo2016deep} and convolutional kernel networks (CKNs) \cite{mairal2014convolutional}, and then to apply these networks into end-to-end fine-tuning for learning similarity \cite{salvador2016faster}. \textcolor{black}{This usually requires the datasets provide well-localized bounding boxes as supervision, \eg the datasets INSTRE \cite{wang2015instre}, Oxford-5k \cite{philbin2007object}, Paris-6k \cite{Lostphilbin2008lost1}, GLD-v2 variant \cite{teichmann2019detect}. Also, in the off-the-shelf scenarios, the way that using the bounding boxes to crop the query images and use as input the DCNNs has been shown to provide better retrieval performance since only the information relevant to the instance is extracted \cite{salvador2016faster},\cite{ tolias2015particular}.
}

\subsubsection{Deep Feature Selection}
\label{Deep_Feature_Selection}

Feature selection decides the receptive field of the extracted features, \ie global-level from fully-connected layers and regional-level from convolutional layers.

\begin{flushleft}
\emph{a. Extracted from Fully-connected Layers}
\end{flushleft}

It is straightforward to select a fully-connected layer as a global feature extractor \cite{sharif2014cnn},\cite{gong2014multi},\cite{babenko2014neural}. With PCA dimensionality reduction and normalization~\cite{sharif2014cnn} image similarity can be measured. Extracting features $ \boldsymbol{B} $ from fully-connected layer leads to two obvious limitations for IIR: including irrelevant information, and a lack of local geometric invariance \cite{gong2014multi}.

With regards to the first limitation, image-level global descriptors may include irrelevant patterns or background clutter, especially when a target instance is only a small portion of an image. It may then be more reasonable to extract region-level features at finer scales, \ie using multiple passes~\cite{gong2014multi},\cite{reddy2015object},\cite{song2017deep}.
For the second limitation, %a lack of local geometric invariance affects the robustness to image transformations, 
an alternative is to extract multi-scale features on a convolutional layer \cite{razavian2016visual},\cite{morere2017nested}. Further, it makes the global features incompatible with techniques such as spatial verification and re-ranking. Several methods then choose to leverage intermediate convolutional layers \cite{babenko2015aggregating},\cite{gong2014multi},\cite{yue2015exploiting},\cite{razavian2016visual}.

\begin{flushleft}
\emph{b. Extracted from Convolutional Layers}
\end{flushleft}

The neurons in a convolutional layer are connected only to a local region of the input image, and this smaller receptive field ensures that the produced features $ \boldsymbol{A} $, usually from the last layer, preserve more local structural information \cite{ng2020solar},\cite{zheng2016good} and are more robust to image transformations \cite{babenko2015aggregating} \textcolor{black}{ thereby address the ``\textit{invariance challenge}''. For instance, Razavian \etal \cite{razavian2016visual} extract multi-scale features on the last convolutional layer and Mor{\`e}re \etal \cite{morere2017nested} incorporate a series of nested pooling layers into CNN. Both of them provide higher feature invariance. Thus, many image retrieval methods use convolutional layers as feature extractors~ \cite{jimenez2017class},\cite{yue2015exploiting},\cite{razavian2016visual},\cite{lou2018multi}.}

Sum/average and max pooling are two simple aggregation methods to produce global features \cite{razavian2016visual}. For a pooled layer, the last convolutional layer usually yields superior accuracy over other shallower or later fully-connected layers \cite{zheng2016good}. There is no other operation on the feature maps before pooling, so we illustrate these methods as ``direct pooling'' in Figure~\ref{PipelineofImageRetrieval}.

%%%% figure
\begin {figure}[!t]
\centering
 {  
   \includegraphics[width=\columnwidth]{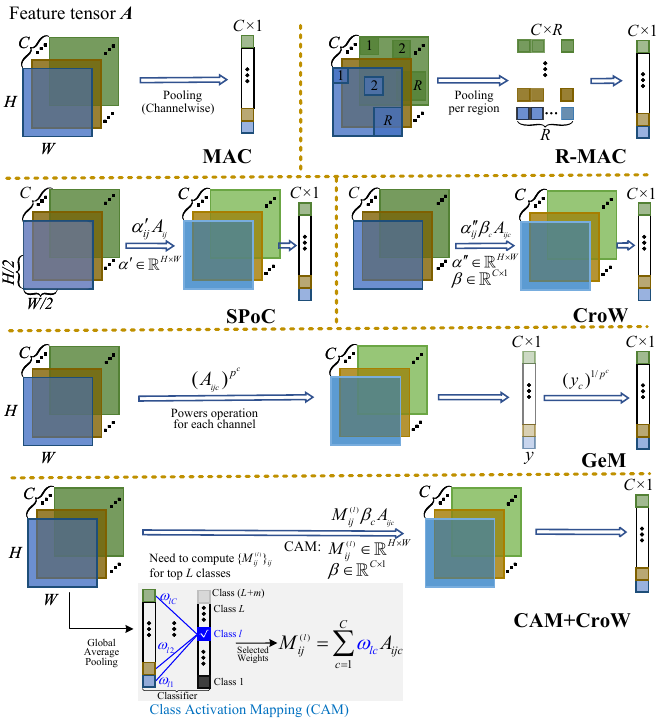} 
 }
 \vspace{-1.5em}
\caption{Representative methods in single-pass methods, focusing on convolutional feature tensor $ \boldsymbol{A} $. We denote the entry in $ \boldsymbol{A} $ corresponding to channel $ c $, at spatial location ($i, j$) as $ A_{ijc} $:
MAC \cite{razavian2016visual}, R-MAC \cite{tolias2015particular}, SPoC with the per-channel Gaussian weighting $\alpha^{\prime}_{ij} A_{ij}$ where $\alpha^{\prime}_{ij} = \exp\left\{-{\textstyle\frac{\left(i-\frac H2\right)^2+\left(j-\frac W2\right)^2}{2\sigma^2}}\right\} $ \cite{babenko2015aggregating}, CroW with $ \alpha^{\prime\prime} $ computed by summing all $C$ feature maps at location ($i, j$)  and $ \beta $ computed by summing the $H\times W$-array at each feature map $c$ \cite{kalantidis2016cross}, GeM with channel-wise powers operation \cite{radenovic2018fine}, and CAM+CroW by performing $ M_{ij}^{(l)}=\sum_{c=1}^C\omega_{lc}A_{ijc} $ where $\omega_{lc}$ are weights activated by $l$-th class \cite{jimenez2017class}. 
}
\vspace{-1em}
\label{SinglePass}
\end {figure}

Instead of direct pooling, many sophisticated aggregation methods have been explored, such as channel-wise or spatial-wise feature weighting on the convolutional feature maps \cite{morere2017nested},\cite{li2017ms},\cite{xiang2019multiple}. These aggregation methods aim to highlight feature importance \cite{kalantidis2016cross} or reduce the undesirable influence of bursty descriptors of some regions \cite{do2018selective},\cite{pang2018deep}. For clarity, we illustrate the representative strategies in Figure \ref{SinglePass}. %, including  MAC \cite{razavian2016visual}, R-MAC \cite{tolias2015particular}, GeM pooling \cite{radenovic2018fine}, SPoC \cite{babenko2015aggregating}, CroW \cite{kalantidis2016cross}, and CAM+CroW \cite{jimenez2017class}. 
Note that these feature aggregation methods are usually performed before channel-wise sum/max pooling and does not embed features into a higher dimensional space.

One rationale behind using convolutional features is that each such vector can act as a ``dense SIFT'' feature \cite{babenko2015aggregating} since each vector corresponds to a region in the input image. Inspired by this perception, many works leverage embedding methods (\eg BoW) used for SIFT features \cite{lowe2004distinctive} on the regional feature vectors and then aggregate them (\eg by sum pooling) into a global descriptor. 
\textcolor{black}{Feature embedding methods address the discriminativity challenge via mapping individual features into a high-dimensional space and make them distinguishable \cite{do2018selective}.} Feature embedding is followed by PCA to reduce feature dimensionality and whitening to down-weight co-occurrence between features.

\subsubsection{Feature Fusion Strategies}
\label{Feature_Fusion_Strategy}

Fusion studies the complementarity of different features which includes layer-level and model-level fusion explorations.

\begin{flushleft}
\emph{a. Layer-level Fusion}
\end{flushleft}

With layer-level fusion it is possible to fuse multiple fully-connected layers in a deep network \cite{liu2015deepindex},\cite{zhou2017collaborative}. For instance, Liu \etal \cite{liu2015deepindex} introduce DeepIndex to incorporate multiple global features from different fully connected layers. The activation from the first fully-connected layer is taken as column indexing, and that from the second layer serves as row indexing.  Similarly, it is also possible to fuse the activations from multiple convolutional layers. For instance, Li \etal \cite{li2017ms} apply the R-MAC encoding scheme on five convolutional layers of VGG-16 and then concatenate them into a multi-scale feature vector.

Features from fully-connected layers retain global high-level semantics, whereas features from convolutional layers can present local low- and intermediate-level cues. Global and local features therefore complement each other when measuring semantic similarity and can, to some extent, guarantee retrieval performance \cite{yu2017exploiting},\cite{yang2021dolg}. Such features can be concatenated directly \cite{cao2020unifying},\cite{yu2017exploiting}, with  convolutional features normally filtered by sliding windows or region proposal nets. Direct concatenation can also be replaced by other advanced methods, such as orthogonal operations \cite{yang2021dolg} or pooling-based methods, such as Multi-layer Orderless Fusion (MOF) of Li \etal \cite{li2016exploiting}, which is inspired by Multi-layer Orderless Pooling (MOP) \cite{gong2014multi}. However local features cannot play a decisive role in distinguishing subtle feature differences if global and local features are treated identically. Yu \etal \cite{yu2017exploiting} use a mapping function to assert local features in refining the return ranking lists, via an exponential mapping function for tapping the complementary strengths of convolutional and fully-connected layers. Similarly, % Cao \etal \cite{cao2020unifying} unify the global and local descriptors for two-stage image retrieval in which attentively selected local features are used to refine the results obtained using global features. 
Liu \etal \cite{liu2019group} design two sub-networks on top of convolutional layers to obtain global and local features and then learn to fuse these features, thereby adaptively adjusting the fusion weights. Instead of directly fusing the layer activations, Zhang \etal \cite{zhang2019effective} fuse the index matrices which are generated based on the two feature types extracted from the same CNN, a feature fusion which has low computational complexity.

It is worth considering which layer combinations are better for fusion given their differences and complementarity. Yu \etal \cite{yu2017exploiting} compare the performance of different combinations between fully-connected and convolutional layers on the Oxford 5k, Holiday, and UKBench datasets. The results show that the combinations including the first fully-connected layer always perform better. Li \etal \cite{li2016exploiting} demonstrate that fusing convolutional and fully-connected layers outperforms the fusion of only convolutional layers. Fusing two convolutional layers with one fully-connected layer achieves the best performance on the Holiday and UKBench datasets.

\begin{flushleft}
\emph{b. Model-level Fusion}
\end{flushleft}

It is possible to combine features from different models; such a fusion focuses more on model complementarity, with methods categorized into \emph{intra-model} and \emph{inter-model}.

Intra-model fusion suggests multiple deep models having similar or highly compatible structures, while inter-model fusion involves models with differing structures. For example, %the widely-used dropout strategy in AlexNet \cite{krizhevsky2012imagenet} can be regarded as intra-model fusion: with random connections of different neurons between two fully-connected layers, each training epoch can be viewed as the combinations of different models. As a second example, 
Simonyan \etal \cite{simonyan2014very} introduce a ConvNet intra-model fusion strategy to improve the feature learning capacity of VGG where VGG-16 and VGG-19 are fused.  
%Similarly, Liu \etal \cite{liu2016deepvehicles} mix different VGG variants to strengthen the learning for fine-grained vehicle retrieval. %Ding \etal \cite{ding2018selective} propose a selective deep ensemble framework to combine ResNet-26 and ResNet-50 improve the accuracy of fine-grained instance retrieval.
To attend to different parts of an image object, Wang \etal \cite{wang2019improving} realize the multi-feature fusion by selecting all convolutional layers of VGG-16 to extract image representations, which is demonstrated to be more robust than using only single-layer features.

Inter-model fusion is a way to bridge different features given the fact that different deep networks have different receptive fields \cite{liu2015deepindex},\cite{ozaki2019large},\cite{yang2017two},\cite{zheng2016good}. For instance, a two-stream attention network \cite{yang2017two} is introduced to implement image retrieval where the main network for semantic prediction is VGG-16 while an auxiliary network is used for predicting attention maps. Similarly, considering the importance and necessity of inter-model fusion to bridge the gap between mid-level and high-level features, Liu \etal \cite{liu2015deepindex} and Zheng \etal \cite{zheng2016good} combine VGG-19 and AlexNet to learn combined features, while Ozaki \etal \cite{ozaki2019large} concatenate descriptors from six different models.

Inter-model and intra-model fusion are relevant to model selection. There are some strategies to determine how to combine the features from two models. It is straightforward to fuse all features from the candidate models and then learn a metric based on the concatenated features \cite{liu2015deepindex},\cite{yang2017two}, which is a kind of ``\emph{early fusion}'' strategy. Alternatively, it is also possible to learn optimal metrics separately for the features from each model, and then to combine these metrics for final retrieval ranking \cite{li2016exploiting},\cite{Zheng2015QueryadaptiveLF}, which is a kind of ``\emph{late fusion}'' strategy.

\textbf{Discussion.} Layer-level fusion and model-level fusion are conditioned on the fact that the associated layers or networks have different feature description capacities. For these fusion strategies, the key question is \textit{what features are the best to be combined?} Some explorations have been made on the basis of off-the-shelf models, such as Xuan \etal \cite{xuan2018deep}, who illustrates the effect of combining different numbers of features and different sizes within the ensemble. Chen \etal \cite{chen2019analysis} analyze the performance of embedded features from off-the-shelf image classification and object detection models with respect to image retrieval.

\subsection{Feature Embedding and Aggregation} 
\label{Deep_Feature_Enhancement}

\textcolor{black}{The primary aim of feature embedding and aggregation is to further promote feature discriminativity, targeting for the ``\textit{discriminativity challenge}'', and obtain final global and/or local features for retrieving specific instances.}

\subsubsection{Matching with Global Features}
\label{Matching_with_Global_Feature}

Global features can be extracted from fully-connected layers, followed by dimensionality reduction and normalization \cite{sharif2014cnn},\cite{babenko2014neural}. They are easy to implement and there is no further aggregation process. Gong \etal \cite{gong2014multi} extract fully-connected activations for local image patches at three scale levels and embed patch-level activations individually using VLAD. Thus, the final concatenated features significantly tackle the invariance challenge caused by image rotations.

Convolutional features can also be aggregated into compact a global feature. Simple aggregation methods are sum/average or max pooling \cite{razavian2016visual},\cite{pang2018unifying}. %For instance, Razavian \etal \cite{razavian2016visual} make the first attempt to perform spatial max pooling on the feature maps of an off-the-shelf DCNN model; 
Sum/average pooling is less discriminative, because it takes into account all activated convolutional outputs, thereby weakening the effect of highly activated features \cite{do2018selective}. As a result, max pooling is particularly well suited for sparse features having a low probability of being active, however max pooling may be inferior to sum/average pooling when image features are whitened \cite{babenko2015aggregating}.

Figure \ref{SinglePass} illustrates sophisticated feature aggregation methods using channel-wise or spatial-wise weighting \cite{babenko2015aggregating},\cite{morere2017nested}. For example, Babenko \etal \cite{babenko2015aggregating} propose sum-pooling convolutional features (SPoC) to obtain compact descriptors weighted by $ \alpha^{\prime} $ with a Gaussian center prior. Similarly, %Ng \etal \cite{ng2020solar} explore the correlations between activations at different locations on the feature maps, thus improving the final descriptors. Also, 
it is possible to treat regions in feature maps as different sub-vectors \cite{tolias2015particular},\cite{razavian2016visual},\cite{zheng2016good}, thus combinations of $R$ sub-vectors are used to represent the input image, such as R-MAC \cite{tolias2015particular}. Since convolutional features may include repetitive patterns and each vector may correspond to identical regions, the resulting descriptors may be bursty, which makes the final aggregated global feature less distinguishable. As a solution, Pang \etal \cite{pang2018deep} leverage heat diffusion to weigh convolutional features at the aggregation stage, and reduce the undesirable influence of burstiness.

Convolutional features have an interpretation as descriptors of local regions, thus many works leverage embedding methods, including BoW, VLAD, and FV, to encode regional feature vectors and then aggregate them into a global descriptor.  Note that BoW and VLAD can be extended by using other metrics, such as a Hamming distance \cite{wang2014hamming}. Here we briefly describe the principle of Euclidean embeddings.

BoW \cite{sivic2003video} is a widely used feature embedding which leads to a sparse vector of occurrence. Let $ \boldsymbol{a} = \left\{a_{1}, a_{2},...,a_{R} \right\} $ be a set of $R$ local features, each of dimensionality $d$. BoW requires a pre-defined codebook $ \boldsymbol{c} = \left\{c_{1}, c_{2},...,c_{K} \right\} $ with $ K $ centroids, usually learned offline, to cluster these local descriptors, and maps each descriptor $a_{t}$ to the nearest centroid $c_{k}$. For each centroid, one can count and normalize the number of occurrences as
\begin{equation}
g(c_{k}) = \frac{1}{R} \sum_{r=1}^{R}\phi(a_{r}, c_{k})
\end{equation}
\begin{equation}
\phi(a_{r}, c_{k}) = \left\{ \begin{array}{ll}

1 & \textrm{if $ c_{k} $ is the closest codeword for $a_{r}$ }\\

0 & \textrm{otherwise}\\

\end{array} \right.
\label{eqn-phi}
\end{equation}
Thus BoW considers the number of descriptors belonging to each $c_{k}$ (\emph{i.e.} 0-order feature statistics), so the BoW representation is the concatenation of all mapped vectors:
\begin{equation}
\label{BoW}
 G_{_{BoW}}(\boldsymbol{a}) =
   \left[
   \begin{array}{ccc}
   g(c_{1}), \cdots ,g(c_{k}) ,\cdots, g(c_{K})
   \end{array}
   \right] \rm ^{\top}
\end{equation}
BoW is simple to implement the encoding of local descriptors, such as convolutional feature maps \cite{li2016exploiting},\cite{mohedano2016bags} or fully-connected activations \cite{zheng2016accurate},\cite{zhang2019effective}, or to encode regional descriptors \cite{mishchuk2017working},\cite{mukherjee2020bag}. Mukherjee \etal \cite{mukherjee2020bag} extract image patches based on information entropy and feed into a pre-trained VGG-16, then use BoW to embed and aggregate the patch-level descriptors from a fully-connected layer.  Embedded BoW vectors are typically high-dimensional and sparse, so not well suited to large-scale datasets in terms of the mentioned efficiency challenge.

VLAD \cite{jegou2010aggregating} stores the sum of residuals for each visual word.  Similar to BoW, it generates $K$ visual word centroids, then each feature $a_{r}$ is assigned to its nearest visual centroid $ c_{k} $:
\begin{equation}
g(c_{k}) = \frac{1}{R} \sum_{r=1}^{R}\phi(a_{r} , c_{k})(a_{r} - c_{k})
\end{equation}
The VLAD representation is stacked by the residuals for all centroids, with dimension ($ d \times K $), \ie 
\begin{equation}
G_{_{VLAD}}(\boldsymbol{a}) \! = \! \left[
   \begin{array}{ccc}
   \cdots, g(c_{k}) \rm ^{\top},\cdots 
   \end{array}
   \right] \rm ^{\top} .
\end{equation}
VLAD captures first-order feature statistics, \ie ($a_{r} - c_{k}$). Similar to BoW, the performance of VLAD is affected by the number of clusters: more centroids produce larger vectors that are harder to index. For instance-level image retrieval, 
Gong \etal \cite{gong2014multi} concatenate the activations of a fully-connected layer with VLAD applied to image-level and patch-level inputs \cite{paulin2015local}. Ng \etal \cite{yue2015exploiting} replace BoW~\cite{sivic2003video} with VLAD~\cite{jegou2010aggregating}, and are the first to encode local features into VLAD representations. This idea inspired another milestone work \cite{arandjelovic2016netvlad} where, for the first time, VLAD is plugged into the last convolutional layer, which allows  end-to-end training via back-propagation.

FV \cite{perronnin2007fisher} extends BoW by encoding the first and second order statistics. FV clusters the set of local descriptors by a Gaussian Mixture Model (GMM) with $K$ components to generate a dictionary $ \boldsymbol{c} = \left\{ \mu_{k}; \Sigma_{k}; w_{k}\right\}_{k=1}^{K} $ made up of mean / covariance / weight triples \cite{jegou2012aggregating}, where the covariance may be simplified by keeping only its diagonal elements. For each local feature  $ a_{r} $, a GMM is given by 
\begin{equation}
\begin{aligned}
\gamma_{k}(a_{r})  =  w_{k}\times p_{k}(a_{r})/\Big(\sum_{k=1}^{K}w_{k}p_{k}(a_{r})\Big) \quad s.t. \sum_{k=1}^K w_{k} = 1
\end{aligned}
\end{equation}
where $ p_{k}(a_{r})=\mathcal{N}(a_{r}, \mu_{k},  \sigma_{k}^2 )$. All local features are assigned into each component $k$ in the dictionary, which is computed as
\begin{equation}
\begin{aligned}
 &g_{w_{k}} =  \frac{1}{R \sqrt{w_{k}}} \sum_{r=1}^{R} \Big(\gamma_{k}(a_{r}) - w_{k}\Big) \\
& g_{u_{k}}  = \frac{\gamma_{k}(a_{r})}{R \sqrt{w_{k}}} \sum_{r=1}^{R} \left( \frac{a_{r} - \mu_{k}}{\sigma_{k}} \right),\\ 
& g_{\sigma_{k}^2} = \frac{\gamma_{k}(a_{r})}{R \sqrt{2w_{k}}} \sum_{r=1}^{R}\left[ {\left( \frac{a_{r} - \mu_{k}}{\sigma_{k}} \right)}^{2} - 1 \right]
\end{aligned}
\end{equation}
The FV representation is produced by concatenating vectors from the $K$ components: 
\begin{equation}
\begin{aligned}
\!\!\! G_{_{FV}}(\boldsymbol{a}) \! = \! \left[
   \begin{array}{ccc}
   g_{w_{1}}, \cdots , g_{w_{K}}, g_{u_{1}}, \cdots, g_{u_{K}}, g_{\sigma_{1}^2}, \cdots, g_{\sigma_{K}^2} 
   \end{array}
   \right] \rm ^{\top}
\end{aligned}
\end{equation}
The FV representation defines a kernel from a generative process and captures more statistics than BoW and VLAD. FV vectors do not increase the computational cost significantly but require more memory.  Applying FV without memory controls may lead to suboptimal performance \cite{sanchez2013image}.

\textbf{Discussion.} Traditionally, pooling-based aggregation methods (\eg in Figure \ref{SinglePass}) are directly plugged into deep networks and then the whole model is used  end-to-end. The three embedding methods (BoW, VLAD, FV) are initially trained with large pre-defined vocabularies \cite{liu2015deepindex},\cite{cao2017local}. One needs to pay attention on their properties before choosing an embedding: BoW and VLAD are computed in the rigid Euclidean space where performance is closely related to the number of centroids, whereas FV can capture higher-order statistics and improves the effectiveness of feature embedding at the expense of a higher memory cost. Further, although vocabularies are usually built separately and pre-trained before encoding deep features, it is necessary to integrate the training of networks and the learning of vocabulary parameters into a unified framework so as to guarantee training and testing efficiency. For example, VLAD is integrated into deep networks where each spatial column feature is used to construct clusters via k-means \cite{yue2015exploiting}. This idea led to NetVLAD \cite{arandjelovic2016netvlad}, where deep networks are fine-tuned with the VLAD vectors. The FV method is also combined with deep networks for retrieval tasks \cite{do2018selective},\cite{ong2017siamese}.

\subsubsection{Matching with Local Features}
\label{Matching_with_Local_Feature}

Although matching with global features has high efficiency for both feature extraction and similarity computation, global features are not compatible with spatial verification and correspondence estimation, which are important procedures for instance-level retrieval tasks, motivating work on matching with local features. In terms of the matching process, global features are matched only once while local feature matching is evaluated by summarizing the similarity across all individual local features (\ie many-to-many matching).

One important aspect of local features is to detect the keypoints for an instance within an image, and then to describe the detected keypoints as a set of local descriptors. Inspired by \cite{dusmanu2019d2}, the common strategies of this whole procedure for IIR can be categorized as \textit{detect-then-describe} and \textit{describe-then-detect}.

In terms of \textit{detect-then-describe}, we regard the descriptors around keypoints as local features, similar to \cite{sharif2014cnn},\cite{gordo2016deep}. Coarse regions can be detected, for example, by using the methods depicted in Figure~\ref{MultiplePass}, and regions of interest in an image can be detected by using region proposal networks (RPNs) \cite{ren2015faster},\cite{teichmann2019detect}. The extracted coarse regions around the keypoints are fed into a DCNN, followed by feature description.  Traditional detectors can also be used to detect fine regions around a keypoint. For instance, Zheng \etal \cite{zheng2016accurate} employ the popular Hessian-Affine detector \cite{mikolajczyk2004scale} to get an affine-invariant local region. Paulin \etal \cite{paulin2015local} and Mishchuk \etal \cite{mishchuk2017working} detect regions using the Hessian-Affine detector and feed into patch-convolutional kernel networks (Patch-CKNs) \cite{mairal2014convolutional}. \textcolor{black}{Note that it becomes more convenient for the case where bounding boxes annotations have been provided by datasets (see Section \ref{Datasets_and_Evaluation_Criteria}), and then the image regions can be cropped directly for further reranking \cite{salvador2016faster}. }

Rather than performing keypoint detection early on, it is possible to postpone the detection stage on the convolutional feature maps, \ie \textit{describe-then-detect}. One can select regions on the convolutional feature maps to obtain a set of local features \cite{tolias2015particular},\cite{sharif2015baseline},\cite{mohedano2016bags}; the local maxima of the feature maps are then detected as keypoints \cite{simeoni2019local}. A similar strategy is also used in network fine-tuning \cite{noh2017largescale},\cite{tan2021instance},\cite{cao2020unifying},\cite{teichmann2019detect},\cite{tolias2020learning}, where the keypoints on the convolutional feature maps can be selected based on attention scores predicted by an attention network \cite{cao2020unifying},\cite{noh2017largescale}, or based on single-head and multi-head attention modules in transformers \cite{tan2021instance},\cite{el2021training}. This approach to keypoint selection is better for achieving computational efficiency. 

After keypoint detection and description, a large number of local features are used in the matching stage to perform instance-level retrieval, and the image similarity is evaluated by matching across all local features. Local matching techniques include spatial verification and selective match kernels (SMK) \cite{tolias2016image}. Spatial verification assumes object instances are rigid so that local matches between images can be estimated as an affine transformation using RANdom SAmple Consensus (RANSAC) \cite{fischler1981random}. One limitation of RANSAC is its high computational complexity of estimating the transformation model when all local descriptors are considered; instead, it is possible to apply RANSAC to a small number of top-ranked local descriptors, such as those selected by approximate nearest neighbor \cite{noh2017largescale}. 
SMK weighs the contributions of individual matches with a non-linear selective function, but is still memory intensive. Its extension, the Aggregated Selective Match Kernel (ASMK), focuses more on aggregating similarities between local features without explicitly modeling the geometric alignment, which can produce a more compact representation \cite{tolias2016image},\cite{tolias2020learning}. Recently, Teichmann \etal \cite{teichmann2019detect} introduced Regional Aggregated Selective Match Kernel (R-ASMK) to combine information from detected regions, boosting image retrieval accuracy compared to the ASMK.

\textbf{Discussion.} Using local descriptors to perform instance retrieval tasks has two limitations. First, the local descriptors for an image are stored individually and independently, which is memory-intensive, and not well-suited for large-scale scenarios.
Second, estimating the similarity between the query and database images depends on cross-matching all local descriptor pairs, which incurs additional searching cost and then a low retrieval efficiency. Therefore, most instance retrieval systems using local features follow a two-stage paradigm: initial filtering and re-ranking \cite{song2017deep},\cite{cao2020unifying},\cite{pang2018improving},\cite{salvador2016faster},\cite{yang2018dynamic}, as in Figure \ref{PipelineofImageRetrieval}. The initial filtering stage is to employ a global descriptor to select a set of candidate matching images, thereby reducing the solution space; the re-ranking stage is to use local descriptors to re-rank the top-ranked images from the global descriptor.

\subsubsection{Attention Mechanism}
\label{Attention_Mechanism}

\begin{figure}[!t]
\centering
\includegraphics[width=0.5\textwidth]{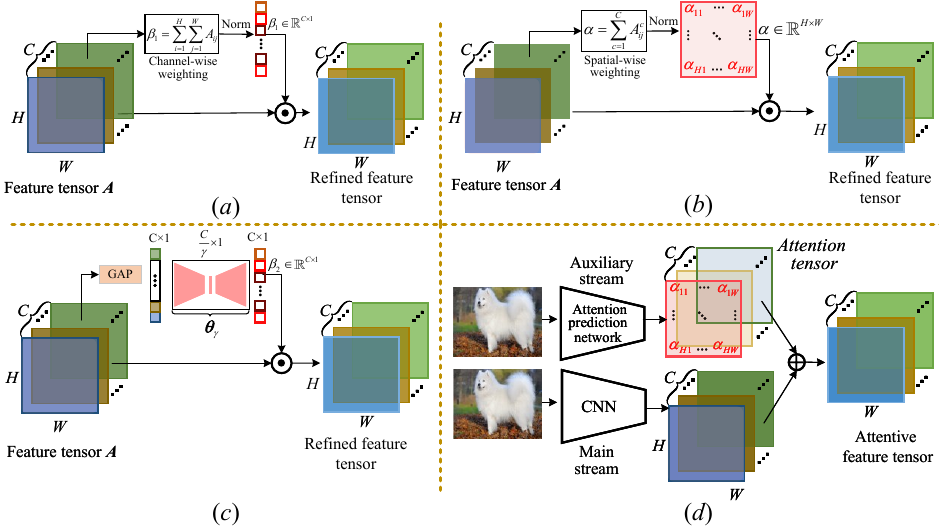}
\vspace{-1.5em}
\caption{Illustration of attention mechanisms. (a)-(b) Non-parametric schemes:  The attention is based on convolutional feature $ \boldsymbol{A} $. Channel-wise attention in (a) produces a $C$-dimensional importance vector \textbf{$\beta_{1}$} \cite{kalantidis2016cross},\cite{xu2018unsupervised}; Spatial-wise attention in (b) computes a 2-dimensional attention map \textbf{$\alpha$} \cite{kalantidis2016cross},\cite{jimenez2017class},\cite{ng2020solar}. (c)-(d) Parametric schemes: The attention weights are learned by a trainable network. In (c), \textbf{$\beta_2$} are provided by a sub-network with parameters $\theta_{\gamma}$ \cite{cao2020unifying},\cite{kim2017learned},\cite{lou2018multi},\cite{kimregional2018Regional}. In (d), the attention maps, as a tensor, are predicted by some auxiliary saliency extraction models from the input image directly \cite{mohedano2017saliency},\cite{yang2017two},\cite{wei2019saliency}.} \label{fig3}
\end{figure}

Attention mechanism can be regarded as a kind of feature aggregation, whose aim is to \textcolor{black}{highlight the most relevant feature parts. It can effectively address the ``\textit{distraction challenge}'' and also promote feature discriminativity \cite{lou2018multi}, realized by computing an attention map.} Approaches to obtaining attention maps can be categorized into non-parametric and parametric groups, as shown in Figure~\ref{fig3}, where the main difference is whether the importance weights in the attention map are learnable.

Non-parametric weighting is a straightforward method to highlight feature importance, and the corresponding attention maps can be obtained by channel-wise or spatial-wise pooling, as in Figure \ref{fig3} (a,b). For spatial-wise pooling, 
%Babenko \etal \cite{babenko2015aggregating} apply a Gaussian center prior scheme to spatially weight the activations of a convolutional layer prior to aggregation. 
Kalantidis \etal \cite{kalantidis2016cross} propose an effective CroW method to weight and pool feature maps, which concentrate on weighting activations at different spatial locations, without considering the relations between these activations. In contrast, Ng \etal \cite{ng2020solar} explore the correlations among activations at different spatial locations on the convolutional feature maps. 

Channel-wise weighting methods are also popular non-parametric attention mechanisms \cite{xu2018unsupervised},\cite{xiang2019multiple}. Xu \etal \cite{xu2018unsupervised} rank the weighted feature maps to build ``probabilistic proposals'' to select regional features. Jimenez \etal \cite{jimenez2017class} combine CroW and R-MAC to propose Classes Activation Maps (CAM) to weigh the feature map per class. Xiang \etal \cite{xiang2019multiple} employ a Gram matrix to analyze the correlations between different channels and then obtain channel sensitivity information to tune the importance of each feature map. Channel-wise and spatial-wise weighting methods are usually integrated into a deep model to highlight feature importance \cite{kalantidis2016cross},\cite{wang2019improving}.

Parametric attention maps, shown in Figure \ref{fig3} (c,d), can be learned via deep networks, where the input can be either image patches or feature maps \cite{kim2017learned},\cite{xiang2019multiple},\cite{yang2021dolg}, approaches which are commonly used in supervised metric learning \cite{lou2018multi}.  Kim \etal \cite{kim2017learned} make the first attempt to propose a shallow network (CRN) to take as input the feature maps of convolutional layers and outputs a weighted mask indicating the importance of spatial regions in the feature maps. The resulting mask modulates feature aggregation to create a global representation of the input image. %Xiang \etal \cite{xiang2019multiple} design a convolutional layer, followed by a sigmoid function, to learn saliency masks as spatial weights and then apply to feature maps. 
Noh \etal \cite{noh2017largescale} design a 2-layer CNN with a softplus output layer to compute scores which indicate the importance of different image regions. Inspired by R-MAC, Kim \etal \cite{kimregional2018Regional} employ a pre-trained ResNet101 to train a context-aware attention network using multi-scale feature maps.

Apart from using feature maps as inputs, a whole image can be used to learn feature importance, for which specific networks are needed \cite{mohedano2017saliency},\cite{yang2017two},\cite{wei2019saliency}. Mohedano \cite{mohedano2017saliency} explores different saliency models, including DeepFixNet and Saliency Attentive Model. Yang \etal \cite{yang2017two} and Wei \etal \cite{wei2019saliency} introduce a two-stream network for image retrieval in which the auxiliary stream, DeepFixNet, is used specifically for predicting attention maps, which are then fused with the feature maps produced by the main network.
%In a nutshell, attention mechanisms offer deep networks the capacity to attend on the most important regions within a given image. 
For image retrieval, attention mechanisms can be combined with supervised metric learning \cite{ng2020solar}.

\subsubsection{ Hashing Embedding }
\label{Deep_Hash_Embedding}

Real-valued features extracted by deep networks are typically high-dimensional, and therefore are not well-suited to retrieval efficiency. As a result, there is significant motivation to transform deep features into more compact codes. \textcolor{black}{Since their computational and storage efficiency are beneficial for the ``\textit{efficiency challenge}'', hashing algorithms have been widely used for global \cite{morere2017nested},\cite{yang2018supervised} and local descriptors \cite{song2017deep},\cite{zhao2017spatial},\cite{zheng2016accurate}.}

Hash functions can be plugged as a layer into deep networks, so that hash codes and deep networks can be  simultaneously trained and optimized, either supervised \cite{yang2018supervised} or unsupervised \cite{lin2018unsupervised}. During hash function training, the hash codes of originally similar images are embedded as closely as possible, and the hash codes of dissimilar images are as separated as possible. $d$-dim hash codes from a hash function $h(\cdot)$ for an image $x$ can be formulated as $ b_{x} = h(x) = h\big(f(x;\bm{\theta})\big) \in  \{+1, -1\}^d $. Because hash codes are non-differentiable their optimization is difficult, so $h(\cdot)$ can be relaxed to be differentiable by using \textit{tanh} or \textit{sigmoid} functions \cite{wang2018survey}.

When binarizing real-valued features, it is crucial to preserve image similarity and to improve hash code quality \cite{wang2018survey}. These two aspects are at the heart of hashing algorithms to maximize retrieval accuracy. 

\begin{flushleft}
\emph{a. Hash Functions to Preserve Image Similarity}
\end{flushleft}
Preserving similarity seeks to minimize the inconsistencies between real-valued features and corresponding hash codes, for which a variety of strategies have been adopted.

Loss functions can significantly influence similarity preservation, which includes both supervised and unsupervised methods. With class labels available, many loss functions are designed to learn hash codes in a Hamming space. As a straightforward method, one can optimize either the difference between matrices computed from the binary codes and their supervision labels \cite{morere2017nested},\cite{liu2018deep} or the difference between the hash codes and real-valued deep features \cite{song2017deep},\cite{lin2018unsupervised}.  Song \etal \cite{song2017deep} propose to learn hash codes for regional features in which each local feature is converted to a set of binary codes by multiplying a hash function and the raw RoI features, then the differences between RoI features and hash codes are characterized by an L$_2$ loss. Do \etal \cite{do2017simultaneous} regularize hash codes with a reconstruction loss, which ensure that codes can be reconstructed to their inputs so that similar/dissimilar inputs are mapped to similar/dissimilar hash codes. Lin \etal \cite{lin2018unsupervised} learn hash codes and address the ``\textit{invariance challenge}'' by introducing an objective function which characterize the difference between the binary codes which are computed from the original image and the geometric transformed one.

\begin{flushleft}
\emph{b. Improving Hash Function Quality}
\end{flushleft}
A good hash function seeks to have binary codes uniformly distributed; that is, maximally filling and using the hash code space, normally on the basis of bit uncorrelation and bit balance \cite{wang2018survey},\cite{lin2018unsupervised}. Bit uncorrelation implies that different bits are as independent as possible, so that a given set of bits can aggregate more information within a given code length \cite{lin2018unsupervised}.  Bit balance means that each bit should have a 50\% chance of being +1 or -1, thereby maximizing code variance and information \cite{wang2018survey}.  Mor{\`e}re \etal \cite{morere2017nested} use the uniform distribution $U$(0,1) to build a regularization term to make hash codes distribute evenly where the codes are learned by a Restricted Boltzmann Machine layer. Likewise, Lin \etal \cite{lin2018unsupervised} optimize the mean of learned hash codes to be close to 0.5 to prevent any bit bias towards zero or one.

\section{$\!\!\!\!\!$Retrieval via Learning DCNN Representations }
\label{Retrieval_via_Learning_DCNN_Representations}

The off-the-shelf DCNNs pre-trained on source datasets for classification are quite robust to inter-class variability. However, in most cases, deep features extracted based on off-the-shelf models may not be sufficient for accurate retrieval, even with the strategies discussed in Section \ref{Retrieval_with_Off_the_Shelf_DCNN_Models}. In order for models  to be more effective for retrieval, a common practice is network fine-tuning, \ie updating the pre-stored parameters \cite{yosinski2014transferable}. Fine-tuning methods have been studied extensively to learn better features, \textcolor{black}{whose primary aim is to address the ``\textit{fine-tune challenge}''.} A standard dataset with clear and well-defined ground-truth labels is indispensable for the supervised fine-tuning and subsequently pair-wise supervisory information is incorporated into ranking loss to update networks by regularizing on retrieval representations, otherwise it is necessary to develop unsupervised fine-tuned methods. After network fine-tuning, features can be organized as global or local to perform retrieval. 

\textcolor{black}{For the most feature strategies we presented in Section \ref{Retrieval_with_Off_the_Shelf_DCNN_Models}, including feature extraction, feature embedding and feature aggregation. Note that fine-tuning does not contradict or render irrelevant these feature processing methods; indeed, these strategies are complementary and can be equivalently incorporated as part of network fine-tuning. To this end, this section will survey the strategies which have been developed, based on the patch-level, image-level, or class-level supervision, to fine-tune deep networks for better instance retrieval.}

\begin{figure*}[!t]
\centering
\includegraphics[width=0.9\textwidth]{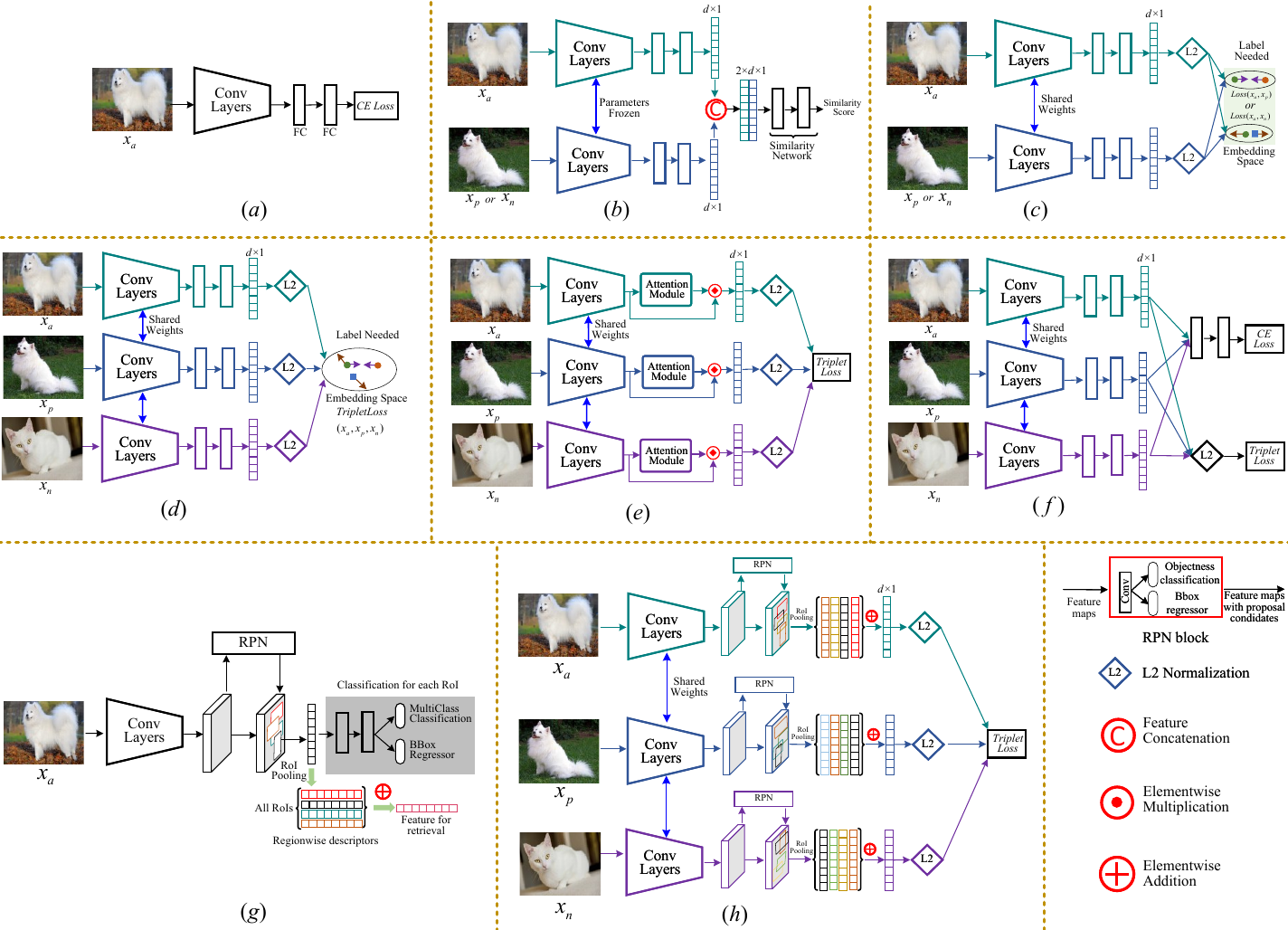}
\vspace{-0.5em}
\caption{Schemes of supervised fine-tuning. Anchor, positive, and negative images are indicated by $x_{a}$, $x_{p}$, $x_{n}$, respectively. (a) classification loss \cite{babenko2014neural};  (b) similarity learning by using a transformation matrix \cite{garcia2017learning}; (c) Siamese loss \cite{radenovic2018fine},\cite{ ong2017siamese},\cite{musgrave2020metric},\cite{lv2018retrieval}; (d) triplet loss \cite{gordo2016deep}; (e) an attention block into DCNNs to highlight regions \cite{song2017deep}; (f) combining classification loss and pairwise ranking loss \cite{xiang2019multiple},\cite{min2020two}; (g) region proposal networks (RPNs) to locate the RoI and highlight specific regions or instances \cite{salvador2016faster}; (h) inserting the RPNs of (g) into DCNNs, such that the RPNs extract regions or instances at the convolutional layer \cite{gordo2016deep},\cite{gordo2017end}.  }
\vspace{-1em}
\label{All_Metric_Learning}
\end{figure*}

\subsection{Supervised Fine-tuning }
\label{Supervised_Fine-tuning}

The way to realize supervised fine-tuning can be determined by the given class labels or pairwise supervisory information.

\subsubsection{Fine-tuning via Classification Loss}
\label{Classification-based_Fine-tuning}

When class labels of a new dataset are available (\eg INSTRE \cite{wang2015instre}, GLDv2 \cite{noh2017largescale},\cite{weyand2020google}), it is preferable to begin with a previously-trained DCNN, trained on a separate dataset, with the backbone DCNN typically chosen from one of AlexNet, VGG, GoogLeNet, or ResNet.
The DCNN can then be fine-tuned, as shown in Figure \ref{All_Metric_Learning} (a), by optimizing its parameters on the basis of a cross entropy loss 
\begin{equation}
\mathcal{L}_{CE}(\hat{p_{i}},y_{i})= - \! \sum^{c}_{i} \! \big(y_{i} \! \times \!log(\hat{p}_{i}) \big)
\label{cross_entropy_loss}
\end{equation}
Here $y_{i}$ and $\hat{p}_{i}$ are the ground-truth labels and the predicted logits, respectively, and $c$ is the total number of categories. The milestone work in such fine-tuning is \cite{babenko2014neural}, in which AlexNet is re-trained on the Landmarks dataset. \textcolor{black}{
According to the class labels, the image-level features are required to compute the logits. Thus, the descriptors extracted from local regions on convolutional feature maps \cite{noh2017largescale},\cite{cao2020unifying} or image patch inputs \cite{teichmann2019detect} are further needed to be aggregated. }

\textcolor{black}{A classification-based fine-tuning method enables to enforce higher similarity for intra-class samples and diversity for inter-class samples. Cao \etal \cite{cao2020unifying} employ the ArcFace loss \cite{Deng2019ArcFaceAA}, which uses the margin-adjusted cosine similarity in the form of softmax loss, to induce smaller intra-class variance and show excellent results for instance retrieval.} Recently, Boudiaf \etal \cite{boudiaf2020unifying} claim that cross entropy loss can minimize intra-class distances while maximizing inter-class distances. Cross entropy loss is, in essence, maximizing a common mutual information between the retrieval features and the ground-truth labels. Therefore, it can be regarded as an upper bound on a new pairwise loss, which has a structure similar to various pairwise ranking losses, of which representatives are introduced below.

\subsubsection{Fine-tuning via Pairwise Ranking Loss}
\label{Verification_based_Learning}

With affinity information (\eg samples from the same group) indicating similar and dissimilar pairs, fine-tuning methods based on pairwise ranking loss learn an optimal metric which minimizes or maximizes the distance of pairs to maintain their similarity. Network fine-tuning via ranking loss involves two types of information \cite{wan2014deep}:
\begin{enumerate}
\item A pair-wise constraint, corresponding to a Siamese network as in Figure \ref{All_Metric_Learning} (c), in which input images are paired with either a positive or negative sample;
\item A triplet constraint, associated with triplet networks as in Figure \ref{All_Metric_Learning} (e), in which anchor images are paired with both similar and dissimilar samples \cite{wan2014deep}.
\end{enumerate}
These pairwise ranking loss based methods are categorized into globally supervised approaches (Figure \ref{All_Metric_Learning} (c,d)) and locally supervised approaches (Figure \ref{All_Metric_Learning} (g,h)), where the former ones learn a metric on global features by satisfying all constraints, whereas the latter ones focus on local areas by only satisfying the given local constraints (\emph{e.g.} region proposals).

To be specific, consider a triplet set $ X \!\! = \!\! \lbrace(x_{a}, x_{p}, x_{n})\} $ in a mini-batch, where $ (x_{a}, x_{p} )$ indicates a similar pair and $ (x_{a}, x_{n} )$ a dissimilar pair. Features $ f(x;\bm{\theta}) $ of one image are extracted by a network $f(\cdot)$ with parameters $\bm{\theta}$, for which we can represent the affinity information for each similar or dissimilar pair as
\begin{equation}
\begin{aligned}
\mathcal{D}_{ij} = \mathcal{D}(x_{i}, x_{j}) = || f(x_{i}; \bm{\theta}) - f(x_{j}; \bm{\theta})||_{2}^{2}
\end{aligned}\label{Distance_pre_defined}
\end{equation}

\begin{flushleft}
\emph{a. Refining with Transformation Matrix}.   
\end{flushleft}

Learning the similarity among input samples can be implemented by optimizing the weights of a linear transformation matrix \cite{garcia2017learning}. It  transforms the concatenated feature pairs into a common latent space using a transformation matrix $ \bm{W} \!\! \in \!\! \mathbb{R}^{2d \times 1} $, where $d$ is the final feature dimension. The similarity score of these pairs are predicted via a sub-network $ \mathcal{S}_{W}(x_{i}, x_{j}) =  f_{W}(f(x_{i}; \bm{\theta})\cup f(x_{j};\bm{\theta}); \bm{W}) $ \cite{garcia2017learning}. In other words, the sub-network $f_{W}$ predicts how similar the feature pairs are. Given the affinity information of feature pairs $ \mathcal{S}_{ij}=\mathcal{S}(x_{i}, x_{j}) \! \in \! \{0,1\} $,  the binary labels 0 and 1 indicate the similar (positive) or dissimilar (negative) pairs, respectively. The training of function $f_{W}$ can be achieved by using a regression loss: 
\begin{equation}
\begin{aligned}
\!\!\!\mathcal{L}_{W}(x_{i}, x_{j}) = & | \mathcal{S}_{W}(x_{i}, x_{j}) - \mathcal{S}_{ij}\big(\text{sim}(x_{i}, x_{j}) + m\big) - \\
&(1-\mathcal{S}_{ij})\big(\text{sim}(x_{i}, x_{j})-m\big) | \label{regressionLoss} 
\end{aligned}
\end{equation}
where $\text{sim}(x_{i}, x_{j}) $ can be the cosine function for guiding the training of $\bm{W}$ and $m$ is a margin. By optimizing the regression loss and updating $ \bm{W} $, deep networks maximize the similarity of similar pairs and minimize that of dissimilar pairs. It is worth noting that the pre-stored parameters in the deep models are frozen when optimizing $ \bm{W} $. The pipeline of this approach is depicted in Figure \ref{All_Metric_Learning} (b).

\begin{flushleft}
\emph{b. Fine-tuning with Siamese Networks}.   
\end{flushleft}

\begin{figure*}[!t]
\centering
\includegraphics[width=\linewidth]{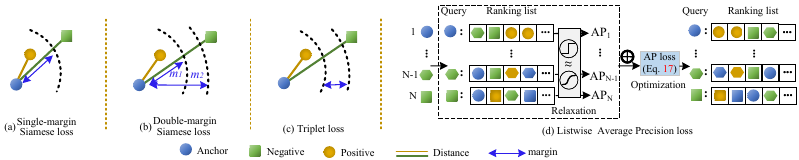}
\vspace{-2em}
\caption{Illustrations of different losses for network fine-tuning. The same shape with different colors denotes images that include the same instance. (a)-(c) have been introduced in the text \cite{radenovic2018fine},\cite{lin2017deephash},\cite{gordo2016deep}. (d) Listwise AP loss considers a mini-batch of $N$ features simultaneously and directly optimizes the Average-Precision computed from these features \cite{revaud2019learning},\cite{brown2020smooth}.
}  \label{mAP_loss_illustration}
\end{figure*}

Siamese networks represent important options in implementing metric learning for fine-tuning, as in Figure \ref{All_Metric_Learning} (c) and Figure \ref{mAP_loss_illustration} (a). It is a structure composed of two branches that share the same weights across layers. Siamese networks are trained on paired data, consisting of an image pair $ (x_{i}, x_{j}) $ such that $ \mathcal{S}(x_{i}, x_{j}) \! \in \! \{0,1\} $. A Siamese loss %, illustrated in Figure~\ref{sampling_strategies}(a), 
is formulated as
\begin{equation}
\begin{aligned}
\mathcal{L}_{Siam}(x_{i}, x_{j})& =  \frac{1}{2}\mathcal{S}(x_{i}, x_{j})\mathcal{D}(x_{i}, x_{j}) \; + \\
&\frac{1}{2}\big(1 - \mathcal{S}(x_{i}, x_{j})\big)\max\big(0,\; m - \mathcal{D}(x_{i}, x_{j})\big)
\end{aligned}\label{contrastive}
\end{equation}
Siamese loss has recently been reaffirmed as a very effective metric in category-level image retrieval, outperforming many more sophisticated losses if implemented carefully \cite{musgrave2020metric}. Enabled by the standard Siamese network, this objective function is used to learn the similarity between semantically relevant samples under different scenarios \cite{radenovic2018fine},\cite{lv2018retrieval}. For example, Radenovi{\'c} \etal \cite{radenovic2018fine} employ a Siamese network on matching and non-matching global feature pairs which are aggregated by GeM-based pooling. The deep network fine-tuned by the Siamese loss generalizes better and converges at higher retrieval performance. Ong \etal \cite{ong2017siamese} leverage the Siamese network to learn image features which are then fed into the Fisher Vector model for further encoding.  Siamese networks can also be applied to hashing learning in which the Euclidean distance $\mathcal{D}(\cdot)$ in Eq. \ref{contrastive} is computed for binary codes \cite{lin2017deephash}.

An implicit drawback of the Siamese loss is that it may penalize similar image pairs even if the margin between these pairs is small or zero \cite{lin2017deephash}, if the constraint is too strong and unbalanced.  At the same time, it is hard to map the features of similar pairs to the same point when images contain complex contents or scenes. To tackle this limitation, Cao \etal \cite{cao2016quartet} adopt a double-margin Siamese loss \cite{lin2017deephash} to relax the penalty for similar pairs by setting a margin $ m_1 $ instead of zero, in which case the original single-margin Siamese loss is re-formulated as 
\begin{equation}
\begin{aligned}
\!\!\!\! \mathcal{L}_{\mathcal{D}\_Siam}(x_{i}, x_{j}) & =  \frac{1}{2}\mathcal{S}(x_{i}, x_{j})\max \big(0, \mathcal{D}(x_{i}, x_{j}) - m_{1} \big) + \\
& \!\! \frac{1}{2}\big(1 - \mathcal{S}(x_{i}, x_{j})\big)\max\big(0, m_{2} - \mathcal{D}(x_{i}, x_{j})\big)
\end{aligned}\label{doublemargincontrastive}
\end{equation}
where $ m_{1} \!\! > \!\! 0 $ and $ m_{2} \!\! > \!\! 0  $ are the margins affecting the similar and dissimilar pairs, respectively, as in Figure \ref{mAP_loss_illustration} (b), meaning that the double margin Siamese loss only applies a contrastive force when the distance of a similar pair is larger than $ m_{1} $. The mAP metric of retrieval is improved when using the double margin Siamese loss \cite{lin2017deephash}.

More recently, transformers have been trained under the regularization of cross entropy \cite{tan2021instance} and Siamese loss \cite{el2021training} for instance-level retrieval and achieved competitive performance, positioning it as an alternative to convolutional architectures. As observed by \cite{el2021training}, the transformer-based architecture is less impacted than convolutional networks by feature collapse since each input feature is projected to different sub-spaces before the multi-headed attention. Moreover, the transformer backbone operates as a learned aggregation operator, thereby avoiding the design of sophisticated feature aggregation methods.

%%%%%% network fine-tuning loss functions

\begin{flushleft}
\emph{c. Fine-tuning with Triplet Networks}.   
\end{flushleft}

Triplet networks optimize similar and dissimilar pairs simultaneously. As shown in Figure \ref{All_Metric_Learning} (d) and Figure \ref{mAP_loss_illustration} (c), the plain triplet networks adopt a ranking loss for training:
\begin{equation}
\begin{aligned}
\mathcal{L}_{Triplet}(x_{a}, x_{p}, x_{n}) =  \max \big( 0, m + \mathcal{D}(x_{a}, x_{p}) - 
 \mathcal{D}(x_{a}, x_{n})\big)\label{triplet}
\end{aligned}
\end{equation}
which indicates that the distance of an anchor-negative pair  $ \mathcal{D}(x_{a}, x_{n}) $ should be larger than that of an anchor-positive pair $ \mathcal{D}(x_{a}, x_{p}) $ by a certain margin $ m $.

\textcolor{black}{Given the datasets that provide bounding box annotations, such as INSTRE, Oxford-5k, Paris-6k, and their variants, the bounding box annotations are used as patch-level supervision to train a region detector which enables the final DCNNs to locate specific regions or objects. As an example, region proposal networks (RPNs) \cite{ren2015faster} is fine-tuned and subsequently plugged into DCNNs and trained end-to-end \cite{salvador2016faster}, as shown in Figure \ref{All_Metric_Learning} (g). RPNs yield the regressed bounding box coordinates of objects and are trained by the multi-class classification loss. Once fine-tuned, RPNs can produce regional features for each detected region by RoI pooling and perform better instance search. }

\textcolor{black}{Further, local supervised metric learning has been explored based on the fact that RPNs \cite{ren2015faster} enable deep models to learn regional features for particular instance objects \cite{min2020two},\cite{song2017deep},\cite{gordo2016deep},\cite{gordo2017end}. RPNs used in the triplet formulation are shown in Figure \ref{All_Metric_Learning} (h). Firstly, regression loss (RPNs loss) is used to minimize the regressed bounding box relative to ground-truth. Then, the regional features for all detected RoIs are aggregated into a global one and L2-normalized for the triplet loss.} Note that, in some cases, jointly training an RPN loss and triplet loss leads to unstable results, a problem addressed in \cite{gordo2016deep} by first training a CNN to produce R-MAC using a rigid grid, after which the parameters in convolutional layers are fixed and RPNs are trained to replace the rigid grid.

Attention mechanisms can also be combined with metric learning for fine-tuning \cite{song2017deep}, as in Figure \ref{All_Metric_Learning} (e), where the attention module is typically end-to-end trainable and takes as input the convolutional feature maps. Song \etal \cite{song2017deep} introduce a convolutional attention layer to explore spatial-semantic information, highlighting regions in images to significantly improve the discrimination for image retrieval.

Recent studies \cite{xiang2019multiple},\cite{min2020two} have jointly optimized the triplet loss and classification loss to further improve network capacity, as shown in Figure \ref{All_Metric_Learning} (f). The overall joint function is 
\begin{equation}
\begin{aligned}
\!\!\!\mathcal{L}_{Joint} = &\lambda_{1} \! \cdot \! \mathcal{L}_{Triplet}(x_{i,a}, x_{i,p}, x_{i,n}) \! + \! \lambda_{2} \!\cdot \! \mathcal{L}_{CE}(\hat{p_{i}},y_{i})
\end{aligned}
\end{equation}
where the cross entropy loss (CE loss) $ \mathcal{L}_{CE} $ is defined in Eq. (\ref{cross_entropy_loss}) and the triplet loss $ \mathcal{L}_{Triplet} $ in Eq. (\ref{triplet}). $ \lambda_{1} $ and $ \lambda_{2} $ are hyper-parameters tuning the tradeoff between the two loss functions.

\subsubsection{Discussion}
\label{Discussion_ranking_loss}

In some cases, pairwise ranking loss cannot effectively learn the variations between samples and still suffers from a weaker generalization capability if the training set is not ordered correctly. Therefore, pairwise ranking loss requires careful sample mining and weighting strategies to obtain the most informative pairs, especially when considering mini-batches.  The hard negative mining strategy is commonly used \cite{arandjelovic2016netvlad},\cite{radenovic2018fine},\cite{tolias2020learning}, however further  sophisticated mining strategies have recently been developed.  Mishchuk \etal \cite{mishchuk2017working} calculate a pair-wise distance matrix on all mini-batch samples to select two closest negative and one anchor-positive pair to form a triplet. %Chen \etal  \cite{chen2017beyond} introduce a margin-based online hard negative mining to select hard samples, where the margin is computed adaptively according to the average distance of two feature distributions from the trained model. 
Instead of traversing all possible two-tuple or three-tuple combinations, it is possible to consider all positive samples in one cluster and negative samples together. Liu \etal \cite{liu2019group} introduce a group-group loss to decrease the intra-group distance and increase the inter-group distance.  Considering all samples may be beneficial for stabilizing optimization and promoting generalization due to a larger data diversity, however the extra computational cost remains an issue to be addressed.

Substantial research has been devoted to pair-wise ranking loss, while cross entropy loss, mainly used for classification, has been largely overlooked. Recently, Boudiaf \etal \cite{boudiaf2020unifying} claim that cross entropy loss can match and even surpass the pair-wise ranking loss when carefully tuned on fine-grained category-level retrieval tasks. In fact, the greatest improvements have come from enhanced training schemes (\eg data augmentation, learning rate polices, batch normalization freeze) rather than intrinsic properties of pairwise ranking loss. Further, although several sophisticated ranking losses have been explored and validated for category-level retrieval, Musgrave \etal \cite{musgrave2020metric} revisited these losses and found that most of them perform on par to vanilla Siamese loss and triplet loss, so there is merit to consider these losses also for instance-level image retrieval tasks.

Both cross entropy loss and pair-wise ranking loss regularize on the embedded features and the corresponding labels so as to maximize their mutual information \cite{boudiaf2020unifying}. Their effectiveness is not guaranteed to give retrieval results that also optimize mAP \cite{revaud2019learning}.  To tackle this limitation one can directly optimize the average precision (AP) metric using the listwise AP loss,
\begin{equation}
\begin{aligned}
\!\!\!\mathcal{L}_{mAP} = 1 - \frac{1}{N} \sum^{N}_{i=1} {\rm AP}(x_{i}^{\top}X_{N}, Y_{i})
\label{AP_loss}
\end{aligned}
\end{equation}
which optimizes the global ranking of thousands of images simultaneously, instead of only a few images at a time. Here $Y_{i}$ is the binary label to evaluate the relevance between batch images. $X_{N} = \{x_{1}, x_{2},...x_{j},...,x_{N}\}$ denotes the features of all images, where each $x_{i}$ is used as a potential query to rank the remaining batch images. Each similarity score $x_{i}^{\top}x_{j}$ can be measured by a cosine function. 

It is demonstrated that training with AP-based loss improves retrieval performance \cite{revaud2019learning},\cite{brown2020smooth}. However average precision, as a metric, is normally non-differentiable. To directly optimize the AP loss during back-propagation, the key is that the indicator function for AP computing  needs to be relaxed using methods such as triangular kernel-based soft assignment \cite{revaud2019learning} or sigmoid function \cite{brown2020smooth}, as shown in Figure \ref{mAP_loss_illustration} (d).

%%%%%%%%%%%%%%%%%%%%%%%%%%%%%%%%%%%%%
\subsection{Unsupervised Fine-tuning }
\label{Unsupervised_Fine-tuning}

Supervised network fine-tuning becomes infeasible when there is insufficient supervisory information, normally because of cost or unavailability. Therefore unsupervised fine-tuning methods for image retrieval are quite necessary, but less studied \cite{iscen2018mining}. 
For unsupervised fine-tuning, two directions are to mine relevance among features via manifold learning, and via clustering techniques, each discussed below.

\subsubsection{Mining Samples with Manifold Learning}
\label{Mining_Samples_with_Manifold_Learning}

\begin{figure}[!t]
\centering
\includegraphics[width=0.5\textwidth]{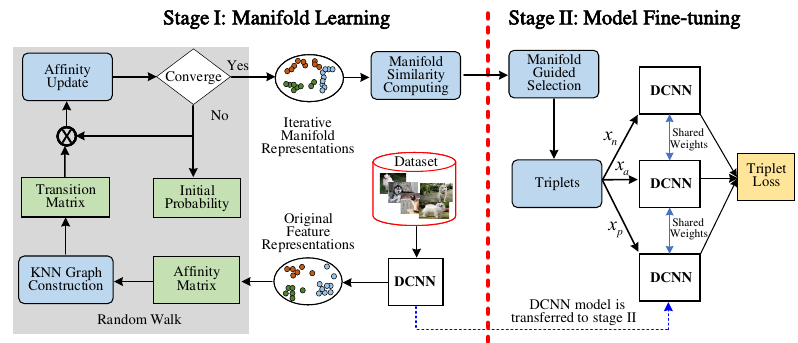}
\caption{Paradigm of manifold learning for unsupervised metric learning, based on triplet loss \cite{iscen2018mining},\cite{zhaomodelling2018}.} 
\label{manifold_learning_pipeline}
\end{figure}

Manifold learning focuses on capturing intrinsic correlations on a manifold structure to mine or deduce relevance, as illustrated in Figure \ref{manifold_learning_pipeline}. Initial similarities between the extracted global features \cite{iscen2018fast} or local features \cite{zhang2016hyperlink},\cite{iscen2017efficient} are used to construct an affinity matrix, which is then re-evaluated and updated using manifold learning \cite{donoser2013diffusion}. According to the manifold similarity in the updated affinity matrix, positive and hard negative samples are selected for metric learning using pairwise ranking loss based functions such as pair loss \cite{gordo2016deep},\cite{iscen2017efficient} or triplet loss \cite{iscen2018mining},\cite{zhaomodelling2018}.  Note that this is different from the aforementioned methods for pairwise ranking loss based fine-tuning methods, where the hard positive and negative samples are explicitly selected from an ordered dataset according to the given affinity information.

It is important to capture the geometry of the manifold of deep features, generally involving two steps \cite{donoser2013diffusion}, known as diffusion. First, the affinity matrix (Figure \ref{manifold_learning_pipeline}) is interpreted as a weighted kNN graph, where each vector is represented by a node, and edges are defined by the pairwise affinities of two connected nodes. Then, the pairwise affinities are re-evaluated in the context of all other elements by diffusing the similarity values through the graph \cite{chang2019explore},\cite{iscen2018mining},\cite{iscen2017efficient},\cite{zhaomodelling2018}, with recent strategies proposed such as regularized diffusion (RDP) \cite{bai2018regularized} and  regional diffusion \cite{iscen2017efficient}. For more details on diffusion methods refer to survey \cite{donoser2013diffusion}.

Most algorithms follow the two steps of \cite{donoser2013diffusion}; the differences among methods lie primarily in three aspects: 
\begin{enumerate}
\item {\bf Similarity initialization},
which affects the subsequent kNN graph construction in an affinity matrix. Usually, an inner product \cite{chang2019explore} or Euclidean distance \cite{xu2017iterative} is directly computed for the affinities.  A Gaussian kernel function can be used \cite{donoser2013diffusion},\cite{zhaomodelling2018}, or consider regional similarity from image patches \cite{iscen2017efficient}.

\item {\bf Transition matrix definition}, a row-stochastic matrix \cite{donoser2013diffusion}, determines the probabilities of transiting from one node to another in the graph. These probabilities are proportional to the affinities between nodes, which can be measured by Geodesic distance (\emph{e.g.} the summation of weights of relevant edges).

\item {\bf Iteration scheme},
to re-valuate and update the values in the affinity matrix by the manifold similarity until some convergence is achieved. Most algorithms are iteration-based \cite{iscen2018mining},\cite{donoser2013diffusion}, as illustrated in Figure \ref{manifold_learning_pipeline}. 
\end{enumerate}
Diffusion process algorithms are indispensable for unsupervised fine-tuning. Better image similarity is guaranteed when it is improved based on initialization (\emph{e.g.} regional similarity \cite{iscen2017efficient} or higher order information \cite{xu2017iterative}). Diffusion is normally iterative and is computationally demanding \cite{zhaomodelling2018}, a limitation which cannot meet the efficiency requirements of image retrieval. To reduce the  computational complexity, Bai \etal \cite{bai2018regularized} propose a regularized diffusion process, facilitated by an efficient iteration-based solver.  Zhao \etal \cite{zhaomodelling2018} regard the diffusion process as a non-linear kernel mapping function, which is then modelled by a deep neural network.  Other studies replace the diffusion process on a kNN graph with a diffusion network \cite{liu2019guided}, which is derived from graph convolution networks \cite{kipf2016semi}, an end-to-end trainable framework which allows efficient computation during training and testing.

Once the manifold space is learned, samples are mined by computing geodesic distances based on the Floyd-Warshall algorithm or by comparing the set difference \cite{iscen2018mining}. The selected samples are fed into deep networks to perform fine-tuning.

\subsubsection{Mining Samples by Clustering}
\label{Mining_Samples_by_Clustering}

Clustering is used to explore proximity information that has been studied in instance-level retrieval \cite{radenovic2016cnn},\cite{tzelepi2018deep},\cite{caron2018deep},\cite{jiang2019unsupervised},\cite{ke2018feature}. The rationale behind these methods is that samples in a cluster are likely to satisfy a degree of similarity.

One class of methods for clustering deep features is via k-means. Given $k$ cluster centroids, during each training epoch a deep network alternates between two steps: first, a soft assignment between the feature representations and the cluster centroids; second, the cluster centroids are refined and, at the same time, the deep network is updated by learning from current high confidence assignments using a certain regularization. These two steps are repeated until a convergence criterion is met, at which point the cluster assignments are used as pseudo-labels \cite{caron2018deep},\cite{jiang2019unsupervised}. Alternatively, the pseudo-labels can be calculated from the samples in a cluster, \eg the mean values. For example, Tzelepi \etal \cite{tzelepi2018deep} compute $ k $ nearest feature representations with respect to a query feature and then compute their mean vectors, which is used as a target for the query feature. In this case, fine-tuning is performed by minimizing the squared distance between each query feature and the mean of its $ k $ nearest features. Liu \etal \cite{liu2018deep} propose a self-taught hashing algorithm using a kNN graph construction to generate pseudo labels that are used to analyze and guide network training. Shen \etal \cite{shen2018matchable} and Radenovi{\'c} \etal \cite{radenovic2016cnn},\cite{radenovic2018fine} use Structure-from-Motion (SfM) for each image cluster to explore sample reconstructions to select images for triplet loss. Clustering methods depend on the Euclidean distance, making it difficult to reveal the intrinsic relationship between objects.

There are further techniques for instance retrieval, such as by using AutoEncoder \cite{do2017simultaneous},\cite{hu2019towards}, generative adversarial networks (GANs)  \cite{bai2021unsupervised}, convolutional kernel networks \cite{paulin2015local},\cite{paulin2017convolutional}, and graph convolutional networks \cite{liu2019guided}. For these methods, they focus on devising novel unsupervised frameworks to realize unsupervised learning, instead of iterative similarity diffusion or cluster refinement on feature space. For example, instead of performing iterative traversal on a set of nearest neighbors defined by kNN graph, Liu \etal \cite{liu2019guided} employ graph convolutional networks \cite{kipf2016semi} to directly encode the neighbor information into image descriptors and then train the deep models to learn a new feature space. This method is demonstrated to significantly improve retrieval accuracy while maintaining efficiency.  GANs are also explored, for the first time, for instance-level retrieval in an unsupervised fashion \cite{bai2021unsupervised}. The generator retrieves images that contain similar instances as a given image, while the discriminator judges whether the retrieved images have the specified instance which appeared in the query image. During training, the discriminator and the generator play a min-max game via an adversarial reward which is computed based on the cosine distance between the query image and the images retrieved by the generator.

\section{State of the Art Performance}
\label{State_of_the_Art_Performance}

\subsection{Datasets}
\label{Datasets_and_Evaluation_Criteria}

To demonstrate the effectiveness of methods, we choose the following commonly-used datasets for performance comparison:

\textbf{UKBench (UKB)} \cite{nister2006scalable} consists of 10,200 images of objects. This dataset has 2,550 groups of images, each group having four images of the same object from different viewpoints or illumination conditions, which can be regarded as a kind of class-level supervision information. All images can be used as a query.

\textbf{Holidays} \cite{jegou2008hamming} consists of 1,491 images collected from personal holiday albums. Most images are scene-related. The dataset comprises 500 groups of similar images with a single query image for each group. The dataset also provides position information of the interest regions for each image.

\textbf{Oxford-5k} \cite{philbin2007object} consists of 5,062 images for 11 Oxford buildings. Each building is associated with five hand-drawn bounding box queries. \textcolor{black}{According to the relevance level, each image of the same building is assigned a label \textit{Good} (\ie \textit{positive}), \textit{OK} (\ie \textit{positive}), \textit{Junk}, or \textit{Bad} (\ie \textit{negative}). \textit{Junk} images can be 
discarded or regarded as \textit{negative} examples \cite{radenovic2018revisiting},\cite{jegou2010aggregating}.
To build a tuple for each given query, one can select a positive example whose label corresponds to \textit{Good} or \textit{OK} in the same category, and select one negative example from each of the remaining building categories. Furthermore, an additional disjoint set of 100,000 distractor images is added to obtain Oxford-105k.}

\textbf{Paris-6k} \cite{Lostphilbin2008lost1} includes 6,412 images and is categorized into 12 groups by architecture. %The dataset has 500 query images for evaluation. 
The supervision information can be used like that of Oxford-5k. Likewise, an additional disjoint set of 100,000 distractor images is added to obtain Paris-106k.

\textbf{INSTRE} \cite{wang2015instre} consists of 28,543 images from 250 different object classes, including three disjoint subsets\footnote{https://github.com/imatge-upc/salbow}: INSTRE-S1, INSTRE-S2, INSTRE-M. 
%Both INSTRE-S1 and INSTRE-S2 have 100 object classes. INSTRE-S1 contains 11,011 images and INSTRE-S2 contains 12,059 images. INSTRE-M contains 5,473 images distributed in 50 object classes. 
\textcolor{black}{INSTRE dataset has bounding box annotations, providing single-labelled and double-labelled class information for single- and multiple-object retrieval, respectively. One can use the class information to build a tuple, with two positive examples from the same class and one negative from one of the remaining classes. The performance evaluation on INSTRE in our experiments follows the protocol in \cite{iscen2017efficient}. }

\textbf{Google Landmarks Dataset (GLD)} \cite{noh2017largescale},\cite{weyand2020google} consists of GLD-v1 and GLD-v2. \textcolor{black}{GLD-v2 is mainly recommended to use and it has the advantage of stability where all images have permissive licenses \cite{cao2020unifying}. GLD-v2 is divided into three subsets: (i) 118k query images with ground-truth annotations, (ii) 4.1M training images of 203k landmarks with labels, and (iii) 762k index images of 101k landmarks. Due to its large scale, GLD-v2 provides class-level ground-truth which can be used to build training tuples. Due to its image diversity, it may produce clutter images for each landmark so it is necessary to introduce pre-processing methods to select the more relevant images \cite{yokoo2020two}. Finally, the training set is cleaned by removing these clutters, consisting of a subset ``GLD-v2-clean'' containing 1.6M images of 81k landmarks. Since Google landmarks dataset stills lack bounding box for objects of interest, Teichmann \etal \cite{teichmann2019detect} provide a new dataset of landmark bounding boxes, based on GLD. This patch-level supervision information can help locate the most relevant regions.  }

\textcolor{black}{Note that, additional queries and distractor images have been added into Oxford-5k and Paris-6k, producing the Revisited Oxford ($\mathcal{R}$Oxford) and Revisited Paris ($\mathcal{R}$Paris) datasets where each image of the same building is assigned a label \textit{Easy}, \textit{Hard}, \textit{Unclear}, or \textit{Negative} \cite{radenovic2018revisiting}. Different label combinations are used as \textit{positive} according to the difficulty level of different setups. During testing, if there are no positive images for a query, then that query is excluded from the evaluation. For details, we refer the reader to \cite{radenovic2018revisiting}. We undertake partial comparisons under the \textit{hard} evaluation protocol on these revisited datasets. }

%%%% Evaluation Metrics
\subsection{Evaluation Metrics}

\textbf{Average precision} (AP) refers to the coverage area under the precision-recall (PR) curve. A larger AP implies a higher PR curve and better retrieval accuracy. AP can be calculated as $ AP=\frac{1}{R} \sum_{k=1}^N P(k)\cdot rel(k) $,
% \begin{equation}
% AP = \frac{\sum_{k=1}^N P(k)\cdot rel(k)}{R} \label{AP}
% \end{equation}
where $ R $ denotes the number of relevant results for the query image from the total number $ N $ of images. $ P(k) $ is the precision of the top $ k $ retrieved images, and $ rel(k) $ is an indicator function equal to 1 if the item within rank $ k $ is a relevant image and 0 otherwise.  Mean average precision (mAP) is adopted for the evaluation over all query images, \ie $mAP=\frac1Q{\sum_{q=1}^Q}AP(q) $,
% \begin{equation} \frac1Q{\sum_{q=1}^Q}AP(q)
% \end{equation}
where $ Q $ is the number of query images.

The \textbf{N-S score} is a metric used for UKBench \cite{nister2006scalable}; the N-S score is the average for the top-4 precision over the dataset.

\subsection{Performance Comparison and Analysis}

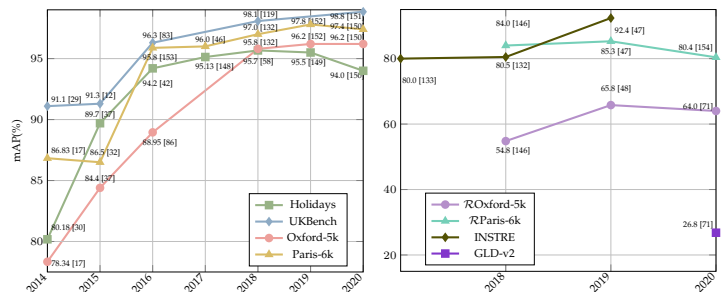
\begin{figure}[t]
\centering
\begin{subfigure}{0.248\textwidth}
\resizebox{\columnwidth}{!}{%
\begin{tikzpicture} %% use pgfplotsset
\scriptsize
\begin{axis}[legend style={at={(0.95,0.3)}, scale=0.5, anchor=north east, font= \scriptsize}, symbolic x coords={2014, 2015, 2016, 2017, 2018, 2019, 2020 }, grid=both, ylabel={mAP(\%)}, ymin=77.5, ymax=99, xmin=2014, xmax=2020, xtick=data, x tick label style={rotate=45,anchor=east} ];

%% Holidays results mAP
\addlegendentry{Holidays}
\addplot[mark=square*,thick, line width=0.45mm, color={rgb, 255:red,152;green,179;blue,134}, on layer=main] coordinates { (2014, 80.18) (2015,89.7) (2016,94.2) (2017,95.13) (2018,95.67) (2019, 95.5) (2020, 94.0) };
\node [above right=0cm, font=\tiny] at (axis cs:2014, 80.5) {80.18 \cite{gong2014multi}}; \node [ font=\tiny]  at (axis cs:2015, 90.4) {89.7 \cite{sharif2015baseline}}; \node [above right=-0.3cm, font=\tiny]  at (axis cs:2016, 93.4) {94.2 \cite{gordo2016deep}}; \node [above right=-0.3cm, font=\tiny]  at (axis cs:2017, 94.8) {95.13 \cite{alzu2017content}}; \node [above right=-0.4cm, font=\tiny]  at (axis cs:2018, 95.6) {95.7 \cite{bai2018regularized}};  \node [above right=-0.5cm, font=\tiny]  at (axis cs:2019, 95.9) {95.5 \cite{husain2019remap}}; \node [above right=-0.8cm, font=\tiny]  at (axis cs:2020, 96) {94.0 \cite{alemu2020multi}};

%% UKBench results mAP
\addlegendentry{UKBench}
\addplot[mark=diamond*,thick,line width=0.45mm, color={rgb,255:red,141;green,170;blue,196}, on layer=main,] coordinates {(2014,91.1) (2015,91.3) (2016,96.3) (2018,98.08) (2020,98.84) }; \node [above right=0cm, font=\tiny] at (axis cs:2014, 91) {91.1 \cite{sharif2014cnn}}; \node [above right=-0.4cm, font=\tiny]  at (axis cs:2015, 92.8) {91.3 \cite{babenko2015aggregating}}; \node [above right=-0.3cm, font=\tiny]  at (axis cs:2016, 97.4) {96.3 \cite{azizpour2016factors}}; \node [above right=-0.4cm, font=\tiny]  at (axis cs:2018, 99.5) {98.1 \cite{yang2018dynamic}}; \node [above right=-0.8cm, font=\tiny]  at (axis cs:2020, 100.8) {98.8 \cite{valem2020graph}};

%% oxford-5k results mAP
\addlegendentry{Oxford-5k}
\addplot[mark=*,thick, on layer=background, line width=0.45mm, color={rgb, 255:red,237;green,160;blue,155}] coordinates { (2014,78.34) (2015,84.4) (2016,88.95) %(2017,95.8) 
(2018,95.8) (2019,96.2)  (2020,96.2) }; \node [above right=0cm, font=\tiny] at (axis cs:2014, 77.5) {78.34 \cite{wan2014deep}};  \node [ font=\tiny]  at (axis cs:2015, 85.15) {84.4 \cite{sharif2015baseline}}; \node [above right=-0.3cm, font=\tiny]  at (axis cs:2016, 88.6) {88.95  \cite{zheng2016accurate}}; %\node [above right=-0.7cm, font=\tiny]  at (axis cs:2017, 97.6) {95.8  \cite{iscen2017efficient}}; 
\node [above right=-0.4cm, font=\tiny]  at (axis cs:2018, 97.2) {95.8 \cite{iscen2018fast}}; \node [above right=-0.5cm, font=\tiny]  at (axis cs:2019, 98) {96.2 \cite{yang2019efficient22}}; \node [above right=-0.8cm, font=\tiny]  at (axis cs:2020, 99.1) {96.2 \cite{alemu2020multi}};

%% Paris-6k results mAP
\addlegendentry{Paris-6k}
\addplot[mark=triangle*,thick,line width=0.45mm, color={rgb, 255:red,217;green,188;blue,102}, on layer=main,] coordinates { (2014,86.83) (2015,86.5) (2016,95.88) (2017,96.0) (2018, 97.0) (2019, 97.8) (2020, 97.4) };
\node [above right=0cm, font=\tiny] at (axis cs:2014, 86.8) {86.83 \cite{wan2014deep}}; \node [above right=-0.3cm, font=\tiny]  at (axis cs:2015, 87.8) {86.5 \cite{tolias2015particular}}; \node [above right=-0.3cm, font=\tiny]  at (axis cs:2016, 95.6) {95.8 \cite{yang2016cross}};\node [above right=-0.3cm, font=\tiny]  at (axis cs:2017, 97.0) {96.0 \cite{radenovic2018fine}}; \node [above right=-0.4cm, font=\tiny]  at (axis cs:2018, 98.4) {97.0 \cite{iscen2018fast}}; \node [above right=-0.5cm, font=\tiny]  at (axis cs:2019, 99.3) {97.8 \cite{yang2019efficient22}}; \node [above right=-0.8cm, font=\tiny]  at (axis cs:2020, 100) {97.4 \cite{alemu2020multi}};

\end{axis}
\end{tikzpicture}
} 
\vspace{-2.1em}
% \caption{}
\end{subfigure}
\begin{subfigure}{0.235\textwidth}
\resizebox{\columnwidth}{!}{%
\begin{tikzpicture} %% use pgfplotsset
\scriptsize
\begin{axis}[legend style={at={(0.425,0.3)}, scale=0.5, anchor=north east, font=\scriptsize}, symbolic x coords={2017, 2018, 2019, 2020}, grid=both, ylabel={}, ymin=15, ymax=95, xmin=2017, xmax=2020, xtick=data, x tick label style={rotate=45, anchor=east} ];

%%%%Revisited Oxford5k
\addlegendentry{$\mathcal{R}$Oxford-5k}
\addplot[mark=*,thick,line width=0.45mm, color={rgb, 255:red,181;green,144;blue,202}, on layer=main,] coordinates { (2018, 54.8) (2019, 65.8) (2020, 64.0) };
\node [above right=-0.3cm, font=\tiny]  at (axis cs:2018, 53.8) {54.8 \cite{radenovic2018revisiting}}; \node [above right=-0.3cm, font=\tiny]  at (axis cs:2019, 72.5) {65.8 \cite{chang2019explore}}; \node [above right=-0.8cm, font=\tiny]  at (axis cs:2020, 74.0) {64.0 \cite{cao2020unifying}};

%%%%Revisited Paris6k
\addlegendentry{$\mathcal{R}$Paris-6k}
\addplot[mark=triangle*,thick,line width=0.45mm, color={rgb, 255:red,117;green,207;blue,184}, on layer=main,] coordinates { (2018, 84.0) (2019, 85.3) (2020, 80.4) };
\node [above right=-0.3cm, font=\tiny]  at (axis cs:2018, 92.5) {84.0 \cite{radenovic2018revisiting}}; \node [above right=-0.3cm, font=\tiny]  at (axis cs:2019, 84.0) {85.3 \cite{liu2019guided}}; \node [above right=-0.9cm, font=\tiny]  at (axis cs:2020, 93.5) {80.4 \cite{ouyang2020collaborative}};

%%%%INSTRE
\addlegendentry{INSTRE}
\addplot[mark=diamond*,thick,line width=0.45mm, color={rgb, 255:red,83;green,82;blue,4}, on layer=main,] coordinates { (2017, 80) (2018, 80.5) (2019, 92.4)};
\node [above right=-0.05cm, font=\tiny]  at (axis cs:2017, 71.5) {80.0 \cite{iscen2017efficient}}; \node [above right=-0.3cm, font=\tiny]  at (axis cs:2018, 79.5) {80.5 \cite{iscen2018fast}}; \node [above right=0cm, font=\tiny]  at (axis cs:2019, 86) {92.4 \cite{liu2019guided}} ;

%%%%GLD-V2
\addlegendentry{GLD-v2}
\addplot[mark=square*,thick,line width=0.45mm, color={rgb, 255:red,130;green,54;blue,203}, on layer=main,] coordinates {(2020, 26.8) };
\node [above right=-0.8cm, font=\tiny] at (axis cs:2020, 38) {26.8 \cite{cao2020unifying}};
\end{axis};
\end{tikzpicture}
}
\vspace{-2.1em}
% \caption{}
\end{subfigure}
\caption{Performance improved from 2014 to 2020.} 
\label{datasets_progress_2014_2_2020}
\end{figure}

\textbf{Overview.} Figure \ref{datasets_progress_2014_2_2020} summarizes the performance over 6 datasets from 2014 to 2020. Early on, the powerful feature extraction of DCNNs led to rapid improvements. Subsequent key ideas have been to extract instance features at the region level to reduce image clutter \cite{gong2014multi}, and to improve feature discriminativity by using methods including feature fusion \cite{zheng2016accurate},\cite{alzu2017content},\cite{valem2020graph}, feature aggregation \cite{tolias2015particular},\cite{razavian2016visual}, and feature embedding \cite{zheng2016accurate}. Fine-tuning is an important strategy to improve performance by tuning deep networks specific for learning instance features \cite{bai2018regularized},\cite{alemu2020multi}. For instance, the accuracy increases steadily from 78.34\% \cite{wan2014deep} to 96.2\% \cite{yang2019efficient22} on the Oxford-5k dataset when manifold learning is used to fine-tune deep networks. The mAP on $\mathcal{R}$Paris-6k and $\mathcal{R}$Oxford-5k is smaller than Paris-6k and Oxford-5k, leaving room for improvement.

We report results using off-the-shelf models (Table \ref{Table_retrieval_off_the_shelf}) and fine-tuning networks (Table \ref{Metric_learning_table_B}). In Table \ref{Table_retrieval_off_the_shelf}, single-pass and multiple-pass are analyzed, while supervised and unsupervised fine-tuning are compared in Table \ref{Metric_learning_table_B}. Since there are many aspects that vary across the different methods, making them not directly comparable, we mainly draw some general claims or trends based on the collected results.

%%%%%%%%%%%%%%%%%%%%%%%%%%%%%%%%%%%%
%%%%%%%%%%%%%%%%%% Tables
%%%%%%%%%%%%%%%%%%%%%%%%%%%%%%%%%%%

\begin{table*}[t]
\centering % Center table 
\caption{\footnotesize Performance evaluation of off-the-shelf DCNN models. ``$ \bullet $'' indicates that  the models or layers are combined to learn features; ``PCA$_{w}$ indicates PCA with whitening on the extracted features to improve robustness; ``MP'' means Max Pooling; ``SP'' means Sum Pooling. The CNN-M network with ``$\ast$'' has an architecture similar to that of AlexNet. ``-'' means that the results were not reported. }
\vspace{-1em}
\label{Table_retrieval_off_the_shelf}
\renewcommand{\arraystretch}{1.0}
\resizebox{19.5cm}{!}{
\begin{tabular}{!{\vrule width1.2bp}c|c|c|c|c|c|c|c|c|c|p{9cm}!{\vrule width1.2bp}}

\Xhline{1pt}
\footnotesize Type  & \footnotesize \shortstack [c]  { Method }  & \footnotesize \shortstack [c] {Backbone \\ DCNN}	& \footnotesize \shortstack [c] {Output \\ Layer} & \footnotesize \shortstack [c] {Embed. \\ Aggre.}	& \footnotesize \shortstack [c] { Feat. \\ Dim } & \footnotesize \shortstack [c] {Holidays} & \footnotesize \shortstack [c] {UKB} & \footnotesize \shortstack [c] {Oxford5k \\ (+100k)} &  \footnotesize \shortstack [c] {Paris6k \\ (+100k)} &  \footnotesize Brief Conclusions and Highlights \\
\Xhline{1.0pt}

\multirow{5}{*}{  \raisebox{-15ex}[0pt]{\begin{tabular}[c]{@{}c@{}}  \rotatebox{90}{Single-pass} \end{tabular}}}  & \footnotesize \raisebox{-3ex}[0pt]{ \shortstack [c] {  Neural \\  codes \cite{babenko2014neural}} } & \footnotesize	\raisebox{-2ex}[0pt]{ AlexNet} & \footnotesize	\raisebox{-2ex}[0pt]{ \shortstack [c] { FC6}}	& \footnotesize \raisebox{-2ex}[0pt]{\shortstack [c] { PCA}}   & \footnotesize \raisebox{-2ex}[0pt]{ $128$} & \footnotesize \raisebox{-2ex}[0pt]{ $74.7$} & \footnotesize \raisebox{-3ex}[0pt]{\shortstack [c] { $ 3.42 $ \\ \textcolor[rgb]{0,0,0}{(N-S)}}}& \footnotesize \raisebox{-3ex}[0pt]{ \shortstack [c] {  $43.3$\\ (38.6)}} & \footnotesize	\raisebox{-2.0ex}[0pt]{\shortstack [c] {$-$}} & \footnotesize  Compressed neural codes of different layers are explored. AlexNet is also fine-tuned for retrieval.\\   \cline{2-11}

%%%%
& \footnotesize \raisebox{-2ex}[0pt]{ \shortstack [c] { SPoC \cite{babenko2015aggregating}} } & \footnotesize	\raisebox{-2ex}[0pt]{ VGG16} & \footnotesize	\raisebox{-2ex}[0pt]{ \shortstack [c] { Conv5}}	& \footnotesize \raisebox{-2.5ex}[0pt]{\shortstack [c] { SPoC + \\ PCA$_{w} $}}   & \footnotesize \raisebox{-2ex}[0pt]{ 256} & \footnotesize \raisebox{-2ex}[0pt]{ 80.2} & \footnotesize \raisebox{-3ex}[0pt]{\shortstack [c] { $ 3.65 $ \\ \textcolor[rgb]{0,0,0}{(N-S)}}} & \footnotesize \raisebox{-3ex}[0pt]{ \shortstack [c] { $58.9$\\ (57.8)}} & \footnotesize	\raisebox{-2.0ex}[0pt]{\shortstack [c] {$-$}} & \footnotesize Exploring Gassian weighting scheme \ie the centering prior, to improve the discrimination of
features. \\   \cline{2-11}

%%%%
& \footnotesize \raisebox{-2ex}[0pt]{ \shortstack [c] {  CroW \cite{kalantidis2016cross}} } & \footnotesize	\raisebox{-2ex}[0pt]{VGG16} & \footnotesize	\raisebox{-2ex}[0pt]{ \shortstack [c] { Conv5}}	& \footnotesize \raisebox{-2.5ex}[0pt]{\shortstack [c] { CroW + \\ PCA$_{w} $}}   & \footnotesize \raisebox{-2ex}[0pt]{256} & \footnotesize \raisebox{-2ex}[0pt]{ $85.1$} & \footnotesize \raisebox{-2ex}[0pt]{\shortstack [c] { $-$}} & \footnotesize \raisebox{-3ex}[0pt]{ \shortstack [c] {$68.4$\\ (63.7)}} & \footnotesize	\raisebox{-3.0ex}[0pt]{\shortstack [c] { $ 76.5 $\\ (69.1) }} & \footnotesize  The spatialwise and channelwise weighting mechanisms are utilized to highlight crucial convolutional features. \\   \cline{2-11}

%%%%
& \footnotesize \raisebox{-2ex}[0pt]{ \shortstack [c] {  R-MAC \cite{tolias2015particular}} } & \footnotesize	\raisebox{-2ex}[0pt]{ VGG16} & \footnotesize	\raisebox{-2ex}[0pt]{ \shortstack [c] {  Conv5}}	& \footnotesize \raisebox{-2.5ex}[0pt]{\shortstack [c] { R-MAC + \\  PCA$_{w} $}}   & \footnotesize \raisebox{-2ex}[0pt]{512} & \footnotesize	\raisebox{-2.0ex}[0pt]{\shortstack [c] {$-$}} & \footnotesize \raisebox{-2ex}[0pt]{\shortstack [c] {$-$ }} & \footnotesize \raisebox{-3ex}[0pt]{ \shortstack [c] {  $66.9$\\ (61.6)}} & \footnotesize	\raisebox{-3.0ex}[0pt]{\shortstack [c] { $-$ \\ (75.7) }} & \footnotesize  Sliding windows with different scales on convolutional feature maps to encode multiple image regions.\\   \cline{2-11}

%%%%

& \footnotesize \raisebox{-3ex}[0pt]{ \shortstack [c] { Multi-layer \\  CNN \cite{yu2017exploiting}} } & \footnotesize	\raisebox{-2ex}[0pt]{VGG16} & \footnotesize	\raisebox{-3ex}[0pt]{ \shortstack [c] { FC6 $\bullet$ \\ Conv4$\sim$5 }}	& \footnotesize \raisebox{-2ex}[0pt]{\shortstack [c] { SP}}   & \footnotesize \raisebox{-2ex}[0pt]{4096} & \footnotesize \raisebox{-2ex}[0pt]{  91.4} & \footnotesize \raisebox{-3ex}[0pt]{\shortstack [c] { $ 3.68 $ \\ \textcolor[rgb]{0,0,0}{(N-S)}}} & \footnotesize \raisebox{-3ex}[0pt]{ \shortstack [c] { $61.5$\\($-$)}} & \footnotesize	\raisebox{-2.0ex}[0pt]{\shortstack [c] {$-$}} & \footnotesize Layer-level feature fusion and the complementary properties of different layers are explored.\\   \cline{2-11} 

%%%%
& \footnotesize \raisebox{-2ex}[0pt]{ \shortstack [c] { BLCF \cite{mohedano2016bags}} } & \footnotesize	\raisebox{-2ex}[0pt]{ VGG16} & \footnotesize	\raisebox{-2ex}[0pt]{ \shortstack [c] {  Conv5}}	& \footnotesize \raisebox{-2.5ex}[0pt]{\shortstack [c] { BoW + \\ PCA$_{w} $}}   & \footnotesize \raisebox{-2ex}[0pt]{ 25k} & \footnotesize \raisebox{-2ex}[0pt]{$-$} & \footnotesize \raisebox{-2ex}[0pt]{\shortstack [c] { $-$}}& \footnotesize \raisebox{-3ex}[0pt]{ \shortstack [c] {  $73.9$\\ (59.3)}} & \footnotesize	\raisebox{-3.0ex}[0pt]{\shortstack [c] { $ 82.0 $\\  (64.8) }} & \footnotesize  Both global features and local features are explored, demonstrating that local features have higher accuracy. \\   \cline{2-11} 

\Xhline{1.0pt}
 
 %%%%%%%%%%%%%%%%%%%%%%%%%%%%%%%%
 %%% multiple pass
\multirow{7}{*}{  \raisebox{-20ex}[0pt]{\begin{tabular}[c]{@{}c@{}}  \rotatebox{90}{Multiple-pass} \end{tabular}}}

%%%
& \footnotesize \raisebox{-2ex}[0pt]{ \shortstack [c] {  OLDFP  \cite{reddy2015object}} } & \footnotesize \raisebox{-2ex}[0pt]{\shortstack [c] {AlexNet }} & \footnotesize	\raisebox{-2ex}[0pt]{ \shortstack [c] {  FC6}}	& \footnotesize \raisebox{-2.6ex}[0pt]{\shortstack [c] { MP \\  +  PCA$_{w} $ }}   & \footnotesize \raisebox{-2ex}[0pt]{ 512} & \footnotesize \raisebox{-2ex}[0pt]{ 88.5} & \footnotesize \raisebox{-3ex}[0pt]{\shortstack [c] {$ 3.81 $ \\ \textcolor[rgb]{0,0,0}{(N-S)}}}  & \footnotesize \raisebox{-3ex}[0pt]{ \shortstack [c] {  $60.7$\\ ($-$)}} & \footnotesize \raisebox{-3ex}[0pt]{ \shortstack [c] { $66.2$\\ ($-$)}} & \footnotesize  Exploring the impact of proposal number. Patches are extracted by RPNs (see Figure \ref{MultiplePass} (d)) and the features are encoded in an orderless way.\\   \cline{2-11}

%%%
& \footnotesize \raisebox{-1.6ex}[0pt]{ \shortstack [c] {  MOP-CNN   \cite{gong2014multi}} } & \footnotesize \raisebox{-2ex}[0pt]{\shortstack [c] { AlexNet }} & \footnotesize	\raisebox{-2ex}[0pt]{ \shortstack [c] { FC7}}  &  \footnotesize \raisebox{-2.5ex}[0pt]{\shortstack [c] { VLAD \\  +  PCA$_{w} $ }}  & \footnotesize \raisebox{-2ex}[0pt]{ 2048} & \footnotesize \raisebox{-2ex}[0pt]{ 80.2} & \footnotesize \raisebox{-2ex}[0pt]{\shortstack [c] { $-$}}   & \footnotesize	\raisebox{-2.0ex}[0pt]{\shortstack [c] {$-$}} &  \footnotesize	\raisebox{-2.0ex}[0pt]{\shortstack [c] {$-$}} & \footnotesize Image patches are extracted densely, as shown in Figure \ref{MultiplePass} (c). Multi-scale patch features are further embedded into VLAD descriptors. \\   \cline{2-11}

%%%
& \footnotesize \raisebox{-2ex}[0pt]{ \shortstack [c] {  CNNaug-ss \cite{sharif2014cnn}} } & \footnotesize \raisebox{-3ex}[0pt]{\shortstack [c] { Overfeat\\  \cite{sermanet2013overfeat}}} & \footnotesize	\raisebox{-2ex}[0pt]{ \shortstack [c] { FC}}		& \footnotesize \raisebox{-2ex}[0pt]{\shortstack [c] {  PCA$_{w} $ }}   & \footnotesize \raisebox{-2ex}[0pt]{ 15k} & \footnotesize \raisebox{-2ex}[0pt]{ 84.3} & \footnotesize \raisebox{-3ex}[0pt]{\shortstack [c] { $ 91.1 $ \\ \textcolor[rgb]{0,0,0}{(mAP)}}}  & \footnotesize \raisebox{-3ex}[0pt]{ \shortstack [c] { $68.0$\\($-$)}} & \footnotesize \raisebox{-3ex}[0pt]{ \shortstack [c] { $79.5$\\ ($-$)}} & \footnotesize Image patches are extracted densely, as shown in Figure \ref{MultiplePass} (c). Image regions at different locations with different sizes are included.\\   \cline{2-11} 

%%%
& \footnotesize \raisebox{-2ex}[0pt]{ \shortstack [c] {  MOF \cite{li2016exploiting}} } &  \footnotesize	\raisebox{-3ex}[0pt]{ \shortstack [c] { CNN-M$^{\ast}$ \\ \cite{chatfield2014return} }} & \footnotesize	\raisebox{-3ex}[0pt]{ \shortstack [c] { FC7 $\bullet$ \\  Conv }}	& \footnotesize \raisebox{-3ex}[0pt]{\shortstack [c] { SP or MP\\ +  BoW }}   & \footnotesize \raisebox{-2ex}[0pt]{ 20k} & \footnotesize \raisebox{-2ex}[0pt]{ 76.8} & \footnotesize \raisebox{-3ex}[0pt]{\shortstack [c] { $ 3.00 $ \\ \textcolor[rgb]{0,0,0}{(N-S)}}}  & \footnotesize	\raisebox{-2.0ex}[0pt]{\shortstack [c] {$-$}} &  \footnotesize	\raisebox{-2.0ex}[0pt]{\shortstack [c] {$-$}} & \footnotesize Exploring layer-level fusion scheme. Image patches are extracted using spatial pyramid modeling, as shown in Figure \ref{MultiplePass} (b).\\   \cline{2-11} 

%%%
& \footnotesize \raisebox{-2.7ex}[0pt]{ \shortstack [c] { Multi-scale \\  CNN \cite{razavian2016visual}} } & \footnotesize	\raisebox{-2ex}[0pt]{ VGG16} & \footnotesize	\raisebox{-2ex}[0pt]{ \shortstack [c] {  Conv5}} & \footnotesize \raisebox{-2.7ex}[0pt]{\shortstack [c] { SP or MP\\  +  PCA$_{w} $ }}   & \footnotesize \raisebox{-2ex}[0pt]{ 32k} & \footnotesize \raisebox{-2ex}[0pt]{  89.6} & \footnotesize \raisebox{-3ex}[0pt]{\shortstack [c] { $ 95.1 $ \\ \textcolor[rgb]{0,0,0}{(mAP)}}}  & \footnotesize \raisebox{-3ex}[0pt]{ \shortstack [c] {  $84.3$\\ ($-$)}} & \footnotesize \raisebox{-3ex}[0pt]{ \shortstack [c] {  $87.9$\\ ($-$)}} & \footnotesize Image patches are extracted in a dense manner, as shown in Figure \ref{MultiplePass} (c). Geometric invariance is considered when aggregating patch features.\\   \cline{2-11} 

%%%
& \footnotesize \raisebox{-2ex}[0pt]{ \shortstack [c] {  LDD \cite{cao2017local}} } & \footnotesize	\raisebox{-2ex}[0pt]{ VGG19} & \footnotesize	\raisebox{-2ex}[0pt]{ \shortstack [c] {  Conv5}}	& \footnotesize \raisebox{-2.6ex}[0pt]{\shortstack [c] {BoW \\ +  PCA$_{w} $ }}   & \footnotesize \raisebox{-2ex}[0pt]{500k} & \footnotesize \raisebox{-2ex}[0pt]{ 84.6 } & \footnotesize \raisebox{-2ex}[0pt]{\shortstack [c] {$-$  }}  & \footnotesize \raisebox{-3ex}[0pt]{ \shortstack [c] {  $83.3$ \\ ($-$)}} & \footnotesize \raisebox{-3ex}[0pt]{ \shortstack [c] {  $87.2$ \\ ($-$)}} & \footnotesize Image patches are obtained using a uniform square mesh, as shown in Figure \ref{MultiplePass} (a). Patch features are encoded into BoW descriptors.\\   \cline{2-11} 

%%%%

& \footnotesize \raisebox{-1.7ex}[0pt]{ \shortstack [c] {  DeepIndex \cite{liu2015deepindex}} } &  \footnotesize	\raisebox{-2.7ex}[0pt]{ \shortstack [c] { AlexNet \\  $\bullet$  VGG19 }} & \footnotesize	\raisebox{-2.7ex}[0pt]{ \shortstack [c] { FC6-7 $\bullet$\\ FC17-18 }}	& \footnotesize \raisebox{-2.7ex}[0pt]{\shortstack [c] {BoW + \\  PCA }}   & \footnotesize \raisebox{-2ex}[0pt]{ 512} & \footnotesize \raisebox{-2ex}[0pt]{ 81.7} & \footnotesize \raisebox{-2.7ex}[0pt]{\shortstack [c] { $ 3.32 $ \\ \textcolor[rgb]{0,0,0}{(N-S)}}}  & \footnotesize	\raisebox{-2.0ex}[0pt]{\shortstack [c] {$-$}} & \footnotesize \raisebox{-2.7ex}[0pt]{ \shortstack [c] { $75.4$\\ ($-$)}} & \footnotesize Exploring layer-level and model-level fusion methods. Image patches are extracted using spatial pyramid modeling, as shown in Figure \ref{MultiplePass} (b).\\   \cline{2-11}

%%%
& \footnotesize \raisebox{-2ex}[0pt]{ \shortstack [c] {  CCS \cite{yan2016cnn}} } & \footnotesize \raisebox{-2ex}[0pt]{\shortstack [c] { GoogLeNet }} & \footnotesize	\raisebox{-2ex}[0pt]{ \shortstack [c] { Conv}}	& \footnotesize \raisebox{-2.7ex}[0pt]{\shortstack [c] { VLAD \\   +  PCA$_{w} $ }}   & \footnotesize \raisebox{-2ex}[0pt]{ 128} & \footnotesize \raisebox{-2ex}[0pt]{ 84.1} & \footnotesize \raisebox{-3ex}[0pt]{\shortstack [c] { $ 3.81 $ \\ \textcolor[rgb]{0,0,0}{(N-S)}}}  & \footnotesize \raisebox{-3ex}[0pt]{ \shortstack [c] { $64.8$\\ ($-$)}} & \footnotesize \raisebox{-3ex}[0pt]{ \shortstack [c] { $76.8$\\ ($-$)}} & \footnotesize Object proposals are extracted by RPNs, as shown in Figure \ref{MultiplePass} (d). Object level and point level feature concatenation schemes are explored.\\   \cline{2-11} 
\Xhline{1.0pt}

\end{tabular}
}
\vspace{-0.5em}
\end{table*}

%%%%%%%%%%%%%%%%%%%%%%%%%%%%%%%%%%%%%
%%%%%%%%%%%%%%%%%%%%%%%%%%%%%%%%%%%%%
%%%%%%%%%% Second table of SOTA%%%%%%
%%%%%%%%%%%%%%%%%%%%%%%%%%%%%%%%%%%%%
%%%%%%%%%%%%%%%%%%%%%%%%%%%%%%%%%%%%%
%%%%%%%%%%%%%%%%%%%%%%%%%%%%%%%%%%%%%

\begin{table*}[!t]
\centering % Center table
 \caption{ \footnotesize Performance evaluation of methods in which DCNN models are fine-tuned, in a supervised or an unsupervised manner. ``CE Loss'' means the models are fine-tuned using the classification-based loss function in the form of Eq. \ref{cross_entropy_loss}. ``Siamese Loss'' is in the form of Eq. \ref{contrastive}. ``Regression Loss'' is in the form of Eq. \ref{regressionLoss}. ``Triplet Loss'' is in the form of Eq. \ref{triplet}. }
 \vspace{-1em}
 \label{Metric_learning_table_B} 
 \renewcommand{\arraystretch}{1.0}
\resizebox{19.5cm}{!}{

\begin{tabular}{!{\vrule width1.2bp}c|p{2.0cm}|c|p{1.35cm}|c|p{1.1cm}|p{1.25cm}|c|c|p{0.7cm}|c|c|p{7.8cm}!{\vrule width1.2bp}}

\Xhline{1pt}
\footnotesize Type  & \footnotesize \shortstack [c]  { Method }  & \footnotesize \shortstack [c] {Backbone \\ DCNN} &   \footnotesize \shortstack [c] {Training \\ set}	& \footnotesize \shortstack [c] {Output \\ Layer} & \footnotesize \shortstack [c] {Embed. \\ Aggre.}	& \footnotesize \shortstack [c] { Loss \\ Function } & \footnotesize \shortstack [c] { Feat. \\ Dim } & \footnotesize \shortstack [c] {Holidays} & \footnotesize \shortstack [c] {UKB} & \footnotesize \shortstack [c] {Oxford5k \\ (+100k)} &  \footnotesize \shortstack [c] {Paris6k \\ (+100k)} &  \footnotesize Brief Conclusions and Highlights \\ \Xhline{1.0pt}

%%%%
\multirow{5}{*}{  \raisebox{-24ex}[0pt]{\begin{tabular}[c]{@{}c@{}}  \rotatebox{90}{Supervised Fine-tuning}\end{tabular}}}  
& \footnotesize \raisebox{-2ex}[0pt]{ \shortstack [c] { $\!\!\!\!\!\! $ Neural codes \cite{babenko2014neural}} } & \footnotesize	\raisebox{-2ex}[0pt]{ AlexNet} & \footnotesize \raisebox{-3.5ex}[0pt]{\shortstack [c] { Landmarks \\ \cite{babenko2014neural} }}  & \footnotesize \raisebox{-2ex}[0pt]{ \shortstack [l] {  FC6}}	&  \footnotesize \raisebox{-2ex}[0pt]{\shortstack [c] {  PCA }} & \footnotesize	\raisebox{-2ex}[0pt]{ \shortstack [c] { CE Loss}}  & \footnotesize \raisebox{-2ex}[0pt]{ 128} & \footnotesize \raisebox{-2ex}[0pt]{ 78.9 } &  \footnotesize \raisebox{-3.5ex}[0pt]{\shortstack [c] { $ 3.29 $ \\ \textcolor[rgb]{0,0,0}{(N-S)}}} & \footnotesize \raisebox{-3.5ex}[0pt]{ \shortstack [c] { $55.7$\\ (52.3)}} & \footnotesize \raisebox{-2.0ex}[0pt]{\shortstack [c] {$-$}} & \footnotesize The first work which fine-tunes deep networks for IIR. Compressed neural codes and different layers are explored. \\  \cline{2-13}

%%%
& \footnotesize \raisebox{-2ex}[0pt]{ \shortstack [c] {  Nonmetric \cite{garcia2017learning}} } & \footnotesize	\raisebox{-2ex}[0pt]{VGG16} & \footnotesize	\raisebox{-2ex}[0pt]{ $\!\!\! $ Landmarks }  & \footnotesize \raisebox{-2ex}[0pt]{ \shortstack [l] { Conv5}}	&  \footnotesize \raisebox{-2ex}[0pt]{\shortstack [c] { PCA$_w$ }} & \footnotesize	\raisebox{-3.2ex}[0pt]{ \shortstack [c] { $\!\!\!\! $ Regression \\ $\!\!\!\! $ Loss}}  & \footnotesize \raisebox{-2ex}[0pt]{512} &  \footnotesize \raisebox{-2ex}[0pt]{ \shortstack [c] {$-$} } & \footnotesize	\raisebox{-2ex}[0pt]{\shortstack [c] {$-$}} & \footnotesize \raisebox{-3.5ex}[0pt]{ \shortstack [c] { $88.2$\\ (82.1)}} &  \footnotesize \raisebox{-3.5ex}[0pt]{ \shortstack [c] {  $88.2$\\ (82.9)}} & \footnotesize Similarity learning of similar and dissimilar pairs is performed by a neural network, optimized using regression loss. \\   \cline{2-13}

%%%
& \footnotesize \raisebox{-2ex}[0pt]{ \shortstack [c] { SIAM-FV  \cite{ong2017siamese}} } & \footnotesize	\raisebox{-2ex}[0pt]{ VGG16}  & \footnotesize	\raisebox{-2ex}[0pt]{ $\!\!\!\! $ Landmarks } & \footnotesize \raisebox{-2ex}[0pt]{ \shortstack [l] {  Conv5}}	&  \footnotesize	\raisebox{-2.7ex}[0pt]{ \shortstack [c] {  FV + \\ PCA$_w$ }} & \footnotesize	\raisebox{-3ex}[0pt]{ \shortstack [c] { Siamese \\  Loss}} & \footnotesize \raisebox{-2ex}[0pt]{ 512} &  \footnotesize \raisebox{-2ex}[0pt]{ \shortstack [c] {$-$} } & \footnotesize	\raisebox{-2ex}[0pt]{\shortstack [c] {$-$}} & \footnotesize \raisebox{-3ex}[0pt]{ \shortstack [c] {  $81.5$\\ (76.6)}} &  \footnotesize \raisebox{-3ex}[0pt]{ \shortstack [c] {  $82.4$\\ ($-$)}} & \footnotesize Fisher Vector is integrated on top of VGG and is trained with VGG simultaneously. \\   \cline{2-13} 

%%%
& \footnotesize \raisebox{-3ex}[0pt]{ \shortstack [c] {  Faster \\  R-CNN  \cite{salvador2016faster}} } & \footnotesize	\raisebox{-2ex}[0pt]{ VGG16} & \footnotesize \raisebox{-3ex}[0pt]{ \shortstack [c] {  Oxford5k \\  Paris6k } } & \footnotesize \raisebox{-2ex}[0pt]{ \shortstack [l] { Conv5}}	&  \footnotesize \raisebox{-2ex}[0pt]{\shortstack [c] {  MP / SP }} & \footnotesize	\raisebox{-3.2ex}[0pt]{ \shortstack [c] { $\!\!\!\! $ Regression \\ $\!\!\!\! $ Loss}}  & \footnotesize \raisebox{-2ex}[0pt]{ 512} &  \footnotesize \raisebox{-2ex}[0pt]{ \shortstack [c] {$-$} } & \footnotesize	\raisebox{-2ex}[0pt]{\shortstack [c] {$-$}} & \footnotesize \raisebox{-3.5ex}[0pt]{ \shortstack [c] {  $75.1$\\ $(-)$}} &  \footnotesize \raisebox{-3.5ex}[0pt]{ \shortstack [c] { $80.7$\\ ($-$)}} & \footnotesize RPN is fine-tuned, based on bounding box coordinates and class scores for specific region query which is region-targeted. \\  \cline{2-13}

%%%
& \footnotesize \raisebox{-2ex}[0pt]{ \shortstack [c] {  SIFT-CNN \cite{lv2018retrieval}} } & \footnotesize	\raisebox{-2ex}[0pt]{VGG16} &  \footnotesize	\raisebox{-3ex}[0pt]{ \shortstack [c] { Holidays \\  UKB }} & \footnotesize \raisebox{-2ex}[0pt]{ \shortstack [l] {  Conv5}}	&  \footnotesize \raisebox{-2ex}[0pt]{ \shortstack [c] {  SP }} & \footnotesize	\raisebox{-3ex}[0pt]{ \shortstack [c] { Siamese \\  Loss}}  & \footnotesize \raisebox{-2ex}[0pt]{ 512} &  \footnotesize \raisebox{-2ex}[0pt]{ 88.4 } &  \footnotesize \raisebox{-3.5ex}[0pt]{\shortstack [c] { $ 3.91 $ \\ \textcolor[rgb]{0,0,0}{(N-S)}}} &  \footnotesize	 \raisebox{-2ex}[0pt]{ \shortstack [c] {$-$}} & \footnotesize \raisebox{-2ex}[0pt]{ \shortstack [c] {$-$}} & \footnotesize SIFT features are used as supervisory information for mining positive and negative samples. \\   \cline{2-13}

 %%%
& \footnotesize \raisebox{-2ex}[0pt]{ \shortstack [c] { $\!\!\!\!\! $  Quartet-Net \cite{cao2016quartet}} } & \footnotesize	\raisebox{-2ex}[0pt]{ VGG16} &  \footnotesize	\raisebox{-3.5ex}[0pt]{ \shortstack [c] { GeoPair \\  \cite{cao2016quartet} }} & \footnotesize \raisebox{-2ex}[0pt]{ \shortstack [l] {  FC6}}	&  \footnotesize \raisebox{-2ex}[0pt]{ \shortstack [c] {  PCA }} & \footnotesize	\raisebox{-3ex}[0pt]{ \shortstack [c] { Siamese \\  Loss}}  & \footnotesize \raisebox{-2ex}[0pt]{ 128} &  \footnotesize \raisebox{-2ex}[0pt]{71.2 } &  \footnotesize \raisebox{-3.5ex}[0pt]{\shortstack [c] { $ 87.5 $ \\ \textcolor[rgb]{0,0,0}{(mAP)}}} & \footnotesize \raisebox{-3.5ex}[0pt]{ \shortstack [c] {  $48.5$\\ ($-$)}} &  \footnotesize \raisebox{-3.5ex}[0pt]{ \shortstack [c] {$48.8$\\($-$)}} & \footnotesize Quartet-net learning is explored to improve feature discrimination where double-margin contrastive loss is used. \\   \cline{2-13}

%%%
& \footnotesize \raisebox{-2ex}[0pt]{ \shortstack [c] {  NetVLAD \cite{arandjelovic2016netvlad}} } & \footnotesize	\raisebox{-2ex}[0pt]{ VGG16} & \footnotesize	\raisebox{-3ex}[0pt]{ \shortstack [c] { $\!\!\!\!\! $ Tokyo Time \\  Machine }}  & \footnotesize	\raisebox{-3ex}[0pt]{ \shortstack [c] {  VLAD \\  Layer }}	&  \footnotesize \raisebox{-2ex}[0pt]{ \shortstack [c] { PCA$_{w}$ }} & \footnotesize	\raisebox{-3ex}[0pt]{ \shortstack [c] { Triplet \\  Loss}}  & \footnotesize \raisebox{-2ex}[0pt]{ 256} &  \footnotesize \raisebox{-2ex}[0pt]{  79.9 } &  \footnotesize	\raisebox{-2ex}[0pt]{\shortstack [c] {$-$}} & \footnotesize \raisebox{-3ex}[0pt]{ \shortstack [c] {  $62.5$\\ ($-$)}} &  \footnotesize \raisebox{-3ex}[0pt]{ \shortstack [c] {  $72.0$\\ ($-$)}} & \footnotesize VLAD is integrated at the last convolutional layer of VGG16 network as a plugged layer. \\   \cline{2-13} \textbf{}

%%%
& \footnotesize \raisebox{-3ex}[0pt]{ \shortstack [c] {  Deep \\ Retrieval \cite{gordo2017end}} } & \footnotesize	\raisebox{-2ex}[0pt]{ ResNet-101} &  \footnotesize	\raisebox{-2ex}[0pt]{ $\!\!\!\!\! $ Landmarks } & \footnotesize	\raisebox{-3ex}[0pt]{ \shortstack [c] {  Conv5 \\  Block }}	&  \footnotesize	\raisebox{-3ex}[0pt]{ \shortstack [c] {  MP + \\  PCA$_w$ }} & \footnotesize	\raisebox{-3ex}[0pt]{ \shortstack [c] { Triplet \\Loss}}  & \footnotesize \raisebox{-2ex}[0pt]{ 2048} &  \footnotesize \raisebox{-2ex}[0pt]{  90.3 } &  \footnotesize	\raisebox{-2ex}[0pt]{\shortstack [c] {$-$}} & \footnotesize \raisebox{-3.5ex}[0pt]{ \shortstack [c] {  $86.1$\\ (82.8)}} &  \footnotesize \raisebox{-3.5ex}[0pt]{ \shortstack [c] {  $94.5$\\ (90.6)}} & \footnotesize Dataset is cleaned automatically. Features are encoded by R-MAC. RPN is used to extract the most relevant regions. \\  \cline{2-13}

& \footnotesize \raisebox{-2ex}[0pt]{ \shortstack [c] {  DELF \cite{noh2017largescale}} } & \footnotesize	\raisebox{-2ex}[0pt]{ ResNet-101}  & \footnotesize	\raisebox{-2ex}[0pt]{ GLDv1 } & \footnotesize	\raisebox{-3ex}[0pt]{ \shortstack [c] {  Conv4 \\ Block }}	& \footnotesize	\raisebox{-2.6ex}[0pt]{ \shortstack [c] { $\!\!\!\! $ Attention \\ + PCA$_w$ }} & \footnotesize	\raisebox{-2ex}[0pt]{ \shortstack [c] { CE Loss}}  & \footnotesize \raisebox{-2ex}[0pt]{ 2048} & \footnotesize \raisebox{-2ex}[0pt]{$-$} &  \footnotesize \raisebox{-2ex}[0pt]{$-$} & \footnotesize \raisebox{-3ex}[0pt]{ \shortstack [c] { $83.8$\\ (82.6)}} & \footnotesize \raisebox{-3ex}[0pt]{ \shortstack [c] { $85.0$\\ (81.7)}} & \footnotesize Exploring the FCN to extract region-level features and construct feature pyramids of different sizes. \\   \cline{2-13}  

 \Xhline{1.0pt}

%%%% for unsupervised method
%%%%
\multirow{7}{*}{ \raisebox{-13ex}[0pt]{\begin{tabular}[c]{@{}c@{}}  \rotatebox{90}{Unsupervised Fine-tuning} \end{tabular}}}
& \footnotesize \raisebox{-2ex}[0pt]{ \shortstack [c] {  MoM \cite{iscen2018mining}} } &  \footnotesize	\raisebox{-2ex}[0pt]{ \shortstack [c] {  VGG16  }} & \footnotesize	\raisebox{-2ex}[0pt]{ Flickr 7M } & \footnotesize \raisebox{-2ex}[0pt]{ \shortstack [l] {  Conv5}}	& \footnotesize \raisebox{-3ex}[0pt]{\shortstack [c] { MP + \\  PCA$_{w} $ }} & \footnotesize	\raisebox{-3ex}[0pt]{ \shortstack [c] { Siamese \\  Loss}} & \footnotesize \raisebox{-2ex}[0pt]{ 64} & \footnotesize \raisebox{-2ex}[0pt]{ 87.5} & \footnotesize	\raisebox{-2ex}[0pt]{\shortstack [c] {$-$}}  & \footnotesize \raisebox{-3.5ex}[0pt]{ \shortstack [c] {  $78.2$\\ (72.6)}} & \footnotesize \raisebox{-3.5ex}[0pt]{ \shortstack [c] {  $85.1$\\ (78.0)}} & \footnotesize Exploring manifold learning for mining dis/similar samples. Features are tested globally and regionally.\\  \cline{2-13}

%%%%
& \footnotesize \raisebox{-2ex}[0pt]{ \shortstack [c] { GeM \cite{radenovic2018fine}} } &  \footnotesize	\raisebox{-2ex}[0pt]{ \shortstack [c] {  VGG16  }}  & \footnotesize	\raisebox{-2ex}[0pt]{ Flickr 7M } & \footnotesize \raisebox{-2ex}[0pt]{ \shortstack [l] {  Conv5}}	& \footnotesize \raisebox{-3ex}[0pt]{\shortstack [c] { GeM  \\  Pooling }} & \footnotesize	\raisebox{-3ex}[0pt]{ \shortstack [c] { Siamese \\  Loss}} & \footnotesize \raisebox{-2ex}[0pt]{ 512} & \footnotesize \raisebox{-2ex}[0pt]{ 83.1} & \footnotesize	\raisebox{-2ex}[0pt]{\shortstack [c] {$-$}}  & \footnotesize \raisebox{-3.5ex}[0pt]{ \shortstack [c] { $82.0$\\ (76.9)}} & \footnotesize \raisebox{-3.5ex}[0pt]{ \shortstack [c] { $79.7$\\ (72.6)}} & \footnotesize Fine-tuning CNNs on an unordered dataset. Samples are selected from an automated 3D reconstruction system.  \\ \cline{2-13}

%%%%
& \footnotesize \raisebox{-2ex}[0pt]{ \shortstack [c] {  SfM-CNN  \cite{radenovic2016cnn}} } &  \footnotesize	\raisebox{-2ex}[0pt]{ \shortstack [c] {  VGG16  }} &  \footnotesize	\raisebox{-2ex}[0pt]{ Flickr 7M } & \footnotesize \raisebox{-2ex}[0pt]{ \shortstack [l] {Conv5}}	& \footnotesize \raisebox{-2ex}[0pt]{ \shortstack [c] { PCA$_{w}$ }}  & \footnotesize	\raisebox{-3ex}[0pt]{ \shortstack [c] { Siamese \\  Loss}} & \footnotesize \raisebox{-2ex}[0pt]{512} & \footnotesize \raisebox{-2ex}[0pt]{  82.5} & \footnotesize	\raisebox{-2ex}[0pt]{\shortstack [c] {$-$}}  & \footnotesize \raisebox{-3.5ex}[0pt]{ \shortstack [c] {  $77.0$\\ (69.2)}} & \footnotesize \raisebox{-3.5ex}[0pt]{ \shortstack [c] {  $83.8$\\ (76.4)}} & \footnotesize  Employing Structure-from-Motion to select positive and negative samples from unordered images.  \\   \cline{2-13} 

%%%%
& \footnotesize \raisebox{-2ex}[0pt]{ \shortstack [c] {  MDP-CNN  \cite{zhaomodelling2018}} } &  \footnotesize	\raisebox{-2ex}[0pt]{ ResNet-101} &  \footnotesize	\raisebox{-2ex}[0pt]{ $\!\!\!\!\! $ Landmarks } & \footnotesize	\raisebox{-3ex}[0pt]{ \shortstack [c] {  Conv5 \\ Block }}	& \footnotesize \raisebox{-2ex}[0pt]{ \shortstack [c] {  SP }}  & \footnotesize	\raisebox{-3ex}[0pt]{ \shortstack [c] { Triplet \\  Loss}}  & \footnotesize \raisebox{-2ex}[0pt]{ 2048} &  \footnotesize \raisebox{-2ex}[0pt]{ \shortstack [c] {$-$} }   & \footnotesize	\raisebox{-2ex}[0pt]{\shortstack [c] {$-$}}  & \footnotesize \raisebox{-3.5ex}[0pt]{ \shortstack [c] {  $85.4$\\ (85.1)}} & \footnotesize \raisebox{-3.5ex}[0pt]{ \shortstack [c] {  $96.3$\\ (94.7)}} & \footnotesize Exploring global feature structure by modeling the manifold learning to select positive and negative pairs.   \\   
\cline{2-13}  

%%%%
& \footnotesize \raisebox{-2ex}[0pt]{ \shortstack [c] {  IME-CNN \cite{xu2017iterative}} } &  \footnotesize	\raisebox{-2ex}[0pt]{ ResNet-101} &  \footnotesize	\raisebox{-3ex}[0pt]{ \shortstack [c] { $\!\!\!\! $ Oxford105k \\  Paris106k }}  & \footnotesize	\raisebox{-3ex}[0pt]{ \shortstack [c] {  IME \\  Layer }}	& \footnotesize \raisebox{-2ex}[0pt]{ \shortstack [c] {  MP }}  & \footnotesize	\raisebox{-3ex}[0pt]{ \shortstack [c] { $\!\!\!\! $ Regression \\ $\!\!\!\!\! $ Loss}} & \footnotesize \raisebox{-2ex}[0pt]{ 2048} &  \footnotesize \raisebox{-2ex}[0pt]{ \shortstack [c] {$-$} }   & \footnotesize	\raisebox{-2ex}[0pt]{\shortstack [c] {$-$}}  & \footnotesize \raisebox{-3.5ex}[0pt]{ \shortstack [c] {  $92.0$\\ (87.2)}} & \footnotesize \raisebox{-3.5ex}[0pt]{ \shortstack [c] {  $96.6$\\ (93.3)}} & \footnotesize Graph-based manifold learning is explored to mine the matching and non-matching pairs in unordered datasets.  \\   \cline{2-13} 
\Xhline{1.0pt}
\end{tabular}
}
\end{table*}

\textbf{Evaluation for single feedforward pass.} \textcolor{black}{In general, we observe that fully-connected layers used as feature extractors may give a lower accuracy (\eg 74.7\% on Holidays in \cite{babenko2014neural}), compared to using convolutional layers in Table \ref{Table_retrieval_off_the_shelf}. For the case where the same VGG net is used, the way to embed or aggregate features is critical. The methods shown in Figure \ref{SinglePass} improve the discrimination of convolutional feature maps and perform differently in Table \ref{Table_retrieval_off_the_shelf}, 66.9\% of R-MAC \cite{Lostphilbin2008lost1} and 58.9\% of SPoC \cite{babenko2015aggregating} on Oxford-5k, differences which we see as critical factors for further analysis. If embedded by a BoW model, the results are competitive on Oxford-5k and Paris-6k (73.9\% and 82.0\%, respectively), while its codebook size is 25k, which may affect retrieval efficiency. Moreover, layer-level feature fusion improves retrieval accuracy. Yu \etal \cite{yu2017exploiting} combine three layers (mAP of 91.4\% on Holidays), outperforming the performance of non-fusion method \cite{babenko2015aggregating} (mAP of 80.2\%).}

\textbf{Evaluation for multiple feedforward pass.}
Results for the methods of Figure \ref{MultiplePass} are reported in Table \ref{Table_retrieval_off_the_shelf}. Among them, extracting image patches densely using VGG \cite{simonyan2014very} has the highest performance on the 4 datasets \cite{sharif2014cnn}, and rigid grid with BoW encoding \cite{cao2017local} is competitive (mAP of 87.2\% on Paris-6k). These two methods consider more patches, even background information, when used for feature extraction. Instead of generating patches densely, region proposals and spatial pyramid modeling introduce a degree of purpose and efficiency in processing image objects. Spatial information is better maintained using multiple-pass schemes than with single-pass. For example, a shallower network (AlexNet) and region proposal networks in \cite{reddy2015object} have a UKBench N-Score of 3.81, higher than using deeper networks \cite{babenko2015aggregating},\cite{babenko2014neural},\cite{yu2017exploiting}. Besides feeding image patches into the same network, model-level fusion also exploits complementary spatial information to improve accuracy. For instance, as reported in \cite{liu2015deepindex}, which combines AlexNet and VGG, the results on Holidays (81.7\% of mAP) and UKBench (3.32 of N-Score) are better than these in \cite{li2016exploiting} (76.75\% and 3.00, respectively).

\textbf{Evaluation for supervised fine-tuning.} Compared to off-the-shelf models, fine-tuning deep networks usually improves accuracy, see Table \ref{Metric_learning_table_B}. For instance, the result on Oxford-5k \cite{tolias2015particular} by using a pre-trained VGG is improved from 66.9\% to 81.5\% in \cite{ong2017siamese} when a single-margin Siamese loss is used. Similar trends can be also observed on the Paris-6k dataset. For classification-based fine-tuning, its performance may be improved by using powerful DCNNs and feature enhancement methods such as the attention mechanism in \cite{noh2017largescale}, with an mAP increased from 55.7\% in \cite{babenko2014neural} to 83.8\% in \cite{noh2017largescale} on Oxford-5k. As for pairwise ranking loss fine-tuning, in some cases the loss used for fine-tuning is essential for performance improvement. For example, RPN is re-trained using regression loss on Oxford-5k and Paris-6k (75.1\% and 80.7\%, respectively) \cite{salvador2016faster}. Its results are lower than the results from \cite{garcia2017learning} (88.2\% and 88.2\%, respectively) where a transformation matrix is used to learn visual similarity. However, when RPN is trained by using triplet loss such as \cite{gordo2017end}, the effectiveness of retrieval is improved significantly where the results are 86.1\% (on Oxford-5k) and 94.5\% (on Paris-6k). Feature embedding methods are important for retrieval accuracy; Ong \etal \cite{ong2017siamese} embedded \textit{Conv5} feature maps by Fisher Vector and achieved an mAP of 81.5\% on Oxford-5k, while embedding feature maps by using VLAD achieves an mAP of 62.5\% on this dataset \cite{arandjelovic2016netvlad},\cite{radenovic2016cnn}.

\textbf{Evaluation for unsupervised fine-tuning.} 
Compared to supervised fine-tuning, unsupervised fine-tuning methods are relatively less explored. The difficulty for unsupervised fine-tuning is to mine sample relevance without ground-truth labels. In general, unsupervised fine-tuning methods should be expected to have lower performance than supervised. For instance, supervised fine-tuning using Siamese loss \cite{lv2018retrieval} achieves an mAP 88.4\% on Holidays, while unsupervised fine-tuning using the same loss function in \cite{radenovic2016cnn},\cite{radenovic2018fine},\cite{iscen2018mining} achieves 82.5\%, 83.1\%, and 87.5\%, respectively. However, unsupervised fine-tuning methods can achieve a similar accuracy, even outperform the supervised fine-tuning, if a suitable feature embedding method is used. For instance, Zhao \etal \cite{zhaomodelling2018} explore global feature structure modeling the manifold learning, producing an mAP of 85.4\% (on Oxford-5k) and 96.3\% (on Paris-6k), which is similar to supervised results \cite{gordo2017end} of 86.1\% (on Oxford-5k) and 94.5\% (on Paris-6k). As another example, the precision of ResNet-101 fine-tuned by cross entropy loss achieves 83.8\% on Oxford-5k \cite{noh2017largescale}, while the precision is further improved to 92.0\% when an IME layer is used to embed features and fine-tuned in an unsupervised way \cite{xu2017iterative}. Note that fine-tuning strategies are related to the type of the target retrieval datasets. As demonstrated in Table \ref{GLD_evaluation_on_training_dataset} and \cite{cao2020unifying}, fine-tuning on different datasets may produce a different final retrieval performance.

\begin{figure*}[!t]
\centering  
 {
  \includegraphics[width=\textwidth]{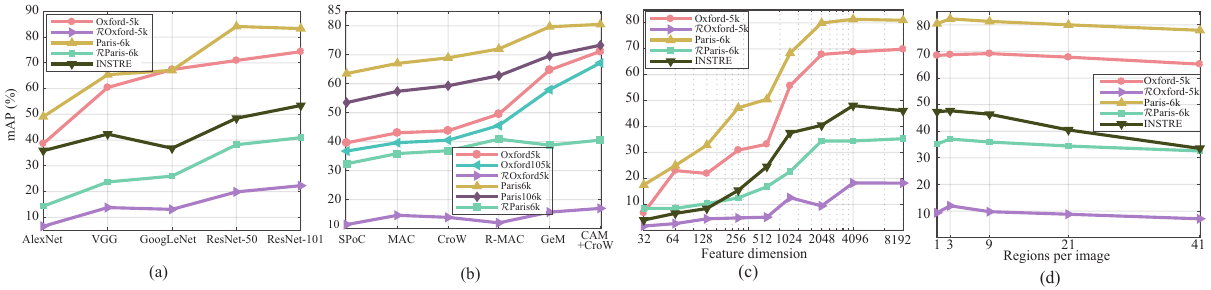}  
 }     
  \vspace{-2em}
  \caption{(a) The effectiveness of different DCNNs; (b) Comparison of the feature aggregation methods in Figure \ref{SinglePass}; (c) The impact of global feature dimension by using ResNet-50; (d) Performance comparison when aggregating different numbers of image regions.} 
  \label{CNNandaggregation_depth_dim_dataset}    
 \end{figure*}

\textbf{Network depth.} We compare the efficacy of DCNNs by depth, following the fine-tuning protocols\footnote{https://github.com/filipradenovic/cnnimageretrieval-pytorch} in \cite{radenovic2018fine}. For fair comparisons, all convolutional features from these backbone DCNNs are aggregated by MAC \cite{razavian2016visual}, and fine-tuned by using the same loss function with the same learning rate, thus the adopted methods are the same except for the DCNN depth. We use the default feature dimension (\emph{i.e.} AlexNet (256), VGG (512), GoogLeNet (1024), ResNet-50/101 (2048)). The results are reported in Figure \ref{CNNandaggregation_depth_dim_dataset} (a). We observe that the deeper networks consistently lead to better accuracy due to extracting more discriminative features.

\textbf{Feature aggregation methods.} The methods of embedding convolutional feature maps were illustrated in Figure \ref{SinglePass}. We use the off-the-shelf VGG (without updating parameters) on the Oxford and Paris datasets. The results are reported in Figure \ref{CNNandaggregation_depth_dim_dataset} (b). We observe that the different ways to aggregate the same off-the-shelf DCNN leads to differences in retrieval performance. These reported results provide a reference for feature aggregation when one uses convolutional layers for performing retrieval tasks.

\textbf{Global feature dimension.} We add fully-connected layers on the top of pooled convolutional features of ResNet-50 to obtain global descriptors with their dimensions varying from 32 to 8192. The results of 5 datasets are shown in Figure \ref{CNNandaggregation_depth_dim_dataset} (c). It is expected that higher-dimension features usually capture more semantics and are helpful for retrieval. The performance
tends to be stable when the dimension is very large.

\textcolor{black}{\textbf{Number of image regions.} We compare the retrieval performance when different number of regions are fed and other components are kept the same, as depicted in Figure \ref{CNNandaggregation_depth_dim_dataset} (d). Convolutional features of each region are pooled as 2048-dim regional features by MAC and then aggregated into a global one. Note that the final memory requirement is identical for the case that a holistic image is used as input (\ie regarded as the case where only one region is used). Regional inputs on an image are extracted with a 40\% overlap of neighboring regions and the number varying from 1 to 41. For Oxford-5k, the best result is given by the case where 9 image regions are used. For the rest datasets, 3 image regions give the best results. Finally, more regions extracted from one image decline the retrieval mAP. A reason is that features of background or irrelevant regions have also been aggregated, and negatively affect the performance.}

\begin{table}[t]
\caption{Evaluations of training sets and retrieval reranking. Numerical results are cited from \cite{cao2020unifying}.
}
\vspace{-1em}
\label{GLD_evaluation_on_training_dataset}
\footnotesize
\setlength{\tabcolsep}{0.6mm}
\begin{tabular}{|cc|c|c|cc|c|c|c|}
\hline
\multicolumn{2}{|c|}{\multirow{3}{*}{Conditions}} 
 & \multirow{3}{*}{Global} & \multirow{3}{*}{\begin{tabular}[c]{@{}c@{}}Local\\ reranking\end{tabular}} & \multicolumn{2}{c|}{Training set}   & \multirow{3}{*}{$\mathcal{R}$Oxf} & \multirow{3}{*}{$\mathcal{R}$Par} & \multirow{3}{*}{\begin{tabular}[c]{@{}c@{}}GLD-v2\\ testing\end{tabular}} \\ \cline{5-6}
 
 &  &  &   & \multicolumn{1}{c|}{GLD-v1} & \begin{tabular}[c]{@{}c@{}}GLD-v2\\ -clean\end{tabular} &  &  &    \\ \hline
\multicolumn{1}{|c|}{\multirow{4}{*}{\begin{tabular}[c]{@{}c@{}}ResNet\\ -50\end{tabular}}} &  Case 1 &  \tickYes  &  \tickNo  & \multicolumn{1}{c|}{\tickYes}   & \tickNo  &  45.1   & 63.4 &  20.4 \\ \cline{2-9} 
\multicolumn{1}{|c|}{} & Case 2  & \tickYes  & \tickNo  & \multicolumn{1}{c|}{\tickNo} & \tickYes  &  51.0  & 71.5 &  24.1  \\ \cline{2-9} 
\multicolumn{1}{|c|}{} & Case 3 &  \tickYes   & \tickYes  & \multicolumn{1}{c|}{\tickYes}   & \tickNo  &  54.2  &  64.9  &  22.3 \\ \cline{2-9} 
  %%%
\multicolumn{1}{|c|}{}& Case 4 &  \tickYes  &  \tickYes & \multicolumn{1}{c|}{\tickNo}    &   \tickYes  &  57.9   &  71.0  & 24.3 \\ \hline

\multicolumn{1}{|c|}{\multirow{4}{*}{\begin{tabular}[c]{@{}c@{}}ResNet\\ -101 \end{tabular}}} & Case 5 &  \tickYes  &  \tickNo  & \multicolumn{1}{c|}{\tickYes}   & \tickNo  &  51.2  &  64.7 &  21.7  \\ \cline{2-9} 
\multicolumn{1}{|c|}{} & Case 6 & \tickYes  & \tickNo  & \multicolumn{1}{c|}{\tickNo} & \tickYes  &  55.6  & 72.4 &  26.0  \\ \cline{2-9} 
\multicolumn{1}{|c|}{} & Case 7  &  \tickYes   & \tickYes  & \multicolumn{1}{c|}{\tickYes}   & \tickNo  &  59.3  &  65.5 &  24.3 \\ \cline{2-9} 
  %%%
\multicolumn{1}{|c|}{} & Case 8 &  \tickYes  &  \tickYes & \multicolumn{1}{c|}{\tickNo}    &   \tickYes  &   64.0  &   72.8 &  26.8 \\ \hline
\end{tabular}
\end{table}

\textbf{Fine-tuning datasets and retrieval reranking.} We compare performance on $\mathcal{R}$Oxford-5k, $\mathcal{R}$Paris-6k, and GLD-v2, aiming at comparing the role of different fine-tuning training sets and the effectiveness of retrieval reranking. Table \ref{GLD_evaluation_on_training_dataset} lists 8 experimental scenarios using two network backbones, as in \cite{cao2020unifying}. 

\textcolor{black}{Since GLD-v2 provides class-level ground-truth, its including images show large context diversity and may pose challenges to the network fine-tuning. Thus, the pre-processing steps, as proposed in \cite{yokoo2020two},\cite{weyand2020google}, are necessary to select the more coherent images, referring to the GLD-v2-clean subset. As a result, when using the global features only, this cleaned version of the training set improves the performance, as observed in Cases 1/5 and Cases 2/6 for ResNet-50/ResNet-101, respectively.} As an important postprocessing strategy, reranking further boosts the accuracy after the initial filtering step by using global features.

% \textbf{Whitening?.}

%  \begin{figure}[!t]
% \centering  
%  {
%   \includegraphics[width=0.5\textwidth]{./Figures/CNNandaggregation.pdf}  
%  }     
%   \vspace{-2em}
%   \caption{(a) The effectiveness of different DCNNs on 5 datasets; (b) Comparison of the feature aggregation methods in Figure \ref{SinglePass}. }  
%   \label{CNNandaggregation}    
%  \end{figure}

%  \begin{figure}[!t]
% \centering  
%  {
%   \includegraphics[width=0.5\textwidth]{./Figures/depth_dim_dataset.pdf}  
%  }     
%   \vspace{-2em}
%   \caption{(a) The impact of global feature dimension by using ResNet-50; (b) Regional search evaluation. }  
%   \label{Feature_dim_region}    
%  \end{figure}

 %%%%%%single column
% \begin{figure*}
%      \centering
%      \begin{subfigure}[b]{0.495\textwidth}
%          \centering
%          \includegraphics[width=\textwidth]{./Figures/depth_dim_dataset.pdf}
%          \caption{$y=x$}
%          \label{fig:y equals x}
%      \end{subfigure}
%      \hfill
%      \begin{subfigure}[b]{0.495\textwidth}
%          \centering
%          \includegraphics[width=\textwidth]{./Figures/depth_dim_dataset.pdf}
%          \caption{$y=3sinx$}
%          \label{fig:three sin x}
%      \end{subfigure}
% \end{figure*}

\section{Conclusions and Outlooks}
\label{Conclusion_and_Future_Directions}

\textcolor{black}{As a comprehensive yet timely survey on instance retrieval using deep learning, this paper has discussed the main challenges, presented a  taxonomy of recent developments according to their roles in implementing instance retrieval,  highlighted the recent representative methods and analyzed their merits and demerits,  discussed the datasets, evaluation protocols, and SOTA performance. Nowadays the exponentially increasing amount of image and video data due to surveillance, e-commerce, medical images, handheld devices, robotics, \etc, offers an endless potential for applications of instance retrieval. 
Although significant progress has been made, as discussed in Section \ref{Keychallenges}, the main challenges in  instance retrieval have not been fully addressed. Below we identify a number of promising directions for
future research. }

\textcolor{black}{\textbf{(1) Accurate and robust feature representations.} 
One of the main challenges in instance retrieval is the large intra-class variations due to changes in viewpoint, scale,  illumination, weather condition and background clutter \emph{etc.}, as we discussed in Section~\ref{Keychallenges}. However, DCNN representations have very little invariance, even though trained with lots of augmented data \cite{radenovic2018revisiting}. Fortunately, before deep learning, with instance retrieval there are lots of important ideas in handling such intra-class variations like local interest point detectors and local invariant descriptors. Therefore, it is worth enabling DCNN to learn more accurate and robust representations via leveraging such traditional wisdom to design better DCNNs. In addition, unlike most objects in existing benchmarks which are rigid, planar and textured, textureless objects, 3D objects, transparent objects, reflective surfaces, \etc  are still very hard to find.}

\textcolor{black}{In addition, pursuing accuracy alone is not sufficient, as instance retrieval systems should be able to resist potential adversarial attacks. 
Recently, deep networks have been proven to be fooled rather easily by adversarial examples \cite{li2019universal}, 
\ie images added with intentionally designed yet nearly imperceptible perturbations, which raises serious safety and robustness concerns. However, adversarial robustness in instance retrieval \cite{li2019universal},\cite{tolias2019targeted} has received very little attention, and should merit further effort.}

\textcolor{black}{\textbf{(2) Compact and efficient deep representations.} In instance retrieval, 
searching efficiently is as critical as searching accurately, especially for the pervasive 
mobile or wearable devices with very limited computing resources. However, existing methods adopt large scale, energy hungry DCNNs that are very difficult to be deployed in mobile devices. Hence, there has been pressing needs to develop compact, efficient, yet reusable deep representations tailored to the resource limited devices, like using binary neural networks \cite{song2017deep},\cite{zhao2017spatial},\cite{zheng2016accurate}.}

\textcolor{black}{\textbf{(3) Learning with fewer labels.} Deep learning require a large amount of high-quality labeled dataset to achieve high accuracy. The presence of labels errors or the limited amount of labeled data can greatly degrade DCNN's accuracy. However, collecting massive amounts of accurately labeled data is costly. In practical scenarios, datasets like GLDv2 \cite{noh2017largescale},\cite{weyand2020google} have long-tailed distributions, and noisy labels. Thus, to address such limitations, few shot learning \cite{triantafillou2017few}, self-supervised learning \cite{Deng2022InsCLRII}, 
imbalanced learning \cite{weyand2020google}, %semi-supervised learning \textcolor{black}{the GLDv2 ????},
 noisy label aware learning \cite{ibrahimi2022learning} \etc should be paid more attention in instance retrieval in the future. }

% It is impossible to annotate every image instance in a large-scale dataset,
%instead the retrieval model should have the ability of learning new knowledge even when many of the training images are unlabeled, \ie the domain of weakly-supervised or self-supervised learning (SSL). The recent work in \cite{Deng2022InsCLR} proposes a new SSL method for instance-level contrast, 
%nevertheless most SSL methods, such as SimCLR~\cite{GeoffreyEHinton2020ASF} and MoCo~\cite{MoCo2020}, are developed for image classification and fail to improve the performance of instance retrieval effectively.

% We would encourage new SSL methods for IIR given only limited labels, since discovering novel visual categories from a set of unlabeled images is a crucial and essential capability 
% for autonomous retrieval systems in real-world applications.
% Recent studies~\cite{DTC2019, AutoNovel2020} on novel category discovery introduce clustering objectives to generate pseudo-labels
% for unlabeled images, learning strategies which are important and practical for IIR, but which have received limited attention.

\textcolor{black}{\textbf{(4) Continual learning for instance retrieval.}
%\textcolor{black}{\textbf{(4) Open-domain instance retrieval.}
%Most IIR methods are examined within a single domain (\ie RGB-based images). However, images for a given object might very well have been collected across different domains, such as Clipart, Infograph, Painting, Quickdraw, Real and Sketch. Learning representations with cross-domain images is beneficial for improving the generalization ability of the retrieval model, %however the challenge is how to disentangle domain-independent and domain-dependent features properly, 
In specific, the current IIR methods make restrictive assumptions, such as the training data being enough and stationary, retraining from scratch being possible when new data becomes available, which is problematic in realistic conditions. Our living world is continuously varying, and in
general data distributions are often non-stationary, new data may be added, and
previously unseen classes may be encountered. Thus, continual learning plays a vital role in continuously updating the IIR systems. The key issues are how to retain and utilize the previously learned knowledge,  how to update the retrieval system as new images becomes available, and how to learn and improve over time.
}

\textbf{(5) Privacy-aware instance retrieval.} Most IIR systems concentrate on improving the accuracy or efficiency performance, and the higher performance might come at the cost of users' privacy. Therefore, in some cases, such as personalized search systems, the privacy protection problem is also an important issue to be considered. Deep models should be privacy-aware and protect users' personalized searching experience to avoid their worries about using such IIR systems.

\textbf{(6) Video instance retrieval.} Searching a specific instance in an image cannot always meet the requirements in some scenarios such as the video surveillance system in the field of searching criminals. Currently, with the rapid growth of video data, retrieving a certain object, place, or action in videos has become more and more important and highly necessary in the future. For video instance retrieval, 3D-CNNs models need to be built to learn video’s spatio-temporal representations to compute the semantic similarity of instances.

\footnotesize
\bibliographystyle{IEEEtran}
\bibliography{IEEEabrv,myreference}

\begin{thebibliography}{100}
\providecommand{\url}[1]{#1}
\csname url@samestyle\endcsname
\providecommand{\newblock}{\relax}
\providecommand{\bibinfo}[2]{#2}
\providecommand{\BIBentrySTDinterwordspacing}{\spaceskip=0pt\relax}
\providecommand{\BIBentryALTinterwordstretchfactor}{4}
\providecommand{\BIBentryALTinterwordspacing}{\spaceskip=\fontdimen2\font plus
\BIBentryALTinterwordstretchfactor\fontdimen3\font minus
  \fontdimen4\font\relax}
\providecommand{\BIBforeignlanguage}[2]{{%
\expandafter\ifx\csname l@#1\endcsname\relax
\typeout{** WARNING: IEEEtran.bst: No hyphenation pattern has been}%
\typeout{** loaded for the language `#1'. Using the pattern for}%
\typeout{** the default language instead.}%
\else
\language=\csname l@#1\endcsname
\fi
#2}}
\providecommand{\BIBdecl}{\relax}
\BIBdecl

\bibitem{smeulders2000content}
A.~W. Smeulders, M.~Worring, S.~Santini, A.~Gupta, and R.~Jain, ``Content-based
  image retrieval at the end of the early years,'' \emph{IEEE Trans. Pattern
  Anal. Mach. Intell.}, vol.~22, no.~12, pp. 1349--1380, 2000.

\bibitem{lew2006content}
M.~S. Lew, N.~Sebe, C.~Djeraba, and R.~Jain, ``Content-based multimedia
  information retrieval: State of the art and challenges,'' \emph{ACM Trans.
  Multimedia Comput. Commun. Appl.}, vol.~2, no.~1, pp. 1--19, 2006.

\bibitem{zheng2015scalable}
L.~Zheng, L.~Shen, L.~Tian, S.~Wang, J.~Wang, and Q.~Tian, ``Scalable person
  re-identification: A benchmark,'' in \emph{ICCV}, 2015, pp. 1116--1124.

\bibitem{liu2019group}
X.~Liu, S.~Zhang, X.~Wang, R.~Hong, and Q.~Tian, ``Group-group loss-based
  global-regional feature learning for vehicle re-identification,'' \emph{IEEE
  Trans. Image Process.}, vol.~29, pp. 2638--2652, 2019.

\bibitem{noh2017largescale}
H.~Noh, A.~Araujo, J.~Sim, T.~Weyand, and B.~Han, ``Largescale image retrieval
  with attentive deep local features,'' in \emph{ICCV}, 2017.

\bibitem{chaudhuri2019siamese}
U.~Chaudhuri, B.~Banerjee, and A.~Bhattacharya, ``Siamese graph convolutional
  network for content based remote sensing image retrieval,'' \emph{Comput.
  Vis. Image Underst.}, vol. 184, pp. 22--30, 2019.

\bibitem{liu2016deepfashion}
Z.~Liu, P.~Luo, S.~Qiu, X.~Wang, and X.~Tang, ``Deepfashion: Powering robust
  clothes recognition and retrieval with rich annotations,'' in \emph{CVPR},
  2016, pp. 1096--1104.

\bibitem{gordo2017beyond}
A.~Gordo and D.~Larlus, ``Beyond instance-level image retrieval: Leveraging
  captions to learn a global visual representation for semantic retrieval,'' in
  \emph{CVPR}, 2017, pp. 6589--6598.

\bibitem{barz2021content}
B.~Barz and J.~Denzler, ``Content-based image retrieval and the semantic gap in
  the deep learning era,'' in \emph{ICPR}, 2021, pp. 245--260.

\bibitem{wang2019multi}
X.~Wang, X.~Han, W.~Huang, D.~Dong, and M.~R. Scott, ``Multi-similarity loss
  with general pair weighting for deep metric learning,'' in \emph{CVPR}, 2019,
  pp. 5022--5030.

\bibitem{wang2014learning}
J.~Wang, Y.~Song, T.~Leung, C.~Rosenberg, J.~Wang, J.~Philbin, B.~Chen, and
  Y.~Wu, ``Learning fine-grained image similarity with deep ranking,'' in
  \emph{CVPR}, 2014, pp. 1386--1393.

\bibitem{babenko2015aggregating}
A.~Babenko and V.~Lempitsky, ``Aggregating local deep features for image
  retrieval,'' in \emph{ICCV}, 2015, pp. 1269--1277.

\bibitem{zheng2018sift}
L.~Zheng, Y.~Yang, and Q.~Tian, ``{SIFT} meets {CNN}: A decade survey of
  instance retrieval,'' \emph{IEEE Trans. Pattern Anal. Mach. Intell.},
  vol.~40, no.~5, pp. 1224--1244, 2018.

\bibitem{zhang2016hyperlink}
W.~Zhang, C.-W. Ngo, and X.~Cao, ``Hyperlink-aware object retrieval,''
  \emph{IEEE Trans. Image Process.}, vol.~25, no.~9, pp. 4186--4198, 2016.

\bibitem{kalantidis2016cross}
Y.~Kalantidis, C.~Mellina, and S.~Osindero, ``Cross-dimensional weighting for
  aggregated deep convolutional features,'' in \emph{ECCV}, 2016.

\bibitem{zhang2013image}
L.~Zhang and Y.~Rui, ``Image search from thousands to billions in 20 years,''
  \emph{ACM Trans. Multimedia Comput. Commun. Appl.}, vol.~9, no.~1s, p.~36,
  2013.

\bibitem{wan2014deep}
J.~Wan, D.~Wang, S.~C.~H. Hoi, P.~Wu, J.~Zhu, Y.~Zhang, and J.~Li, ``Deep
  learning for content-based image retrieval: A comprehensive study,'' in
  \emph{ACM MM}, 2014, pp. 157--166.

\bibitem{alzu2015semantic}
A.~Alzu’bi, A.~Amira, and N.~Ramzan, ``Semantic content-based image
  retrieval: A comprehensive study,'' \emph{J. Vis. Commun. Image Represent.},
  vol.~32, pp. 20--54, 2015.

\bibitem{li2016socializing}
X.~Li, T.~Uricchio, L.~Ballan, M.~Bertini, C.~G. Snoek, and A.~D. Bimbo,
  ``Socializing the semantic gap: A comparative survey on image tag assignment,
  refinement, and retrieval,'' \emph{ACM Comput. Surv. (CSUR)}, vol.~49, no.~1,
  pp. 1--39, 2016.

\bibitem{zhou2017recent}
W.~Zhou, H.~Li, and Q.~Tian, ``Recent advance in content-based image retrieval:
  A literature survey,'' \emph{arXiv preprint arXiv:1706.06064}, 2017.

\bibitem{piras2017information}
L.~Piras and G.~Giacinto, ``Information fusion in content based image
  retrieval: A comprehensive overview,'' \emph{Inf. Fusion}, vol.~37, pp.
  50--60, 2017.

\bibitem{wang2018survey}
J.~Wang, T.~Zhang, N.~Sebe, H.~T. Shen \emph{et~al.}, ``A survey on learning to
  hash,'' \emph{IEEE Trans. Pattern Anal. Mach. Intell.}, vol.~40, no.~4, pp.
  769--790, 2018.

\bibitem{lowe2004distinctive}
D.~G. Lowe, ``Distinctive image features from scale-invariant keypoints,''
  \emph{Int. J. Comput. Vis.}, vol.~60, no.~2, pp. 91--110, 2004.

\bibitem{sivic2003video}
J.~Sivic and A.~Zisserman, ``Video google: A text retrieval approach to object
  matching in videos,'' in \emph{CVPR}, 2003, pp. 1470--1477.

\bibitem{krizhevsky2012imagenet}
A.~Krizhevsky, I.~Sutskever, and G.~E. Hinton, ``Imagenet classification with
  deep convolutional neural networks,'' in \emph{NeurIPS}, 2012.

\bibitem{he2016deep}
K.~He, X.~Zhang, S.~Ren, and J.~Sun, ``Deep residual learning for image
  recognition,'' in \emph{CVPR}, 2016, pp. 770--778.

\bibitem{ren2015faster}
S.~Ren, K.~He, R.~Girshick, and J.~Sun, ``Faster {R-CNN}: Towards real-time
  object detection with region proposal networks,'' in \emph{NeurIPS}, 2015,
  pp. 91--99.

\bibitem{minaee2021image}
S.~Minaee, Y.~Y. Boykov, F.~Porikli, A.~J. Plaza, N.~Kehtarnavaz, and
  D.~Terzopoulos, ``Image segmentation using deep learning: A survey,''
  \emph{IEEE Trans. Pattern Anal. Mach. Intell.}, 2021.

\bibitem{sharif2014cnn}
A.~Sharif~Razavian, H.~Azizpour, J.~Sullivan, and S.~Carlsson, ``{CNN} features
  off-the-shelf: an astounding baseline for recognition,'' in \emph{CVPR
  workshops}, 2014, pp. 806--813.

\bibitem{gong2014multi}
Y.~Gong, L.~Wang, R.~Guo, and S.~Lazebnik, ``Multi-scale orderless pooling of
  deep convolutional activation features,'' in \emph{ECCV}, 2014.

\bibitem{yosinski2014transferable}
J.~Yosinski, J.~Clune, Y.~Bengio, and H.~Lipson, ``How transferable are
  features in deep neural networks?'' in \emph{NeurIPS}, 2014, pp. 3320--3328.

\bibitem{tolias2015particular}
G.~Tolias, R.~Sicre, and H.~J{\'e}gou, ``Particular object retrieval with
  integral max-pooling of {CNN} activations,'' in \emph{ICLR}, 2015, pp. 1--12.

\bibitem{jimenez2017class}
A.~Jim{\'e}nez, J.~M. Alvarez, and X.~Gir{\'o}~Nieto, ``Class-weighted
  convolutional features for visual instance search,'' in \emph{BMVC}, 2017,
  pp. 1--12.

\bibitem{do2018selective}
T.-T. Do, T.~Hoang, D.-K.~L. Tan, H.~Le, T.~V. Nguyen, and N.-M. Cheung, ``From
  selective deep convolutional features to compact binary representations for
  image retrieval,'' \emph{ACM Trans. Multimedia Comput. Commun. Appl.},
  vol.~15, no.~2, pp. 1--22, 2018.

\bibitem{xu2018unsupervised}
J.~Xu, C.~Wang, C.~Qi, C.~Shi, and B.~Xiao, ``Unsupervised part-based weighting
  aggregation of deep convolutional features for image retrieval,'' in
  \emph{AAAI}, 2018, pp. 7436--7443.

\bibitem{li2016exploiting}
Y.~Li, X.~Kong, L.~Zheng, and Q.~Tian, ``Exploiting hierarchical activations of
  neural network for image retrieval,'' in \emph{ACM MM}, 2016.

\bibitem{sharif2015baseline}
A.~Sharif~Razavian, J.~Sullivan, A.~Maki, and S.~Carlsson, ``A baseline for
  visual instance retrieval with deep convolutional networks,'' in \emph{ICLR},
  2015.

\bibitem{do2017embedding}
T.-T. Do and N.-M. Cheung, ``Embedding based on function approximation for
  large scale image search,'' \emph{IEEE Trans. Pattern Anal. Mach. Intell.},
  vol.~40, no.~3, pp. 626--638, 2017.

\bibitem{babenko2014neural}
A.~Babenko, A.~Slesarev, A.~Chigorin, and V.~Lempitsky, ``Neural codes for
  image retrieval,'' in \emph{ECCV}, 2014, pp. 584--599.

\bibitem{garcia2017learning}
N.~Garcia and G.~Vogiatzis, ``Learning non-metric visual similarity for image
  retrieval,'' \emph{Image Vis. Comput.}, vol.~82, pp. 18--25, 2019.

\bibitem{ong2017siamese}
E.-J. Ong, S.~Husain, and M.~Bober, ``Siamese network of deep fisher-vector
  descriptors for image retrieval,'' \emph{arXiv preprint arXiv:1702.00338},
  2017.

\bibitem{gordo2016deep}
A.~Gordo, J.~Almaz{\'a}n, J.~Revaud, and D.~Larlus, ``Deep image retrieval:
  Learning global representations for image search,'' in \emph{ECCV}, 2016.

\bibitem{arandjelovic2016netvlad}
R.~Arandjelovic, P.~Gronat, A.~Torii, T.~Pajdla, and J.~Sivic, ``{NetVLAD}:
  {CNN} architecture for weakly supervised place recognition,'' in \emph{CVPR},
  2015, pp. 5297--5307.

\bibitem{radenovic2016cnn}
F.~Radenovi{\'c}, G.~Tolias, and O.~Chum, ``{CNN} image retrieval learns from
  {BoW}: Unsupervised fine-tuning with hard examples,'' in \emph{ECCV}, 2016,
  pp. 3--20.

\bibitem{xu2017iterative}
J.~Xu, C.~Wang, C.~Qi, C.~Shi, and B.~Xiao, ``Iterative manifold embedding
  layer learned by incomplete data for large-scale image retrieval,''
  \emph{IEEE Trans. Multimedia}, vol.~21, no.~6, pp. 1551--1562, 2017.

\bibitem{radenovic2018fine}
F.~Radenovi{\'c}, G.~Tolias, and O.~Chum, ``Fine-tuning {CNN} image retrieval
  with no human annotation,'' \emph{IEEE Trans. Pattern Anal. Mach. Intell.},
  vol.~41, no.~7, pp. 1655--1668, 2018.

\bibitem{liu2019guided}
C.~Liu, G.~Yu, M.~Volkovs, C.~Chang, H.~Rai, J.~Ma, and S.~K. Gorti, ``Guided
  similarity separation for image retrieval,'' in \emph{NeurIPS}, 2019.

\bibitem{chang2019explore}
C.~Chang, G.~Yu, C.~Liu, and M.~Volkovs, ``Explore-exploit graph traversal for
  image retrieval,'' in \emph{CVPR}, 2019, pp. 9423--9431.

\bibitem{simonyan2014very}
K.~Simonyan and A.~Zisserman, ``Very deep convolutional networks for
  large-scale image recognition,'' \emph{arXiv preprint arXiv:1409.1556}, 2014.

\bibitem{yue2015exploiting}
J.~Yue-Hei~Ng, F.~Yang, and L.~S. Davis, ``Exploiting local features from deep
  networks for image retrieval,'' in \emph{CVPR workshops}, 2015.

\bibitem{jegou2008hamming}
H.~Jegou, M.~Douze, and C.~Schmid, ``Hamming embedding and weak geometric
  consistency for large scale image search,'' in \emph{ECCV}, 2008.

\bibitem{liu2015deepindex}
Y.~Liu, Y.~Guo, S.~Wu, and M.~S. Lew, ``Deepindex for accurate and efficient
  image retrieval,'' in \emph{ICMR}, 2015, pp. 43--50.

\bibitem{yan2016cnn}
K.~Yan, Y.~Wang, D.~Liang, T.~Huang, and Y.~Tian, ``{CNN} vs. {SIFT} for image
  retrieval: Alternative or complementary?'' in \emph{ACM MM}, 2016.

\bibitem{jegou2010aggregating}
H.~J{\'e}gou, M.~Douze, C.~Schmid, and P.~P{\'e}rez, ``Aggregating local
  descriptors into a compact image representation,'' in \emph{CVPR}, 2010.

\bibitem{perronnin2007fisher}
F.~Perronnin and C.~Dance, ``Fisher kernels on visual vocabularies for image
  categorization,'' in \emph{CVPR}, 2007, pp. 1--8.

\bibitem{simeoni2019local}
O.~Sim{\'e}oni, Y.~Avrithis, and O.~Chum, ``Local features and visual words
  emerge in activations,'' in \emph{CVPR}, 2019, pp. 11\,651--11\,660.

\bibitem{revaud2019learning}
J.~Revaud, J.~Almaz{\'a}n, R.~S. Rezende, and C.~R.~d. Souza, ``Learning with
  average precision: Training image retrieval with a listwise loss,'' in
  \emph{ICCV}, 2019, pp. 5107--5116.

\bibitem{bai2018regularized}
B.~Song, X.~Bai, Q.~Tian, and L.~J. Latecki, ``Regularized diffusion process on
  bidirectional context for object retrieval,'' \emph{IEEE Trans. Pattern Anal.
  Mach. Intell.}, vol.~41, no.~5, pp. 1213--1226, 2018.

\bibitem{el2021training}
A.~El-Nouby, N.~Neverova, I.~Laptev, and H.~J{\'e}gou, ``Training vision
  transformers for image retrieval,'' \emph{arXiv preprint arXiv:2102.05644},
  2021.

\bibitem{tan2021instance}
F.~Tan, J.~Yuan, and V.~Ordonez, ``Instance-level image retrieval using
  reranking transformers,'' \emph{ICCV}, 2021.

\bibitem{reddy2015object}
K.~Reddy~Mopuri and R.~Venkatesh~Babu, ``Object level deep feature pooling for
  compact image representation,'' in \emph{CVPR Workshops}, 2015.

\bibitem{morere2017nested}
O.~Morere, J.~Lin, A.~Veillard, L.-Y. Duan, V.~Chandrasekhar, and T.~Poggio,
  ``Nested invariance pooling and {RBM} hashing for image instance retrieval,''
  in \emph{ICMR}, 2017, pp. 260--268.

\bibitem{weyand2020google}
T.~Weyand, A.~Araujo, B.~Cao, and J.~Sim, ``Google landmarks dataset v2-a
  large-scale benchmark for instance-level recognition and retrieval,'' in
  \emph{CVPR}, 2020, pp. 2575--2584.

\bibitem{song2017deep}
J.~Song, T.~He, L.~Gao, X.~Xu, and H.~T. Shen, ``Deep region hashing for
  efficient large-scale instance search from images,'' in \emph{AAAI}, 2018.

\bibitem{lin2018unsupervised}
K.~Lin, J.~Lu, C.-S. Chen, J.~Zhou, and M.-T. Sun, ``Unsupervised deep learning
  of compact binary descriptors,'' \emph{IEEE Trans. Pattern Anal. Mach.
  Intell.}, 2018.

\bibitem{zhou2017collaborative}
W.~Zhou, H.~Li, J.~Sun, and Q.~Tian, ``Collaborative index embedding for image
  retrieval,'' \emph{IEEE Trans. Pattern Anal. Mach. Intell.}, vol.~40, no.~5,
  pp. 1154--1166, 2017.

\bibitem{liu2016deepvehicles}
H.~Liu, Y.~Tian, Y.~Yang, L.~Pang, and T.~Huang, ``Deep relative distance
  learning: Tell the difference between similar vehicles,'' in \emph{CVPR},
  2016.

\bibitem{ozaki2019large}
K.~Ozaki and S.~Yokoo, ``Large-scale landmark retrieval/recognition under a
  noisy and diverse dataset,'' in \emph{CVPR Workshop}, 2019.

\bibitem{razavian2016visual}
A.~S. Razavian, J.~Sullivan, S.~Carlsson, and A.~Maki, ``Visual instance
  retrieval with deep convolutional networks,'' \emph{ITE Trans. Media Technol.
  Appl.}, vol.~4, no.~3, pp. 251--258, 2016.

\bibitem{pang2018unifying}
S.~Pang, J.~Xue, J.~Zhu, L.~Zhu, and Q.~Tian, ``Unifying sum and weighted
  aggregations for efficient yet effective image representation computation,''
  \emph{IEEE Trans. Image Process.}, vol.~28, no.~2, pp. 841--852, 2018.

\bibitem{cao2020unifying}
B.~Cao, A.~Araujo, and J.~Sim, ``Unifying deep local and global features for
  efficient image search,'' in \emph{ECCV}, 2020, pp. 726--743.

\bibitem{fischler1981random}
M.~A. Fischler and R.~C. Bolles, ``Random sample consensus: a paradigm for
  model fitting with applications to image analysis and automated
  cartography,'' \emph{Communications of the ACM}, vol.~24, no.~6, pp.
  381--395, 1981.

\bibitem{tolias2016image}
G.~Tolias, Y.~Avrithis, and H.~J{\'e}gou, ``Image search with selective match
  kernels: aggregation across single and multiple images,'' \emph{Int. J.
  Comput. Vis.}, vol. 116, no.~3, pp. 247--261, 2016.

\bibitem{teichmann2019detect}
M.~Teichmann, A.~Araujo, M.~Zhu, and J.~Sim, ``Detect-to-retrieve: Efficient
  regional aggregation for image search,'' in \emph{CVPR}, 2019.

\bibitem{pang2018improving}
S.~Pang, J.~Ma, J.~Zhu, J.~Xue, and Q.~Tian, ``Improving object retrieval
  quality by integration of similarity propagation and query expansion,''
  \emph{IEEE Trans. Multimedia}, vol.~21, no.~3, pp. 760--770, 2018.

\bibitem{qi2017spatial}
C.~Qi, C.~Shi, J.~Xu, C.~Wang, and B.~Xiao, ``Spatial weighted fisher vector
  for image retrieval,'' in \emph{ICME}, 2017, pp. 463--468.

\bibitem{kim2017learned}
H.~J. Kim, E.~Dunn, and J.-M. Frahm, ``Learned contextual feature reweighting
  for image geo-localization,'' in \emph{CVPR}, 2017, pp. 3251--3260.

\bibitem{mohedano2017saliency}
E.~Mohedano, K.~McGuinness, X.~Gir{\'o}-i Nieto, and N.~E. O'Connor, ``Saliency
  weighted convolutional features for instance search,'' in \emph{CBMI}, 2018,
  pp. 1--6.

\bibitem{yang2017two}
F.~Yang, J.~Li, S.~Wei, Q.~Zheng, T.~Liu, and Y.~Zhao, ``Two-stream attentive
  {CNN}s for image retrieval,'' in \emph{ACM MM}, 2017, pp. 1513--1521.

\bibitem{yang2018supervised}
H.-F. Yang, K.~Lin, and C.-S. Chen, ``Supervised learning of
  semantics-preserving hash via deep convolutional neural networks,''
  \emph{IEEE Trans. Pattern Anal. Mach. Intell.}, vol.~40, no.~2, pp. 437--451,
  2018.

\bibitem{liu2018deep}
Y.~Liu, J.~Song, K.~Zhou, L.~Yan, L.~Liu, F.~Zou, and L.~Shao, ``Deep
  self-taught hashing for image retrieval,'' \emph{IEEE Trans Cybern.},
  vol.~49, no.~6, pp. 2229--2241, 2018.

\bibitem{gordo2017end}
A.~Gordo, J.~Almazan, J.~Revaud, and D.~Larlus, ``End-to-end learning of deep
  visual representations for image retrieval,'' \emph{Int. J. Comput. Vis.},
  vol. 124, no.~2, pp. 237--254, 2017.

\bibitem{azizpour2016factors}
H.~Azizpour, A.~S. Razavian, J.~Sullivan, A.~Maki, and S.~Carlsson, ``Factors
  of transferability for a generic convnet representation,'' \emph{IEEE Trans.
  Pattern Anal. Mach. Intell.}, vol.~38, no.~9, pp. 1790--1802, 2016.

\bibitem{mohedano2016bags}
E.~Mohedano, K.~McGuinness, N.~E. O'Connor, A.~Salvador, F.~Marqu{\'e}s, and
  X.~Giro-i Nieto, ``Bags of local convolutional features for scalable instance
  search,'' in \emph{ICMR}, 2016, pp. 327--331.

\bibitem{zhao2017spatial}
W.~Zhao, H.~Luo, J.~Peng, and J.~Fan, ``Spatial pyramid deep hashing for
  large-scale image retrieval,'' \emph{Neurocomputing}, vol. 243, pp. 166--173,
  2017.

\bibitem{zheng2016accurate}
L.~Zheng, S.~Wang, J.~Wang, and Q.~Tian, ``Accurate image search with
  multi-scale contextual evidences,'' \emph{Int. J. Comput. Vis.}, vol. 120,
  no.~1, pp. 1--13, 2016.

\bibitem{cao2016focus}
J.~Cao, L.~Liu, P.~Wang, Z.~Huang, C.~Shen, and H.~T. Shen, ``Where to focus:
  Query adaptive matching for instance retrieval using convolutional feature
  maps,'' \emph{arXiv preprint arXiv:1606.06811}, 2016.

\bibitem{zitnick2014edge}
C.~L. Zitnick and P.~Doll{\'a}r, ``Edge boxes: Locating object proposals from
  edges,'' in \emph{ECCV}, 2014, pp. 391--405.

\bibitem{yu2017efficient}
T.~Yu, Y.~Wu, S.~D. Bhattacharjee, and J.~Yuan, ``Efficient object instance
  search using fuzzy objects matching.'' in \emph{AAAI}, 2017, pp. 4320--4326.

\bibitem{sun2015scalable}
S.~Sun, W.~Zhou, Q.~Tian, and H.~Li, ``Scalable object retrieval with compact
  image representation from generic object regions,'' \emph{ACM Trans.
  Multimedia Comput. Commun. Appl.}, vol.~12, no.~2, pp. 1--21, 2015.

\bibitem{mairal2014convolutional}
J.~Mairal, P.~Koniusz, Z.~Harchaoui, and C.~Schmid, ``Convolutional kernel
  networks,'' in \emph{NeurIPS}, 2014, pp. 2627--2635.

\bibitem{salvador2016faster}
A.~Salvador, X.~Gir{\'o}-i Nieto, F.~Marqu{\'e}s, and S.~Satoh, ``Faster
  {R-CNN} features for instance search,'' in \emph{CVPR Workshops}, 2016, pp.
  9--16.

\bibitem{wang2015instre}
S.~Wang and S.~Jiang, ``{INSTRE}: a new benchmark for instance-level object
  retrieval and recognition,'' \emph{ACM Trans. Multimedia Comput. Commun.
  Appl.}, vol.~11, no.~3, pp. 37:1--37:21, 2015.

\bibitem{philbin2007object}
J.~Philbin, O.~Chum, M.~Isard, J.~Sivic, and A.~Zisserman, ``Object retrieval
  with large vocabularies and fast spatial matching,'' in \emph{CVPR}, 2007,
  pp. 1--8.

\bibitem{Lostphilbin2008lost1}
------, ``Lost in quantization: Improving particular object retrieval in large
  scale image databases,'' in \emph{CVPR}, 2008, pp. 1--8.

\bibitem{ng2020solar}
T.~Ng, V.~Balntas, Y.~Tian, and K.~Mikolajczyk, ``{SOLAR}: Second-order loss
  and attention for image retrieval,'' in \emph{ECCV}, 2020, pp. 253--270.

\bibitem{zheng2016good}
L.~Zheng, Y.~Zhao, S.~Wang, J.~Wang, and Q.~Tian, ``Good practice in {CNN}
  feature transfer,'' \emph{CoRR}, vol. abs/1604.00133, 2016.

\bibitem{lou2018multi}
Y.~Lou, Y.~Bai, S.~Wang, and L.-Y. Duan, ``Multi-scale context attention
  network for image retrieval,'' in \emph{ACM MM}, 2018, pp. 1128--1136.

\bibitem{li2017ms}
Y.~Li, Y.~Xu, J.~Wang, Z.~Miao, and Y.~Zhang, ``{MS-RMAC}: Multiscale regional
  maximum activation of convolutions for image retrieval,'' \emph{IEEE Signal
  Process Lett.}, vol.~24, no.~5, pp. 609--613, 2017.

\bibitem{xiang2019multiple}
X.~Xiang, Z.~Wang, Z.~Zhao, and F.~Su, ``Multiple saliency and channel
  sensitivity network for aggregated convolutional feature,'' in \emph{AAAI},
  vol.~33, no.~01, 2019, pp. 9013--9020.

\bibitem{pang2018deep}
S.~Pang, J.~Ma, J.~Xue, J.~Zhu, and V.~Ordonez, ``Deep feature aggregation and
  image re-ranking with heat diffusion for image retrieval,'' \emph{IEEE Trans.
  Multimedia}, vol.~21, no.~6, pp. 1513--1523, 2018.

\bibitem{yu2017exploiting}
W.~Yu, K.~Yang, H.~Yao, X.~Sun, and P.~Xu, ``Exploiting the complementary
  strengths of multi-layer {CNN} features for image retrieval,''
  \emph{Neurocomputing}, vol. 237, pp. 235--241, 2017.

\bibitem{yang2021dolg}
M.~Yang, D.~He, M.~Fan, B.~Shi, X.~Xue, F.~Li, E.~Ding, and J.~Huang, ``Dolg:
  Single-stage image retrieval with deep orthogonal fusion of local and global
  features,'' in \emph{ICCV}, 2021, pp. 11\,772--11\,781.

\bibitem{zhang2019effective}
Z.~Zhang, Y.~Xie, W.~Zhang, and Q.~Tian, ``Effective image retrieval via
  multilinear multi-index fusion,'' \emph{IEEE Trans. Multimedia}, vol.~21,
  no.~11, pp. 2878--2890, 2019.

\bibitem{wang2019improving}
Q.~Wang, J.~Lai, Z.~Yang, K.~Xu, P.~Kan, W.~Liu, and L.~Lei, ``Improving
  cross-dimensional weighting pooling with multi-scale feature fusion for image
  retrieval,'' \emph{Neurocomputing}, vol. 363, pp. 17--26, 2019.

\bibitem{Zheng2015QueryadaptiveLF}
L.~Zheng, S.~Wang, L.~Tian, F.~He, Z.~Liu, and Q.~Tian, ``Query-adaptive late
  fusion for image search and person re-identification,'' in \emph{CVPR}, 2015,
  pp. 1741--1750.

\bibitem{xuan2018deep}
H.~Xuan, R.~Souvenir, and R.~Pless, ``Deep randomized ensembles for metric
  learning,'' in \emph{ECCV}, 2018, pp. 723--734.

\bibitem{chen2019analysis}
B.-C. Chen, L.~S. Davis, and S.-N. Lim, ``An analysis of object embeddings for
  image retrieval,'' \emph{arXiv preprint arXiv:1905.11903}.

\bibitem{wang2014hamming}
F.~Wang, W.-L. Zhao, C.-W. Ngo, and B.~Merialdo, ``A hamming embedding kernel
  with informative bag-of-visual words for video semantic indexing,'' \emph{ACM
  Trans. Multimedia Comput. Commun. Appl.}, vol.~10, no.~3, pp. 1--20, 2014.

\bibitem{mishchuk2017working}
A.~Mishchuk, D.~Mishkin, F.~Radenovi{\'c}, and J.~Matas, ``Working hard to know
  your neighbor's margins: Local descriptor learning loss,'' in \emph{NeurIPS},
  2017, pp. 4827--4838.

\bibitem{mukherjee2020bag}
A.~Mukherjee, J.~Sil, A.~Sahu, and A.~S. Chowdhury, ``A bag of constrained
  informative deep visual words for image retrieval,'' \emph{Pattern
  Recognition Letters}, vol. 129, pp. 158--165, 2020.

\bibitem{paulin2015local}
M.~Paulin, M.~Douze, Z.~Harchaoui, J.~Mairal, F.~Perronin, and C.~Schmid,
  ``Local convolutional features with unsupervised training for image
  retrieval,'' in \emph{ICCV}, 2015, pp. 91--99.

\bibitem{jegou2012aggregating}
H.~Jegou, F.~Perronnin, M.~Douze, J.~S{\'a}nchez, P.~Perez, and C.~Schmid,
  ``Aggregating local image descriptors into compact codes,'' \emph{IEEE Trans.
  Pattern Anal. Mach. Intell.}, vol.~34, no.~9, pp. 1704--1716, 2012.

\bibitem{sanchez2013image}
J.~S{\'a}nchez, F.~Perronnin, T.~Mensink, and J.~Verbeek, ``Image
  classification with the fisher vector: Theory and practice,'' \emph{Int. J.
  Comput. Vis.}, vol. 105, no.~3, pp. 222--245, 2013.

\bibitem{cao2017local}
J.~Cao, Z.~Huang, and H.~T. Shen, ``Local deep descriptors in bag-of-words for
  image retrieval,'' in \emph{ACM MM}, 2017, pp. 52--58.

\bibitem{dusmanu2019d2}
M.~Dusmanu, I.~Rocco, T.~Pajdla, M.~Pollefeys, J.~Sivic, A.~Torii, and
  T.~Sattler, ``D2-net: A trainable cnn for joint description and detection of
  local features,'' in \emph{CVPR}, 2019, pp. 8092--8101.

\bibitem{mikolajczyk2004scale}
K.~Mikolajczyk and C.~Schmid, ``Scale \& affine invariant interest point
  detectors,'' \emph{Int. J. Comput. Vis.}, vol.~60, no.~1, pp. 63--86, 2004.

\bibitem{tolias2020learning}
G.~Tolias, T.~Jenicek, and O.~Chum, ``Learning and aggregating deep local
  descriptors for instance-level recognition,'' in \emph{ECCV}, 2020.

\bibitem{yang2018dynamic}
J.~Yang, J.~Liang, H.~Shen, K.~Wang, P.~L. Rosin, and M.-H. Yang, ``Dynamic
  match kernel with deep convolutional features for image retrieval,''
  \emph{IEEE Trans. Image Process.}, vol.~27, no.~11, pp. 5288--5302, 2018.

\bibitem{kimregional2018Regional}
J.~Kim and S.-E. Yoon, ``Regional attention based deep feature for image
  retrieval,'' in \emph{BMVC}, 2018, pp. 209--223.

\bibitem{wei2019saliency}
S.~Wei, L.~Liao, J.~Li, Q.~Zheng, F.~Yang, and Y.~Zhao, ``Saliency inside:
  Learning attentive {CNN}s for content-based image retrieval,'' \emph{IEEE
  Trans. Image Process.}, vol.~28, no.~9, pp. 4580--4593, 2019.

\bibitem{do2017simultaneous}
T.-T. Do, D.-K. Le~Tan, T.~T. Pham, and N.-M. Cheung, ``Simultaneous feature
  aggregating and hashing for large-scale image search,'' in \emph{CVPR}, 2017,
  pp. 6618--6627.

\bibitem{musgrave2020metric}
K.~Musgrave, S.~Belongie, and S.-N. Lim, ``A metric learning reality check,''
  in \emph{ECCV}, 2020, pp. 681--699.

\bibitem{lv2018retrieval}
Y.~Lv, W.~Zhou, Q.~Tian, S.~Sun, and H.~Li, ``Retrieval oriented deep feature
  learning with complementary supervision mining,'' \emph{IEEE Trans. Image
  Process.}, vol.~27, no.~10, pp. 4945--4957, 2018.

\bibitem{min2020two}
W.~Min, S.~Mei, Z.~Li, and S.~Jiang, ``A two-stage triplet network training
  framework for image retrieval,'' \emph{IEEE Trans. Multimedia}, vol.~22,
  no.~12, pp. 3128--3138, 2020.

\bibitem{Deng2019ArcFaceAA}
J.~Deng, J.~Guo, and S.~Zafeiriou, ``Arcface: Additive angular margin loss for
  deep face recognition,'' \emph{CVPR}, pp. 4685--4694, 2019.

\bibitem{boudiaf2020unifying}
M.~Boudiaf, J.~Rony, I.~M. Ziko, E.~Granger, M.~Pedersoli, P.~Piantanida, and
  I.~B. Ayed, ``A unifying mutual information view of metric learning:
  cross-entropy vs. pairwise losses,'' in \emph{ECCV}, 2020, pp. 548--564.

\bibitem{lin2017deephash}
J.~Lin, O.~Morere, A.~Veillard, L.-Y. Duan, H.~Goh, and V.~Chandrasekhar,
  ``Deephash for image instance retrieval: Getting regularization, depth and
  fine-tuning right,'' in \emph{ICMR}, 2017, pp. 133--141.

\bibitem{cao2016quartet}
J.~Cao, Z.~Huang, P.~Wang, C.~Li, X.~Sun, and H.~T. Shen, ``Quartet-net
  learning for visual instance retrieval,'' in \emph{ACM MM}, 2016, pp.
  456--460.

\bibitem{brown2020smooth}
A.~Brown, W.~Xie, V.~Kalogeiton, and A.~Zisserman, ``Smooth-{AP}: Smoothing the
  path towards large-scale image retrieval,'' in \emph{ECCV}, 2020, pp.
  677--694.

\bibitem{iscen2018mining}
A.~Iscen, G.~Tolias, Y.~Avrithis, and O.~Chum, ``Mining on manifolds: Metric
  learning without labels,'' in \emph{CVPR}, 2018, pp. 7642--7651.

\bibitem{iscen2018fast}
A.~Iscen, Y.~Avrithis, G.~Tolias, T.~Furon, and O.~Chum, ``Fast spectral
  ranking for similarity search,'' in \emph{CVPR}, 2018, pp. 7632--7641.

\bibitem{iscen2017efficient}
A.~Iscen, G.~Tolias, Y.~Avrithis, T.~Furon, and O.~Chum, ``Efficient diffusion
  on region manifolds: Recovering small objects with compact {CNN}
  representations,'' in \emph{CVPR}, 2017, pp. 2077--2086.

\bibitem{donoser2013diffusion}
M.~Donoser and H.~Bischof, ``Diffusion processes for retrieval revisited,'' in
  \emph{CVPR}, 2013, pp. 1320--1327.

\bibitem{zhaomodelling2018}
Y.~Zhao, L.~Wang, L.~Zhou, Y.~Shi, and Y.~Gao, ``Modelling diffusion process by
  deep neural networks for image retrieval,'' in \emph{BMVC}, 2018.

\bibitem{kipf2016semi}
T.~N. Kipf and M.~Welling, ``Semi-supervised classification with graph
  convolutional networks,'' in \emph{ICLR}, 2017.

\bibitem{tzelepi2018deep}
T.~Maria and T.~Anastasios, ``Deep convolutional image retrieval: A general
  framework,'' \emph{Signal Process. Image Commun.}, vol.~63, pp. 30--43, 2018.

\bibitem{caron2018deep}
M.~Caron, P.~Bojanowski, A.~Joulin, and M.~Douze, ``Deep clustering for
  unsupervised learning of visual features,'' in \emph{ECCV}, 2018, pp.
  132--149.

\bibitem{jiang2019unsupervised}
W.~Jiang, Y.~Wu, C.~Jing, T.~Yu, and Y.~Jia, ``Unsupervised deep quantization
  for object instance search,'' \emph{Neurocomputing}, vol. 362, pp. 60--71,
  2019.

\bibitem{ke2018feature}
B.~Ke, J.~Shao, Z.~Huang, and H.~T. Shen, ``Feature reconstruction by laplacian
  eigenmaps for efficient instance search,'' in \emph{ICMR}, 2018.

\bibitem{shen2018matchable}
T.~Shen, Z.~Luo, L.~Zhou, R.~Zhang, S.~Zhu, T.~Fang, and L.~Quan, ``Matchable
  image retrieval by learning from surface reconstruction,'' in \emph{ACCV},
  2018, pp. 415--431.

\bibitem{hu2019towards}
J.~Hu, R.~Ji, H.~Liu, S.~Zhang, C.~Deng, and Q.~Tian, ``Towards visual feature
  translation,'' in \emph{CVPR}, 2019, pp. 3004--3013.

\bibitem{bai2021unsupervised}
C.~Bai, H.~Li, J.~Zhang, L.~Huang, and L.~Zhang, ``Unsupervised adversarial
  instance-level image retrieval,'' \emph{IEEE Trans. Multimedia}, 2021.

\bibitem{paulin2017convolutional}
M.~Paulin, J.~Mairal, M.~Douze, Z.~Harchaoui, F.~Perronnin, and C.~Schmid,
  ``Convolutional patch representations for image retrieval: an unsupervised
  approach,'' \emph{Int. J. Comput. Vis.}, vol. 121, no.~1, pp. 149--168, 2017.

\bibitem{nister2006scalable}
D.~Nister and H.~Stewenius, ``Scalable recognition with a vocabulary tree,'' in
  \emph{CVPR}, 2006, pp. 2161--2168.

\bibitem{radenovic2018revisiting}
F.~Radenovic, A.~Iscen, G.~Tolias, Y.~Avrithis, and O.~Chum, ``Revisiting
  oxford and paris: Large-scale image retrieval benchmarking,'' in \emph{CVPR},
  2018.

\bibitem{yokoo2020two}
S.~Yokoo, K.~Ozaki, E.~Simo-Serra, and S.~Iizuka, ``Two-stage discriminative
  re-ranking for large-scale landmark retrieval,'' in \emph{CVPR Workshops},
  2020, pp. 1012--1013.

\bibitem{alzu2017content}
A.~Alzu'bi, A.~Amira, and N.~Ramzan, ``Content-based image retrieval with
  compact deep convolutional features,'' \emph{Neurocomputing}, vol. 249, pp.
  95--105, 2017.

\bibitem{husain2019remap}
S.~S. Husain and M.~Bober, ``{REMAP}: Multi-layer entropy-guided pooling of
  dense {CNN} features for image retrieval,'' \emph{IEEE Trans. Image
  Process.}, vol.~28, no.~10, pp. 5201--5213, 2019.

\bibitem{alemu2020multi}
L.~T. Alemu and M.~Pelillo, ``Multi-feature fusion for image retrieval using
  constrained dominant sets,'' \emph{Image Vis Comput}, vol.~94, p. 103862,
  2020.

\bibitem{valem2020graph}
L.~P. Valem and D.~C.~G. Pedronette, ``Graph-based selective rank fusion for
  unsupervised image retrieval,'' \emph{Pattern Recognit Lett}, 2020.

\bibitem{yang2019efficient22}
F.~Yang, R.~Hinami, Y.~Matsui, S.~Ly, and S.~Satoh, ``Efficient image retrieval
  via decoupling diffusion into online and offline processing,'' in
  \emph{AAAI}, vol.~33, 2019, pp. 9087--9094.

\bibitem{yang2016cross}
H.-F. Yang, K.~Lin, and C.-S. Chen, ``Cross-batch reference learning for deep
  classification and retrieval,'' in \emph{ACM MM}, 2016, pp. 1237--1246.

\bibitem{ouyang2020collaborative}
J.~Ouyang, W.~Zhou, M.~Wang, Q.~Tian, and H.~Li, ``Collaborative image
  relevance learning for visual re-ranking,'' \emph{IEEE Trans. Multimedia},
  2020.

\bibitem{sermanet2013overfeat}
P.~Sermanet, D.~Eigen, X.~Zhang, M.~Mathieu, R.~Fergus, and Y.~LeCun,
  ``Overfeat: Integrated recognition, localization and detection using
  convolutional networks,'' in \emph{ICLR}, 2014.

\bibitem{chatfield2014return}
K.~Chatfield, K.~Simonyan, A.~Vedaldi, and A.~Zisserman, ``Return of the devil
  in the details: Delving deep into convolutional nets,'' in \emph{BMVC}, 2014.

\bibitem{li2019universal}
J.~Li, R.~Ji, H.~Liu, X.~Hong, Y.~Gao, and Q.~Tian, ``Universal perturbation
  attack against image retrieval,'' in \emph{ICCV}, 2019, pp. 4899--4908.

\bibitem{tolias2019targeted}
G.~Tolias, F.~Radenovic, and O.~Chum, ``Targeted mismatch adversarial attack:
  Query with a flower to retrieve the tower,'' in \emph{ICCV}, 2019.

\bibitem{triantafillou2017few}
E.~Triantafillou, R.~Zemel, and R.~Urtasun, ``Few-shot learning through an
  information retrieval lens,'' \emph{NeurIPS}, vol.~30, 2017.

\bibitem{Deng2022InsCLRII}
Z.~Deng, Y.~Zhong, S.~Guo, and W.~Huang, ``{InsCLR}: Improving instance
  retrieval with self-supervision,'' in \emph{AAAI}, 2022.

\bibitem{ibrahimi2022learning}
S.~Ibrahimi, A.~Sors, R.~S. de~Rezende, and S.~Clinchant, ``Learning with label
  noise for image retrieval by selecting interactions,'' in \emph{WACV}, 2022,
  pp. 2181--2190.

\end{thebibliography}

\vspace{-2.0 cm}
%% author- WeiChen
\begin{IEEEbiography}[{\includegraphics[width=1in,height=1.25in,clip,keepaspectratio]{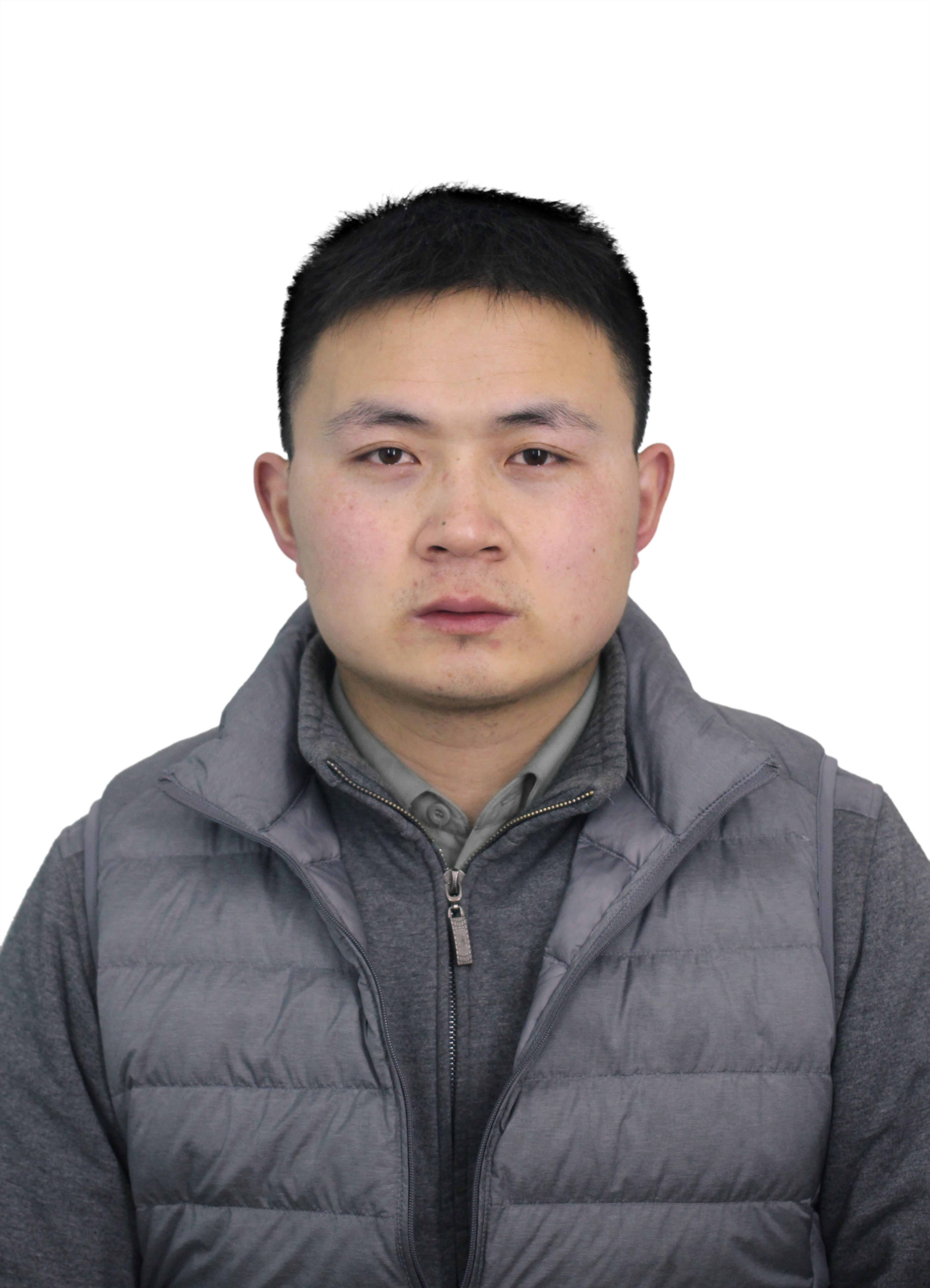}}]{Wei Chen} received the Ph.D. degree at Leiden University, in 2021. Before starting with PhD study in Leiden University, he received his Master degree from the National University of Defense Technology, China, in 2016. His research interest focuses on cross-modal retrieval with deep learning methods, and also in the context of incremental learning. He has published papers in international conferences and journal including CVPR, ACM MM, PR, Neurocomputing, and IEEE TMM \etc
\end{IEEEbiography}
\vspace{-1.7 cm} 

%% author
\begin{IEEEbiography}[{\includegraphics[width=1in,height=1.25in,clip,keepaspectratio]{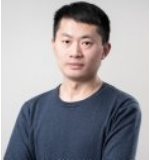}}]{Yu Liu} is currently an Associate Professor in International School
of Information Science and Engineering at Dalian University of Technology, China.
He was a post-doctoral researcher in the PSI group of KU Leuven, Belgium. 
In 2018, he received the PhD degree from Leiden University, Netherlands.
His research interests include object recognition and retrieval in the context of continual learning and zero-shot learning. 
He has co-organized several workshops at ICCV, CVPR and ECCV, respectively. 
He has published papers in CVPR, ICCV, ECCV, ACM MM, IEEE TIP, etc, and 
received the best paper award at MMM2017.
\end{IEEEbiography}
\vspace{-1.7 cm} 

%% author
\begin{IEEEbiography}[{\includegraphics[width=1in,height=1.25in,clip,keepaspectratio]{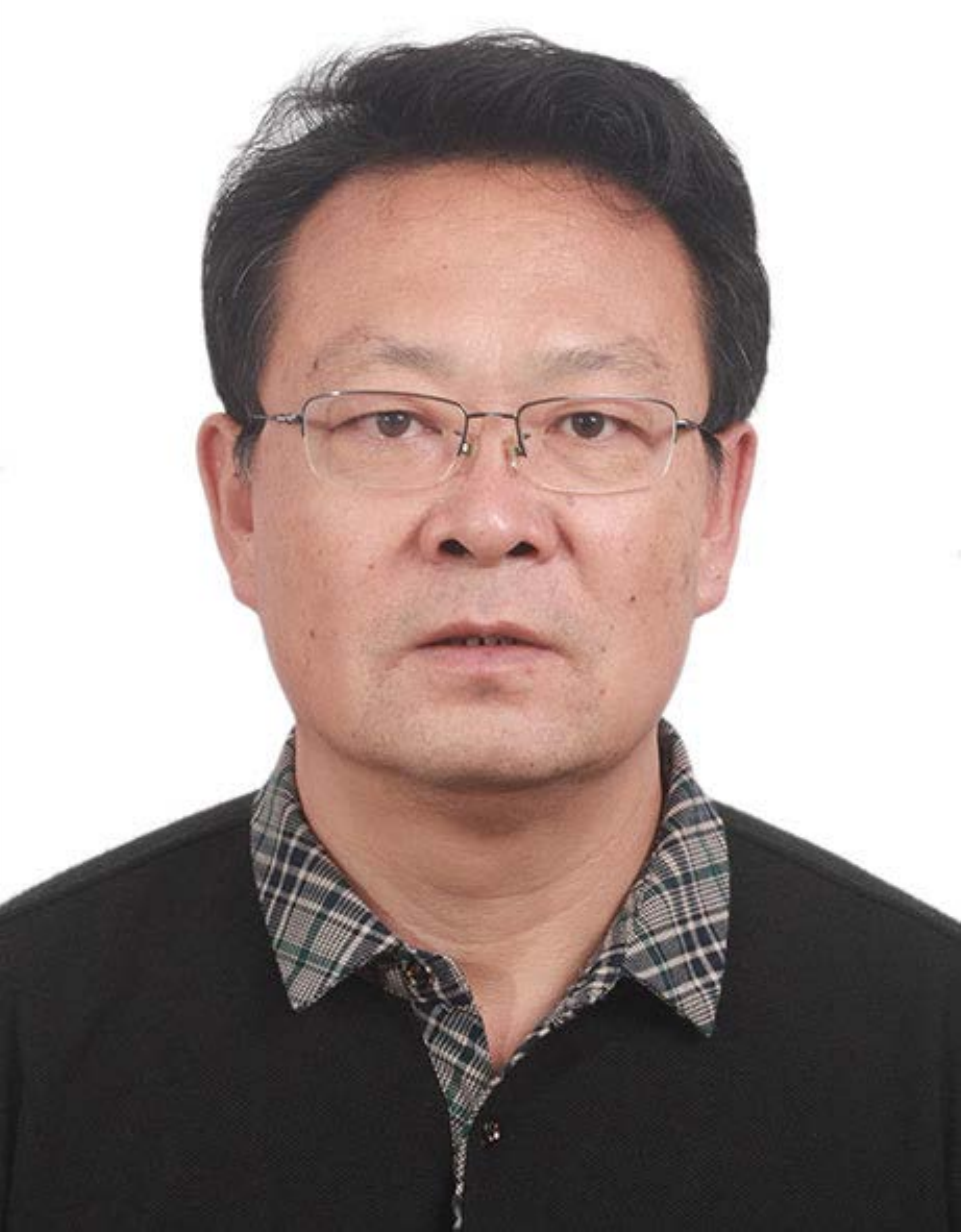}}]{Weiping Wang} received the Ph.D. degree in systems engineering from the National University of Defense Technology, Changsha, China. He is a Professor at Academy of Advanced Technology Research of Hunan. He has more than 200 papers published on journals and conferences including IEEE Transactions multimedia, Transactions on Vehicular Technology, \etc His current research interests focus on intelligent decision experimentation, multi-modal knowledge extraction and knowledge integrated computation.

\end{IEEEbiography}
\vspace{-2.0 cm} 

%% author
\begin{IEEEbiography}[{\includegraphics[width=1in,height=1.25in,clip,keepaspectratio]{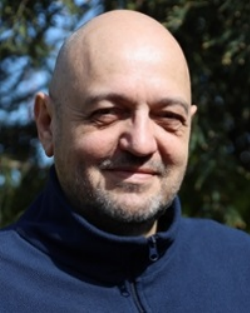}}]{Erwin M. Bakker} is co-director of the LIACS Media Lab at Leiden University. He has published widely in the fields of image retrieval, audio analysis and retrieval and bioinformatics. He was closely involved with the start of the International Conference on Image and Video Retrieval (CIVR). Moreover, he regularly serves as a program committee member or organizing committee member for scientific multimedia and human-computer interaction conferences and workshops.
\end{IEEEbiography}
\vspace{-2.0 cm} 

%%author
\begin{IEEEbiography}[{\includegraphics[width=1in,height=1.25in,clip,keepaspectratio]{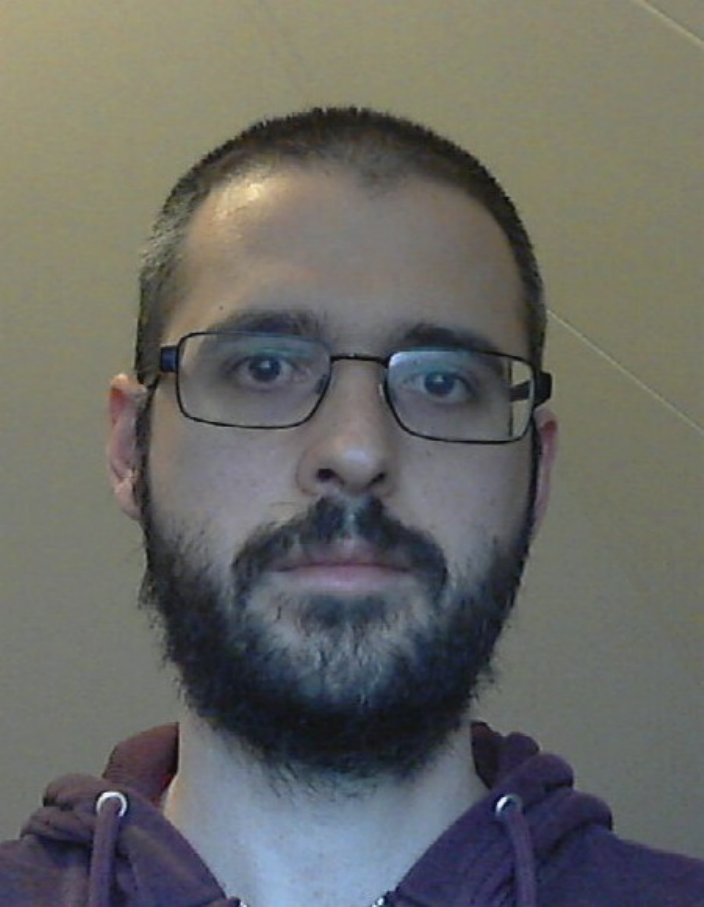}}]{Theodoros Georgiou} received the Ph.D. degree at Leiden University, in 2021. His research interest focuses on deep learning methods applied on higher than two dimensional data. Before starting with PhD study in Leiden University, he received his Master degree from the Leiden University in 2016. He has published  papers in international conferences  and journal including ICPR, CBMI, WCCI and IJMIR.
\end{IEEEbiography}

\vspace{-2.0 cm} 
%% author
\begin{IEEEbiography}[{\includegraphics[width=1in,height=1.25in,clip,keepaspectratio]{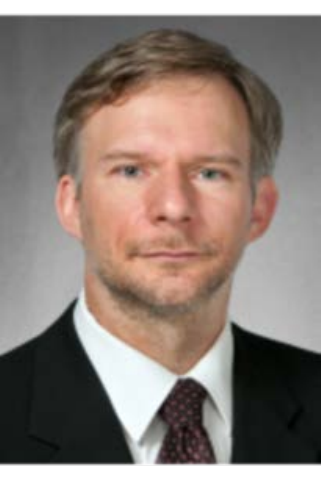}}]{Paul Fieguth} is co-director of the Vision \& Image Processing Group in Systems Design Engineering at the University of Waterloo, where he is Professor and Associate Dean.  He received the Ph.D.\ degree from the Massachusetts Institute of Technology, Cambridge, in 1995, and has held visiting appointments at the University of Heidelberg in Germany, at INRIA/Sophia in France, at the Cambridge Research Laboratory in Boston, at Oxford University and the Rutherford Appleton Laboratory in England. His research interests include statistical signal and image processing, hierarchical algorithms, data fusion, machine learning, and the interdisciplinary applications of such methods.  He has published textbooks on Statistical Image Processing and on Complex Systems theory.
\end{IEEEbiography}
\vspace{-1.7 cm}

%% author
\begin{IEEEbiography}[{\includegraphics[width=1in,height=1.25in,clip,keepaspectratio]{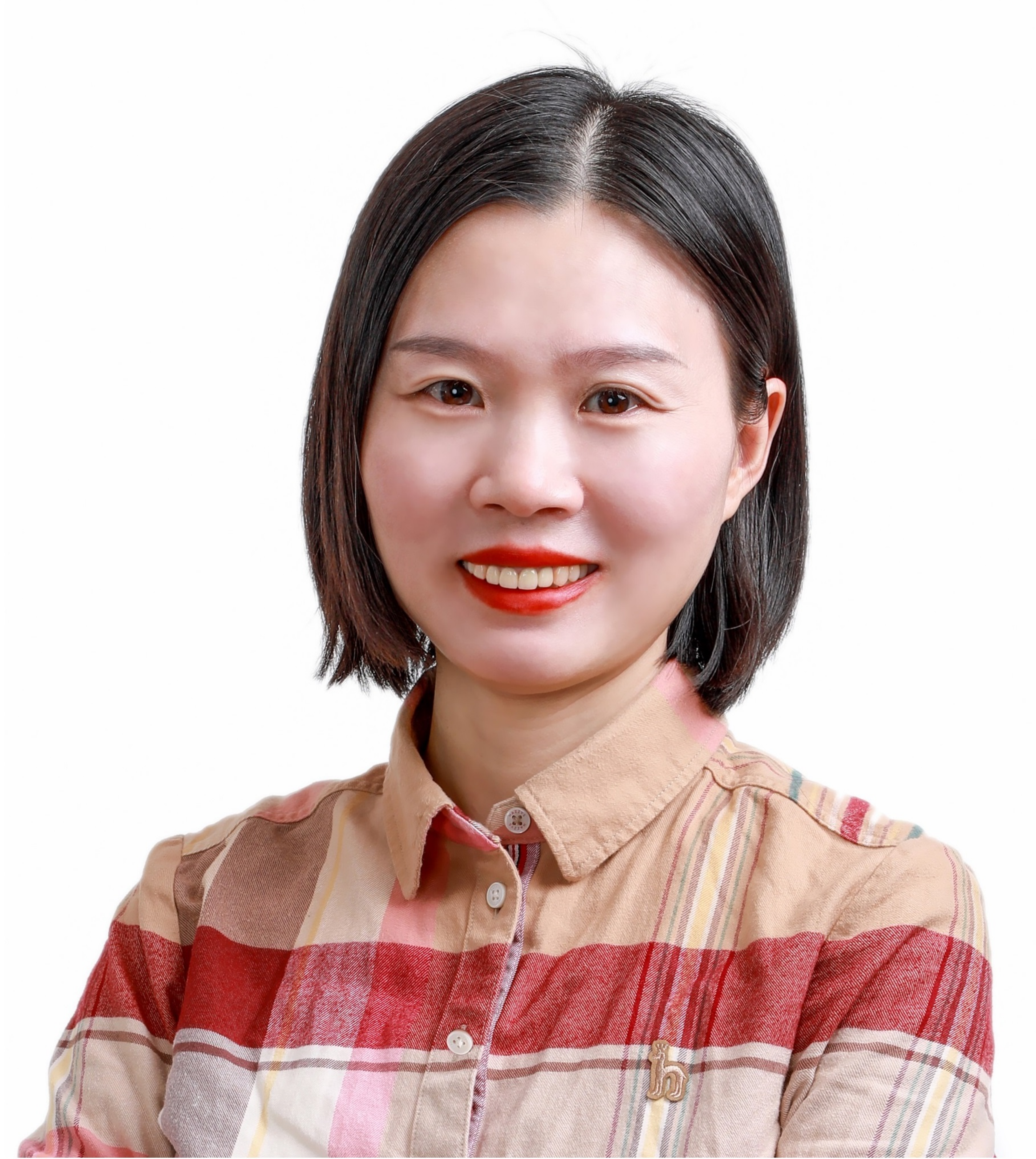}}]{Li Liu} received the Ph.D. degree in information and communication engineering from the National University of Defense Technology, China, in 2012. She is currently a Professor with the College of System Engineering. She has held visiting appointments at the University of Waterloo, Canada, at the Chinese University of Hong Kong, and at the University of Oulu, Finland. Her current research interests include computer vision, pattern recognition and machine learning. Her papers have currently over 7800 citations in Google Scholar.
\end{IEEEbiography}
\vspace{-1.7 cm}

%% author
\begin{IEEEbiography}[{\includegraphics[width=1in,height=1.25in,clip,keepaspectratio]{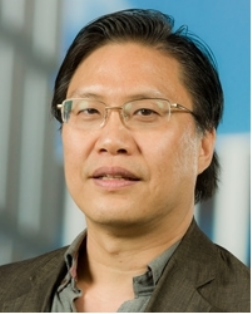}}]{Michael S. Lew} is the head of the Deep Learning and Computer Vision Research Group and a full Professor at LIACS, Leiden University. He has published over a dozen books and 190 peer-reviewed scientific articles in the areas of image retrieval, computer vision, and deep learning. Notably, he had the most cited paper in the ACM Transactions on Multimedia and one of the top 10 most cited articles in the history (out of more than 16,000 articles) of the ACM SIGMM.  He was also a founding member of the advisory committee for the TRECVID video retrieval evaluation project, chair of the steering committee for the ACM International Conference on Multimedia Retrieval and a member of the ACM SIGMM Executive Committee. 
\end{IEEEbiography}
\vspace{-1.7 cm}

%\end{IEEEbiography}

 \newpage

\end{document}